\documentclass[twoside,11pt]{article}

\usepackage{ArXiv_version}

\usepackage{lastpage}
\jmlrheading{23}{2022}{1-\pageref{LastPage}}{6/21; Revised 3/22}{4/22}{21-0622}{Shu Hu, Yiming Ying, Xin Wang, and Siwei Lyu}

\ShortHeadings{Sum of Ranked Range Loss for Supervised Learning}{Hu, Ying, Wang, and Lyu}

\usepackage[utf8]{inputenc} 
\usepackage[T1]{fontenc}    
\hypersetup{ hidelinks }
\usepackage{hyperref}       
\usepackage{url}            
\usepackage{booktabs}       
\usepackage{amsfonts}       
\usepackage{nicefrac}       
\usepackage{microtype}      
\usepackage{graphicx}
\usepackage{color}
\usepackage{amsmath,bm}
\usepackage{amsmath}
\usepackage{amsbsy}
\usepackage{amssymb}
\usepackage{dsfont}
\usepackage{mathrsfs}
\usepackage{tcolorbox,multicol}
\usepackage[ruled, vlined, linesnumbered]{algorithm2e}%
\usepackage{threeparttable}
\usepackage{wrapfig}
\usepackage{multirow}
\usepackage{cleveref}
\usepackage{subcaption}
\newcommand{\rulesep}{\unskip\ \vrule\ }
\usepackage{paralist}
\usepackage{makecell}
\usepackage{pifont}
\usepackage{paralist}
\usepackage{diagbox}
\usepackage{float}
\usepackage[format=hang]{caption}

\let\oldemptyset\emptyset
\let\emptyset\varnothing

\DeclareMathOperator*{\argmin}{argmin}

\def\aorr{\texttt{AoRR}}
\def\sorr{\texttt{SoRR}}
\def\tkml{\texttt{TKML}}

\def\X{\mathcal{X}}

\def\cE{\mathcal{E}}
\def\EX{\mathbb{E}}
\def\cR{\mathcal{R}}
\newcommand{\sgn}{\text{sign}}
\def\mbI{\mathbb{I}}
\def\R{\mathbb{R}}
\def\gl{\lambda}
\def\hgl{\hat{\lambda}}

\usepackage{enumerate}

\def\ie{{\em i.e.}}

\def\sl#1{\textcolor{orange}{[Siwei:#1]}}

\def\L{\mathcal{L}}

\begin{document}

\title{Sum of Ranked Range Loss for Supervised Learning}

\author{\name Shu Hu \email shuhu@buffalo.edu\\ 
        \addr Department of Computer Science and Engineering\\
      University at Buffalo, State University of New York\\
      Buffalo, NY 14260-2500, USA
       \AND
       \name Yiming Ying \email yying@albany.edu \\
      \addr Department of Mathematics and Statistics\\
      University at Albany, State University of New York\\
      Albany, NY 12222, USA
       \AND
       \name Xin Wang \email xwang264@buffalo.edu \\
      \addr Department of Computer Science and Engineering\\
      University at Buffalo, State University of New York\\
      Buffalo, NY 14260-2500, USA 
       \AND
      \name Siwei Lyu\thanks{Corresponding author.} \email siweilyu@buffalo.edu \\
      \addr Department of Computer Science and Engineering\\
      University at Buffalo, State University of New York\\
      Buffalo, NY 14260-2500, USA}

\editor{Lorenzo Rosasco}

\maketitle

\begin{abstract}
In forming learning objectives, one oftentimes needs to aggregate a set of individual values to a single output. Such cases occur in the aggregate loss, which  combines individual losses of a learning model over each training sample, and in the individual loss for multi-label learning, which combines prediction scores over all class labels. In this work, we introduce the \underline{s}um \underline{o}f \underline{r}anked \underline{r}ange (\sorr) as a general approach to form learning objectives. A ranked range is a consecutive sequence of sorted values of a set of real numbers. The minimization of \sorr~is solved with the difference of convex algorithm (DCA). We explore two applications in machine learning of the minimization of the \sorr~framework, namely the \aorr~aggregate loss for binary/multi-class classification at the sample level and the \tkml~individual loss for multi-label/multi-class classification at the label level. A combination loss of \aorr~and \tkml~is proposed as a new learning objective for improving the robustness of multi-label learning in the face of outliers in sample and labels alike. Our empirical results highlight the effectiveness of the proposed optimization frameworks and demonstrate the applicability of proposed losses using synthetic and real data sets. 
\end{abstract}

\begin{keywords}
  {Learning objective, Aggregate Loss, Rank-based Losses, Multi-class classification, Multi-label classification} 
\end{keywords}

\section{Introduction}


Learning objective is a fundamental component in any machine learning system. In forming learning objectives, we often need to aggregate a set of individual values to a single numerical value. Such cases occur in the aggregate loss, which combines individual losses of a learning model over each training sample, and in the individual loss for multi-label learning, which combines prediction scores over all class labels. In this paper, we refer to the loss over all training data as the \textbf{aggregate loss}, in order to distinguish it from the
\textbf{individual loss} that measures the quality of the model on a
single training example. For a set of real numbers representing individual values, the ranking order reflects the most basic relation among them.  Therefore, designing learning objectives can be achieved by choosing operations defined based on the ranking order of the individual values.

\begin{figure}
\centering
\includegraphics[width=0.8\textwidth]{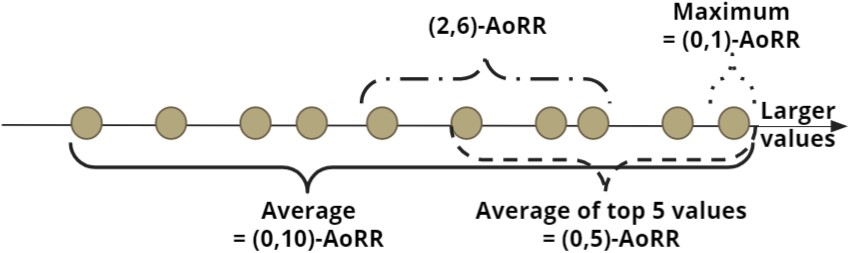}
\caption{ Illustrative examples of different approaches to aggregate individual values to form learning objectives in machine learning. \aorr~represents the average of ranked range method.
}
\label{fig:illustrate}
\end{figure}

Straightforward choices for such operations are the average and the maximum. Both are widely used in forming aggregate losses \citep{vapnik1992principles,shalev2016minimizing} and multi-label losses \citep{madjarov2012extensive}, yet each has its own drawbacks. The average is {\em insensitive} to minority sub-groups while the maximum is {\em sensitive} to outliers, which usually appear as the top individual values. The average top-$k$ loss is introduced as a compromise between the average and the maximum for aggregate loss \citep{fan2017learning} and multi-label individual loss \citep{fan2020groupwise}. However, it dilutes but not excludes the influences of the outliers. The situation is graphically illustrated in Figure \ref{fig:illustrate}.

In this work, we introduce the {\bf \underline{s}um \underline{o}f \underline{r}anked \underline{r}ange} (\sorr) as a new form of learning objectives that aggregate a set of individual values to a single value.  A ranked range is a consecutive sequence of sorted values of a set of real numbers. The \sorr~can be expressed as the difference between two sums of the top ranked values, which are convex functions themselves. As such, the \sorr~is the difference of two convex functions, and its optimization is an instance of the difference-of-convex (DC) programming problems \citep{le2018dc}. Therefore, it is natural to adopt the existing DC algorithm (DCA) \citep{tao1986algorithms} to efficiently solve the \sorr~related learning problems.

We explore two applications in machine learning of the minimization of the \sorr~framework. The first is to use the {\em average} of ranked range (\aorr) as an aggregate loss for binary/multi-class classification. Specifically, we consider individual logistic loss and individual hinge loss in \aorr~for binary classification. On the other hand, we consider softmax loss as an individual loss in \aorr~for multi-class classification. The \aorr~can be easily obtained by considering using an average operator on \sorr~(see Section \ref{sec:aorr} for more details).  Unlike previous aggregate losses, the \aorr~aggregate loss can completely eliminate the influence of outliers if their proportion in training data is known. We also propose a framework that can optimize \aorr~without prior knowledge about the proportion of outliers by utilizing a clean validation set extracted from the training data set. Second, we use a special case of \sorr~as a new type of individual loss for multi-label, the \textit{top-$k$ multi-label} (\tkml) loss, which explicitly encourages the true labels in the top $k$ range. Furthermore, we propose a \tkml-\aorr~loss, which combines the \aorr~and the \tkml~methods. The proposed loss can eliminate the outliers in the top-$k$ multi-label learning. Then a heuristic algorithm is designed to optimize \tkml-\aorr~.
The new learning objectives are tested and compared experimentally on several synthetic and real data sets\footnote{Code available at \url{https://github.com/discovershu/SoRR}.}. 

The main contributions of this work can be summarized as follows:
\begin{itemize}
\item We introduce \sorr~as a general learning objective and show that it can be formulated as the difference of two convex functions. 
\item Based on \sorr, we introduce the \aorr~aggregate loss for binary/multi-class classification at the sample level and the \tkml~individual loss for multi-label/multi-class learning at the label level. The two \sorr-based losses are further combined to form the \tkml-\aorr~loss that can improve the robustness of multi-label learning in the face of outliers in data and labels alike.
\item We also explore several theoretical aspects of \sorr-based losses, including connecting \aorr~to the Condition Value at Risk (CVaR) and explore its generalization property, establishing its classification calibration with regards to the optimal Bayes classifier. 
\item The learning of \sorr-based losses can be generally solved by the DC algorithm, but we also show their connections with  the bilevel optimization. Furthermore, we propose practical methods to better determine the hyper-parameters in \sorr-based losses.
\item We empirically demonstrate the robustness and effectiveness of the proposed \aorr, \tkml, \tkml-\aorr, and their optimization frameworks on both synthetic and real data sets.
\end{itemize} 

This paper significantly extends our previous conference paper \citep{hu2020learning} both theoretically and experimentally in the following aspects:  1) We establish the relations between \sorr~and the bilevel optimization problem \citep{borsos2020coresets} (Section \ref{sec:bilevel}); 2) The \aorr~aggregate loss is  reformulated by the Condition Value at Risks (CVaRs) and  its generalization property is studied (Section \ref{sec:cvar}); 3) We propose a new framework for automatically determining the hyper-parameters in \sorr-based losses and empirically demonstrate its effectiveness (Section \ref{sec:determin_k_m} and \ref{sec:aorr_multiclass}); 4) We combine \aorr~aggregate loss at the sample level and \tkml~individual loss at the label level to reduce the influence of the outliers in the top-$k$ multi-label learning procedure. We also propose a heuristic algorithm to optimize this combined loss and verify its effectiveness through experiments (Section \ref{sec:tkml-aorr}).

{The rest of the paper is organized as follows. In Section \ref{sec:related-work}, we summarize the existing works that are related to this work and discuss how this work differs from them. In Section \ref{sec:dca}, we define the sum of ranked range \sorr, formulate it as a DC problem and provide a DC algorithm for solving it. Then we connect the \sorr~to the bilevel optimization problem. In Section \ref{sec:aorr}, we propose a new form of aggregate loss named \aorr~aggregate loss based on \sorr~for binary and multi-class classification. We provide its interpretation and study its classification calibration property. We also connect it to the CVaR problem and provide a generalization bound. Furthermore, to make it more reliable in the practice, we propose a new framework for determining the hyper-parameters of \aorr. In Section \ref{sec:tkml}, we define \tkml~individual loss based on \sorr~for multi-label/multi-class learning. In Section \ref{sec:tkml-aorr}, we define \tkml-\aorr~loss, which combines \tkml~loss for label level and  \aorr~loss for sample level. The \tkml-\aorr~loss can eliminate the influence of the outliers in the top-$k$ multi-label learning. Then we propose a heuristic algorithm to optimize it. Section \ref{sec:conclusion} concludes the paper with discussions.} 


\section{{Related Work}} \label{sec:related-work}
There are a large body of works that focus on designing different forms of individual losses to solve specific problems and studying their properties, especially for binary classification. The earliest work can be traced back to 1960s by \cite{shuford1966admissible} and \cite{savage1971elicitation}. Specifically, \cite{shuford1966admissible} propose  admissible probability measurement procedures as individual losses to measure students' degree-of-belief probabilities in an educational environment. \cite{savage1971elicitation} characterizes scoring rules for probabilistic forecasts of categorical and binary variables.
More recent work of \cite{buja2005loss, masnadi2008design, lin2017focal, reid2010composite} continue to study this topic. For example, \cite{buja2005loss} study loss functions for binary class probability estimation
and classification. \cite{masnadi2008design} discuss  the robustness of outliers from a theoretical aspect in designing loss functions for classification. \cite{reid2010composite} study general composite binary losses. Many researchers prefer to select and use individual losses in terms of a convex function because of  the good global convergence properties in its optimization.
However, non-convex individual losses are also explored in recent work, such as \cite{lin2017focal, he2010maximum, wu2007robust, yu2010relaxed}. Concretely, to handle the imbalanced data problem, \cite{lin2017focal} design focal loss for object detection. To handle the outliers problem, \cite{he2010maximum} propose to learn a robust sparse representation for face recognition based on the correntropy \citep{liu2007correntropy} along with the use of an $\ell_1$ norm penalty, which is insensitive to outliers. \cite{wu2007robust} propose a robust truncated hinge loss for SVM. \cite{yu2010relaxed} propose a robust estimation method for classification based on loss clipping.  All of them are trying to design non-convex individual losses for robust learning.

Rank-based aggregate losses are oftentimes overlooked in existing machine learning literature. One relevant topic is the data subset selection problem \citep{wei2015submodularity}, which is about selecting a subset from a large training dataset to train a model while incurring a minimal average loss, the learning objective of which can be regarded as a special aggregate loss that averages over the individual losses corresponding to data selected into the subset.
Hard example mining is an effective method in the training and is widely used in existing works \citep{gidaris2015object, liu2016ssd, shrivastava2016training, lin2017focal}. For instance, in online hard example mining, \cite{shrivastava2016training} use top-$k$ samples with the largest losses for each training image to update the model parameter. In object detection, \cite{lin2017focal} propose a weighted cross-entropy loss, which assigns more weights on candidate bounding box with large losses. They obtain a better performance with this method than using the conventional cross-entropy loss. However, it should be mentioned that the connections of these works to the aggregate loss are incidental. The most relevant works on the rank-based aggregate loss are \cite{fan2017learning} and \cite{lyu2020average}. In their works, Fan et al. and Lyu et al. propose an average top-$k$ loss as a type of aggregate loss. They also mention that the maximum loss \citep{shalev2016minimizing} and the average loss \citep{vapnik1992principles} are two special cases of their average top-$k$ loss. They demonstrate that their loss can alleviate the influence of data with imbalance and outlier problems in the learning process. However, as we mentioned before, their method can dilute but not completely eliminate the influence of the outliers.

Rank-based individual losses have been studied in many tasks. For example, multi-label/multi-class learning, ranking, and information retrieval. The work from \cite{rudin2009p, usunier2009ranking} propose a form of individual loss that gives more weights to the samples at the top of a ranked list. This idea is further extended to multi-label learning in the work of \cite{weston2011wsabie}. Top-1 loss is commonly used in multi-class learning, which causes more penalties when the 
class corresponding to the top-1 prediction score
inconsistent with the ground-truth class label \citep{crammer2001algorithmic}. Since a sample may contain multiple classes and some classes may overlap, \cite{lapin2015top, lapin2016loss, lapin2017analysis} propose top-$k$ multi-class loss that can introduce penalties when 
the ground-truth label does not appear in the set of top-$k$ labels as measured by their prediction scores.
In multi-label learning, many works focus on the ranking-based approaches, such as \cite{zhang2006multilabel, fan2020multi}. The loss functions from their methods encourage  the predicted relevancy scores of the ground-truth positive labels to be higher than that of the negative ones.  Similar to the motivation of the top-$k$ multi-class classification, in multi-label learning, the classifier is expected to include as many true labels as possible in the top $k$ outputs. However, there are no dedicated methods for the top-$k$ multi-label learning. Therefore, in this paper, we propose \tkml~individual loss for multi-label learning to fill this gap.

Rank-based losses are also popular to be used at the sample and label levels simultaneously. For example, the work of \cite{rawat2020doubly} proposed a doubly-stochastic mining method, which combines the methods of \cite{kawaguchiordered} and \cite{lapin2015top}. They use the average top-$k$ methods both at the data sample and label levels to construct the final loss function. Then the constructed loss is applied to solve modern retrieval problems, which are characterized by training sets with a huge number of labels, and heterogeneous data distributions across sub-populations. However, their method does not consider the outliers or noisy labels that may exist in real-world data sets.  Such problems can make the performance of the proposed model plummet. In addition, their model only works on multi-class problems and cannot be applied to multi-label learning directly. In this paper, we fill this gap and propose a \tkml-\aorr~loss for solving the noise problem in top-$k$ multi-label learning. Furthermore, we  design a heuristic algorithm for optimizing \tkml-\aorr~loss and show it can be generalized to other existing algorithms in  \cite{shen2019learning, shah2020choosing, kawaguchiordered, rawat2020doubly}.


\section{Sum of Ranked Range (\sorr)}
\label{sec:dca}

\begin{table*}[]
\centering
\scalebox{0.8} { 
\begin{tabular}{|c|c|}
\hline
Symbol & Description  \\ \hline
$S=\{s_1,\cdots,s_n\}$ & A set of real numbers   \\ \hline
$s_{[k]}$ ($s_{[m]}$) & The $k$-th ($m$-th) largest value after sorting the elements in $S$ \\ \hline
$s_{i}(\cdot)$ & The $i$-th individual loss \\ \hline
$\phi_{k}(\cdot)$ ($\phi_{m}(\cdot)$) & The sum of top-$k$ (top-$m$) values \\ \hline
$\psi_{m,k}(\cdot)$ & The sum of ($m$, $k$)-ranked range values \\ \hline
$q_i$ & The weight of $s_{i}(\cdot)$ \\ \hline
$x$ & A multi-dimensional data instance\\ \hline
$y$ & A label (class) of $x$ \\ \hline
$f_{\theta}(\cdot)$, $f(\cdot;\theta)$ & A parametric function (classifier) with parameters $\theta$ \\ \hline
$\mathcal{L}(\cdot)$ & Loss function \\ \hline
$\mathcal{D}$ & Training data set  \\ \hline
$\widetilde{\mathcal{D}}$ & Validation data set \\ \hline
$\mathcal{Y}=\{1,\cdots,l\}$ & A set of labels (classes) with size $l$   \\ \hline
$Y$ & Ground truth labels of $x$ and it is a subset of $\mathcal{Y}$ \\ \hline
\end{tabular}
}
\caption{Frequently used symbols in this paper.}
\label{tab:symbols}
\end{table*}

For a set of real numbers $S = \{s_1, \cdots, s_n\}$, we use $s_{[k]}$ to denote the {\em top-$k$ value}, which is the $k$-th largest value after sorting the elements in $S$ (ties can be broken in any consistent way). Correspondingly, we define $\phi_k(S) = \sum_{i=1}^k s_{[i]}$ as the {\em sum of the top-$k$} values of $S$. 
For two integers $k$ and $m$, $0 \le m < k \le n$, the $(m,k)$-ranked range is the set of sorted values $\{s_{[m+1]},\cdots,s_{[k]}\}$. The sum of $(m,k)$-ranked range ($(m,k)$-\sorr) is defined as $\psi_{m,k}(S) =\sum_{i=m+1}^k s_{[i]}$, and the average of $(m,k)$-ranked range ($(m,k)$-\aorr) is ${1 \over k-m}\psi_{m,k}(S)$. It is easy to see that the sum of ranked range (\sorr) is the difference between two sums of top values as, $\psi_{m,k}(S) =\phi_k(S) - \phi_m(S)$. Also, the top-$k$ value corresponds to the $(k-1,k)$-\sorr, as  $\psi_{k-1,k}(S)=s_{[k]}$. Similarly, the median can also be obtained from \aorr, as $\frac{1}{\lceil \frac{n+1}{2} \rceil - \lfloor \frac{n+1}{2}\rfloor +1 }\psi_{\lfloor \frac{n+1}{2}\rfloor-1, \lceil \frac{n+1}{2} \rceil}(S)$. We collect symbols and notations to be used throughout the paper in Table \ref{tab:symbols}.

In machine learning problems, we are interested in the set $S(\theta) = \{s_1(\theta),\cdots,s_n(\theta)\}$ formed from a family of functions where each $s_i$ is a convex function of parameter $\theta$. For example, in practice, $s$ can be a convex loss function. $n$ is the total number of training samples. $s_i$ is the individual loss function of sample $i$. We can use \sorr~to form learning objectives. In particular, we can eliminate the ranking operation and use the equivalent form of \sorr~in the following result. Denote $[a]_+=\max\{0,a\}$ as the hinge function. 
\vspace*{-1mm}
\begin{theorem}\label{theorem:sorr} Suppose $s_i(\theta)$ is convex with respect to $\theta$ for any $i\in [1,n]$, then 
\begin{equation}
\min_{\theta}\psi_{m,k}(S(\theta)) = \min_{\theta}\bigg[ \min_{\lambda\in \mathbb{R}}\Big\{k\lambda+\sum_{i=1}^n[s_i(\theta)-\lambda]_+\Big\}-\min_{\hat{\lambda}\in \mathbb{R}}\Big\{m\hat{\lambda}+\sum_{i=1}^n[s_i(\theta)-\hat{\lambda}]_+\Big\}\bigg].
\label{eq:0}    
\end{equation}
Furthermore, $\hat{\lambda}>\lambda$, when the optimal solution is achieved. 
\end{theorem} 
The proof of Theorem \ref{theorem:sorr} is in Appendix \ref{proof_theorem_sorr}. Note that $\psi_{m,k}(S(\theta))$ is not a convex function of $\theta$. But its equivalence to the difference between $\phi_k(S(\theta))$ and $\phi_m(S(\theta))$ suggests that $\psi_{m,k}(S(\theta))$ is a difference-of-convex (DC) function, because $\phi_k(S(\theta))$ and $\phi_m(S(\theta))$ are convex functions of $\theta$ in this setting. As such, a natural choice for its optimization is the DC algorithm (DCA) \citep{tao1986algorithms}. 

To be specific, for a general DC problem formed from two convex functions $g(\theta), h(\theta)$ as $\overline{s}(\theta)=g(\theta)- h(\theta)$, DCA iteratively search for a critical point of $\overline{s}(\theta)$ \citep{thi2017stochastic}. At each iteration of DCA, we form an affine majorization of function $h$ using  its sub-gradient at $\theta^{(t)}$, \ie, $\hat{\theta}^{(t)} \in \partial h(\theta^{(t)})$, and then update $\theta^{(t+1)} \in \arg\!\min_\theta \left \{ g(\theta) -\theta^\top \hat{\theta}^{ (t)} \right \}$. DCA is a descent method without line search, which means the objective function is monotonically decreased at each iteration \citep{tao1997convex}. It does not require the differentiability of $g(\theta)$ and $h(\theta)$ to assure its convergence. Moreover, it is known that DCA converges from an arbitrary initial point and often converges to a global solution \citep{le2018dc}. For example, the authors in \cite{tao1997convex} proved that $\theta^*$ is a critical point of $\overline{s}(\theta)$ if $\partial g(\theta^*)\cap \partial h(\theta^*)\neq \oldemptyset$, or equivalently, $0\in \partial g(\theta^*)- \partial h(\theta^*)$. A critical point can lead to a local minimizer of $\overline{s}(\theta)$ if it admits a neighborhood $U$ such that $\partial h(\theta)\cap \partial g(\theta^*)\neq \oldemptyset$, $\forall \theta \in U\cap$ dom $g$. While a DC problem can be solved based on standard (sub-)gradient descent methods, DCA seems to be more amenable to our task because of  its appealing properties  and the natural DC structure of our objective function. In addition, as shown in  \cite{piot2016difference} with extensive experiments,  DCA empirically outperforms the gradient descent method on various problems.


\begin{algorithm}[t]
    \caption{DCA for Minimizing \sorr}\label{Alg0}
    \SetAlgoLined
    \textbf{Initialization:} $\theta^{(0)}$, $\lambda^{(0)}$, $\eta_l$, and two hyperparameters $k$ and $m$ \\

    
    \For{$t=0,1,...$}{
    Compute $\hat{\theta}^{(t)}$ with equation (\ref{eq:subgradient_phi_m})

    \For{$l=0,1,...$}{

    Compute $\theta^{(l+1)}$ and $\lambda^{(l+1)}$ with equation (\ref{eq:SGD})
    }
    
    Update $\theta^{(t+1)} \leftarrow \theta^{(l+1)}$ 

    }
    
\end{algorithm}

To use DCA to optimize \sorr, we need to solve the convex sub-optimization problem 
\[
\min_{\theta}\bigg[\min_{\lambda}\Big\{k\lambda+\sum_{i=1}^n[s_i(\theta)-\lambda]_+\Big\}- \theta^T\hat{\theta} \bigg].
\]
This problem can be solved using a stochastic sub-gradient method \citep{bottou2008tradeoffs, rakhlin2011making, srebro2010stochastic}. We first randomly sample $s_{i_l}(\theta^{(l)})$ from the collection of $\{s_i(\theta^{(l)})\}_{i=1}^n$ and then perform the following steps:
\begin{equation}\small
    \begin{aligned}
        \theta^{(l+1)} \leftarrow \theta^{(l)}-\eta_l\left(\partial s_{i_l}(\theta^{(l)})\cdot \mathbb{I}_{[s_{i_l}(\theta^{(l)})>\lambda^{(l)}]} - \hat{\theta}^{ (t)}\right), \ \
      \lambda^{(l+1)}\leftarrow \lambda^{(l)}- \eta_l\left(k - \mathbb{I}_{[s_{i_l}(\theta^{(l)})>\lambda^{(l)}]}\right),
    \end{aligned}
\label{eq:SGD}
\end{equation}
where $\eta_l$ is the step size. $\mathbb{I}_{[a]}$ is an indicator function with $\mathbb{I}_{[a]}=1$ if $a$ is true and 0 otherwise. In equation \eqref{eq:SGD}, we use the fact that the sub-gradient of $\phi_m(S(\theta))$ is computed, as
\begin{equation}
\hat{\theta} \in \partial \phi_m(S(\theta)) = \sum_{i=1}^n \partial s_i(\theta)\cdot \mathbb{I}_{[s_i(\theta)>s_{[m]}(\theta)]},
\label{eq:subgradient_phi_m} 
\end{equation}
where $\partial s_i(\theta)$ is the gradient or a sub-gradient of convex function $s_i(\theta)$ (Proof can be found in Appendix \ref{prooftheorem1})\footnote{For large data sets, we can use a stochastic
version of DCA, which is more efficient with a provable convergence to a critical point \citep{thi2019stochastic}.}.
The pseudo-code of minimizing \sorr~is described in Algorithm \ref{Alg0}.
For a given outer loop size $|t|$, an inner loop size $|l|$, and training sample size $n$, the time complexity of Algorithm \ref{Alg0} is $O(|t|(n\log n+|l|))$.

To better understand \sorr, we provide more discussion about it. Since \sorr~is non-convex, an intuitive strategy to optimize a non-convex loss function is to use a surrogate convex loss to replace it and apply existing convex optimization approaches to solve the proposed surrogate convex loss. However, there are two drawbacks to  this strategy. First, a significant learning property of the surrogate loss is its consistency, which requires the optimal minimizers of the surrogate loss to be near to or exact the optimal minimizers of the original loss. As we shall see, the study of consistency is significantly more complex for \sorr. Fortunately, we rewrite \sorr~to a DC problem (equation (\ref{eq:0})), and then we can show \aorr~aggregate loss (a variant of \sorr~only add an additional factor) satisfies the classification calibration (a sufficient condition for consistency) under several moderate conditions in Section 4.1. Second, it is very hard to find a suitable convex loss to surrogate \sorr. If we can find a surrogate convex loss, the optimal solution obtained by learning this convex loss and the actual optimal solution based on \sorr~are not guaranteed to be equal. This difference comes not only from the optimization algorithm, but also from the learning objective. For example, we may select the sum of the top-$k$ loss $\phi_k(S)$ to surrogate $\psi_{m,k}(S)$ since it is a convex loss. But $\phi_k(S)$ will use extremely largest values (top-1 value to top-$m$ value) to train the model, while $\psi_{m,k}(S)$ is not. Therefore, they have different meanings in model learning. In other words, the learned models are exactly different. The model learned by using $\phi_k(S)$ may not satisfy the original requirements in specific tasks such as robust to outliers. However, in equation (\ref{eq:0}), the DC term (right-hand side) is exactly equivalent to the original \sorr~formula (left-hand side). Since the DC problem is well studied in the existing works and can be solved by DCA with provable convergence properties,  it is beneficial and natural to use the DC function to replace the original ranking formula even if the DC function is also non-convex.


\subsection{Connection with Bilevel Optimization} \label{sec:bilevel}

We give another intuitive interpretation of \sorr. Suppose we have 10 individual losses in the ranked list. To extract the sum of (2, 6)-ranked range, we can select a subset, which contains the bottom 8 individual losses from the ranked list in the beginning. Then we select the top 4 individual losses from this subset as the finalized (2, 6)-ranked range. Suppose $s_i(\theta)\geq 0$, based on this new intuition, we can reformulate \sorr. First, we sum the bottom $n-m$ losses as follows,
\begin{equation*}
    \begin{aligned}
    \sum_{i=m+1}^n s_{[i]}(\theta) = \min_{q} \sum_{i=1}^n q_i s_i(\theta) \ \ \ \text{s.t.} \ q_i\in\{0,1\}, \ ||q||_0 =n-m,
    \end{aligned}
\end{equation*}
where $q=\{q_1,\cdots,q_n\}\in \{0,1\}^n$, and $q_i$ is an indicator. When $q_i=0$, it indicates that the $i$-th individual loss is not included in the objective function. Otherwise, the objective function should include this individual loss. Next, we sum the top-($k-m$) individual losses from the bottom $n-m$ individual losses as follows,
\begin{equation}
    \begin{aligned}
    &\min_{q}\sum_{i=1}^{k-m} (q s(\theta))_{[i]} \ \ \ \text{s.t.} \ q_i\in\{0,1\}, \ ||q||_0 =n-m \\
    =&\min_{\lambda,q} (k-m)\lambda+\sum_{i=1}^n [q_is_i(\theta)-\lambda]_+\ \ \ \text{s.t.} \ q_i\in\{0,1\}, \ ||q||_0 =n-m\\
    =&\min_{\lambda,q}(k-m)\lambda+\sum_{i=1}^n q_i[s_i(\theta)-\lambda]_+\ \ \ \text{s.t.} \ q_i\in [0,1], \ ||q||_0 =n-m,
    \end{aligned}
\label{eq:aorr_newform}
\end{equation}
where $qs(\theta)=\{q_1s_1(\theta),\cdots,q_ns_n(\theta)\}$. The first equation holds because of Lemma \ref{lemma:convex}. Since $q_is_i(\theta)\geq 0$,  we know the optimal $\lambda^*\geq 0$ from Lemma \ref{lemma:convex}. If $q_i=0$, $[q_is_i(\theta)-\lambda^*]_+=0=q_i[s_i(\theta)-\lambda^*]_+$. If $q_i=1$, $[q_is_i(\theta)-\lambda^*]_+=[s_i(\theta)-\lambda^*]_+=q_i[s_i(\theta)-\lambda^*]_+$. Thus the second equation holds. It should be mentioned that the discrete indicator $q_i$ can be replaced by a continue one, which means $q_i\in [0,1]$. The reason is that the optimal $q$ is at a corner of the hypercube $[0,1]^n$. In fact, with the outer minimization of model parameter $\theta$, the above objective function is equivalent to equation (\ref{eq:0}). Specifically, we obtain a theorem as follows,
\begin{theorem}
With relaxing $q$ to $[0,1]$, equation (\ref{eq:aorr_newform}) is equivalent to  equation (\ref{eq:0}):
\begin{equation*}
    \begin{aligned}
    &\min_{\lambda,q}(k-m)\lambda+\sum_{i=1}^n q_i[s_i(\theta)-\lambda]_+\ \ \ \text{s.t.} \ q_i\in [0,1], \ ||q||_0 =n-m \\
    =&\min_{\lambda}\Big\{k\lambda+\sum_{i=1}^n[s_i(\theta)-\lambda]_+\Big\}-\min_{\hat{\lambda}}\Big\{m\hat{\lambda}+\sum_{i=1}^n[s_i(\theta)-\hat{\lambda}]_+\Big\}.
    \end{aligned}
\label{eq:aorr-equal}
\end{equation*}
\label{theorem:aorr-equal}
\end{theorem}
Proof can be found in Appendix \ref{appendix:proof_theorem_aorr_equal}. Combining the optimization of the model parameter $\theta$, we can reform equation (\ref{eq:aorr_newform}) to a bilevel optimization problem \citep{borsos2020coresets, jenni2018deep} as follows,
\begin{equation}
    \begin{aligned}
    \min_{\lambda, q} \ \ \ & (k-m)\lambda+\sum_{i=1}^n q_i[s_i(\theta^*(q))-\lambda]_+\\
    \text{s.t.} \ \ \ & \theta^*(q)\in \argmin_{\theta} (k-m)\lambda+\sum_{i=1}^n q_i[s_i(\theta)-\lambda]_+\\
    & q_i\in [0,1], \ ||q||_0 =n-m,
    \end{aligned}
\label{eq:bilevel}
\end{equation}
where we minimize an \textit{outer} objective, here the first row of equation (\ref{eq:bilevel}), which in turn depends on the solution $\theta^*(q)$ to an \textit{inner} optimization problem (the second row of equation (\ref{eq:bilevel})) with the constraint on $q$. This problem (equation (\ref{eq:bilevel})) can be solved through some existing bilevel optimization algorithms \citep{borsos2020coresets}. For example, we can adopt the \textit{coresets via bilevel optimization} algorithm from \cite{borsos2020coresets} to solve our cardinality-constrained bilevel optimization problem. We only need to take our specific inner problem, outer problem, and constraints into their optimization framework.

\section{\aorr~Aggregate Loss for Binary and Multi-class Classification} \label{sec:aorr}

\sorr~provides a general framework to aggregate individual values to form a learning objective. Here, we examine its use as an aggregate loss in supervised learning problems in detail and optimize it using the DC algorithm.
Specifically, we aim to find a parametric function $f_\theta$ with parameter $\theta$ that can predict a target $y$ from the input data or features $x$ using a set of labeled training samples $\{(x_i,y_i)\}_{i=1}^n$.  We assume that the individual loss for a sample $(x_i,y_i)$ is $s_i(\theta) = s(f(x_i;\theta),y_i) \ge 0$. The learning objective for supervised learning problem is constructed from the aggregate loss $\mathcal{L}(S(\theta))$ that accumulates all individual losses over training samples, $S(\theta)=\{s_i(\theta)\}_{i=1}^n$. Specifically, we define the \aorr~aggregate loss as
\[
\L_{aorr}(S(\theta)) = {1 \over k-m} \psi_{m,k}(S(\theta)) = {1 \over k-m} \sum_{i=m+1}^{k} s_{[i]}(\theta).
\]
If we choose the $\ell_2$ individual loss or the hinge individual loss, we get the learning objectives in \cite{ortis2019predicting} and \cite{kanamori2017robustness}, respectively. For $m\ge 1$, we can optimize \aorr~using the DCA as described in Section \ref{sec:dca}.

The \aorr~aggregate loss is related to previous aggregate losses that are widely used to form learning objectives.
\begin{compactitem}
    \item the {\em average loss} \citep{vapnik2013nature}:   $\L_{avg}(S(\theta))=\frac{1}{n}\sum_{i=1}^n s_i(\theta)$;
    \item the {\em maximum loss} \citep{shalev2016minimizing}: $\L_{max}(S(\theta)) = \mbox{max}_{1\leq i \leq n}s_i(\theta)$;
    \item the {\em median loss} \citep{ma2011robust}:  $\L_{med}(S(\theta)) = {1 \over 2}
    \left(s_{\left[\lfloor \frac{n+1}{2}\rfloor\right]}(\theta) + s_{\left[\lceil \frac{n+1}{2} \rceil\right]}(\theta)\right)$;
    \item the {\em average top-$k$ loss} (AT$_k$) \citep{fan2017learning}: $\L_{avt-k}(S(\theta)) = \frac{1}{k} \sum_{i=1}^k s_{[i]}(\theta)$, for $1\leq k \leq n$.
\end{compactitem}
The \aorr~aggregate loss generalizes the average loss ($k=n$ and $m=0$), the maximum loss ($k=1$ and $m=0$), the median loss ($k=\lceil \frac{n+1}{2} \rceil$, $m=\lfloor \frac{n+1}{2}\rfloor-1$) , and the average top-$k$ loss ($m=0$). Interestingly, the average of the bottom-$(n-m)$ loss, $\L_{abt-m}(S(\theta)) = \frac{1}{n-m} \sum_{i=m+1}^n s_{[i]}(\theta)$, which is not widely studied in the literature as a learning objective, is an instance of the \aorr~aggregate loss ($k=n$). {In addition, the robust version of the maximum loss \citep{shalev2016minimizing}, which is a maximum loss on a subset of samples of size at least $n-(k-1)$, where the number of outliers is at most $k-1$, is equivalent to the top-$k$ loss, a special case of the \aorr~ aggregate loss ($m=k-1$).}  

\begin{figure}
\centering
    \includegraphics[trim=1 1 1 1, clip,keepaspectratio, width=0.8\textwidth]{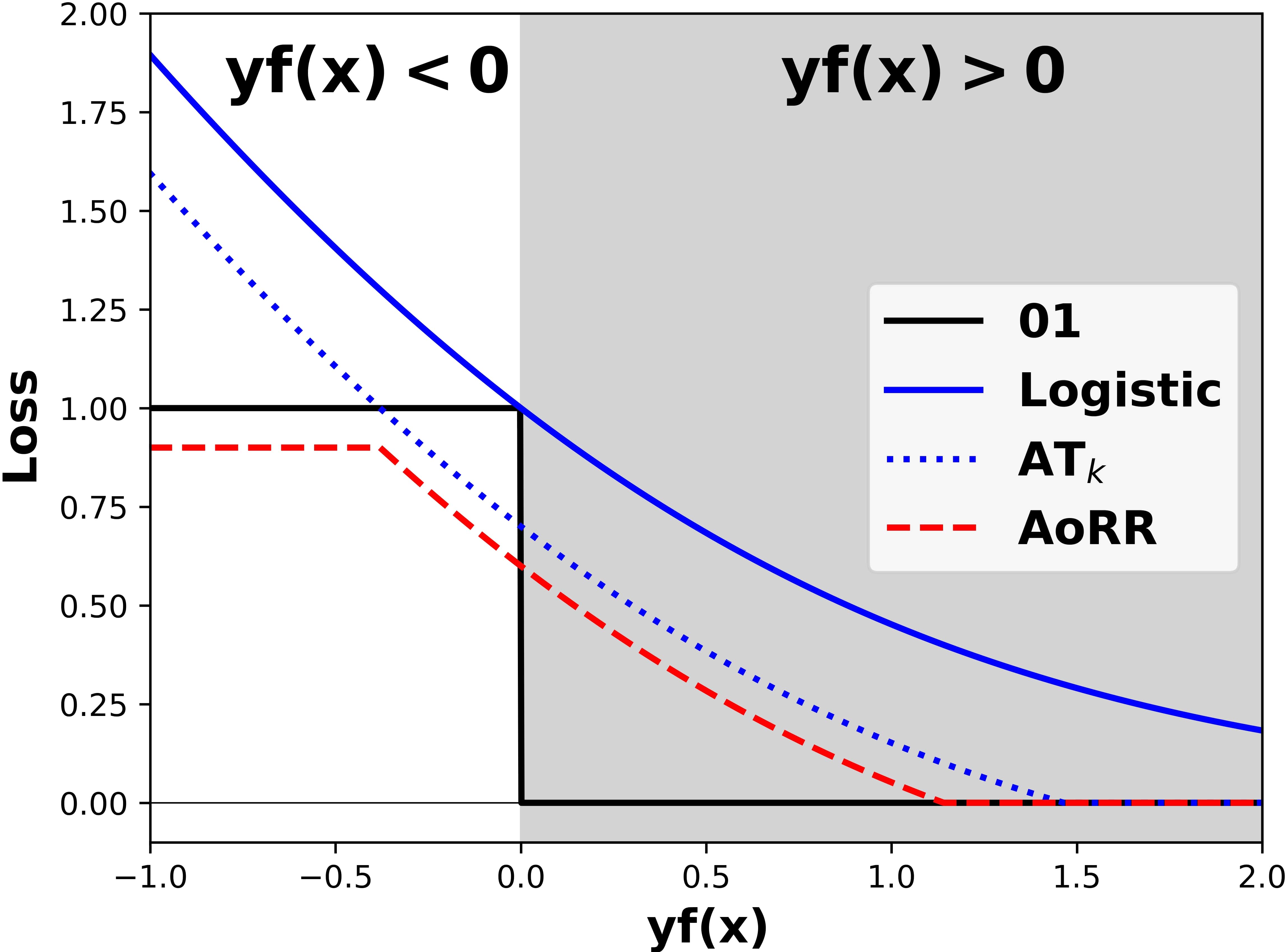}
  \caption{ The \aorr~loss and other losses interpreted at
the individual sample level. The shaded area over 0.00 loss
corresponds to data/target with the correct
classification. Note that we fix $\lambda=0.4$ and $\hat{\lambda}=1.3$ to draw the \aorr~curve.}
\label{fig:interpretation}
\end{figure}

Using the \aorr~aggregate loss can bring flexibility in designing learning objectives and alleviate drawbacks of previous aggregate losses. In particular, the average loss, the maximum loss, and the AT$_k$ loss are all influenced by outliers in training data since outliers usually correspond to extremely large individual losses and all of these methods inevitably use large individual losses to construct aggregate losses.
They only differ in the degree of influence, with the maximum loss being the most sensitive to outliers. In comparison, \aorr~loss can completely eliminate the influence of the top individual losses by excluding the top $m$ individual losses from the learning objective.

In addition, traditional approaches to handling outliers focus on the design of robust {\em individual losses} over training samples, notable examples include the Huber loss \citep{ friedman2001elements} and the capped hinge loss \citep{nie2017multiclass}. Changing individual losses may not be desirable, as they are usually relevant to the learning problem and application. On the other hand, 
instead of changing the definition of individual losses, our method builds in robustness at the aggregate loss level, using the \aorr~aggregate loss. The resulting learning algorithm based on Algorithm \ref{Alg0} is more flexible and allows the user to choose an individual loss form that is relevant to the learning problem.

The robustness to outliers of the \aorr~loss can be more clearly understood at the individual sample level, with fixed $\gl$ and $\hat{\gl}$. We use binary classification to illustrate with $s_i(\theta) = s(y_if_\theta(x_i))$ where $f_\theta$ is the parametric predictor and $y_i\in \{\pm 1\}.$ In this case,  $y_if_\theta(x_i)>0$ and $y_if_\theta(x_i)<0$ corresponds to the correct and false predictions, respectively. Specifically, noting that $s_i(\theta)\geq 0$, we can rearrange terms in equation \eqref{eq:0} to obtain
\begin{equation}
\L_{aorr}(S(\theta)) = {1 \over k-m}\min_{\lambda > 0}\max_{\hat{\lambda} > \lambda} \sum_{i=1}^n \Big\{[s_i(\theta)-\lambda]_+ -[s_i(\theta)-\hat{\lambda}]_+ \Big\}+ k\lambda - m\hat{\lambda}.
\label{eq:aorr-loss}
\end{equation}
We are particularly interested in the term inside the summation in equation \eqref{eq:aorr-loss} 
\[
[s(yf_\theta(x))-\lambda]_+ -[s(yf_\theta(x))-\hat{\lambda}]_+ = \left\{
\begin{array}{cc}
     \hat{\lambda}-\lambda & s(yf_\theta(x)) > \hat{\lambda} \\
    s(yf_\theta(x))-\lambda &  \lambda < s(yf_\theta(x)) \le \hat{\lambda} \\
    0 & s(yf_\theta(x)) \le \lambda
\end{array}
\right..
\]
According to this, at the level of individual training samples, the equivalent effect of using the \aorr~loss is to uniformly reduce the individual losses by $\lambda$, but truncate the reduced individual loss at values below zero or above $\hat{\lambda}$. The situation is illustrated in Figure \ref{fig:interpretation} for the logistic individual  loss $s(yf(x)) = \log_2(1+e^{-yf(x)})$, which is a convex and smooth surrogate to the ideal $01$-loss. The effect of reducing and truncating from below and above has two interesting consequences. First, note that the use of convex and smooth surrogate loss inevitably introduces penalties to samples that are correctly classified but are ``too close'' to the boundary. The reduction of the individual loss alleviates that improper penalty. This property is also shared by the AT$_k$ loss. On the other hand, the ideal $01$-loss exerts the same penalty to all incorrect classified samples regardless of their margin value, while the surrogate has unbounded penalties. This is the exact cause of the sensitivity to outliers of the previous aggregate losses, but the truncation of \aorr~loss is similar to the $01$-loss, and thus is more robust to the outliers. It is worth emphasizing that the above explanation of the \aorr~ loss has been illustrated at the individual sample level with fixed $\gl$ and $\hat{\gl}. $ The aggregate \aorr~loss defined  by equation \eqref{eq:aorr-loss} as  a whole is not an average sample-based loss because it can not be decomposed into the summation of individual losses over samples.

\subsection{Classification Calibration} 

A fundamental question in learning theory for classification \citep{bartlett2006convexity,vapnik2013nature} is to investigate when the best possible estimator from a learning objective is consistent with the best possible, \ie, the Bayes rule.  Here we investigate this statistical question for the \aorr~loss by considering its infinite sample case, \ie, $n\to \infty.$ {As mentioned above, the \aorr~loss as a whole is not the average of individual losses over samples, and therefore the analysis for the standard ERM \citep{bartlett2006convexity,lin2004note}  does not apply to our case. }

We assume that the training data $\{(x_i,y_i)\}_{i=1}^n$ are i.i.d. from an unknown distribution $p$ on $\X \times \{\pm 1\}$.  The misclassification error measures the quality of a classifier $f:\X \to \{\pm 1\}$ and is denoted by $\cR(f) = \Pr( Y \neq f(X))=  \EX[\mbI_{Yf(X)\le 0}]$. The Bayes error leads to the least expected error, which is defined by $\cR^\ast = \inf_{f} \cR(f)$. 
No function can achieve the Bayes error than the Bayes rule
$f_c(x)=\sgn(\eta(x) - {1 \over 2})$, where $\eta(x) = P(Y=1|X=x).$ It is well noted that, in practice, one uses a surrogate loss $\ell: \R \to [0,\infty)$ which is a continuous function and upper-bounds the $01$-loss. Its true risk is given by $\cE_\ell(f) =  \EX[\ell(Yf(X))]$.  Denote the optimal $\ell$-risk by $\cE^\ast_\ell = \inf_f \cE_\ell(f)$, the {\em classification calibration} (point-wise form of Fisher consistency) for loss $\ell$ \citep{bartlett2006convexity,lin2004note} holds true if the minimizer $f^\ast_\ell = \inf_{f}\cE_\ell(f)$ has  the same sign as the Bayes rule $f_c(x)$, \ie, $\sgn(f^\ast_\ell(x)) = \sgn(f_c(x))$ whenever $f_c(x) \neq 0$.

In analogy, we can investigate the classification calibration property of the \aorr~loss. Specifically, we first obtain the population form of the \aorr~loss using the infinite limit of the empirical one given by equation \eqref{eq:aorr-loss}. Indeed, 
we know from \citep{bhat2019concentration,brown2007large} that, for any bounded $f$ and $\alpha\in (0,1]$, there holds $ \inf_{\lambda\ge 0}\alpha\lambda+\frac{1}{n}\sum_{i=1}^n[s(y_if_\theta(x_i))-\lambda]_+  \to \inf_{\lambda\ge 0}\alpha\lambda+\EX[s(Yf(X))-\lambda]_+$  as $n\to \infty.$ Consequently, we have the limit case of the \aorr~loss $ \L_{aorr}(S(\theta)) $  restated as follows:
\begin{equation}\small
\begin{aligned}
&\frac{n}{k-m}\bigg[\min_{\lambda}\Big\{\frac{k}{n}\lambda+\frac{1}{n}\sum_{i=1}^n[s(y_if_\theta(x_i))-\lambda]_+\Big\}-\min_{\hat{\lambda}}\Big\{\frac{m}{n}\hat{\lambda}+\frac{1}{n}\sum_{i=1}^n[s(y_if_\theta(x_i))-\hat{\lambda}]_+\Big\}\bigg]\\
&\xrightarrow[n\rightarrow \infty]{\frac{k}{n}\rightarrow \nu, \frac{m}{n} \rightarrow \mu} \frac{n}{k-m}\bigg[\min_{\lambda \ge 0}\Big\{\mathbb{E}[[s(Yf(X))-\lambda]_+]+\nu\lambda\Big\}-\min_{\hat{\lambda} \ge 0}\Big\{\mathbb{E}[[s(Yf(X))-\hat{\lambda}]_+]+\mu\hat{\lambda}\Big\}\bigg].
\end{aligned}
\end{equation}
Throughout the paper, we assume that $\nu>\mu$ which is reasonable as $k>m.$ In particular, we assume that $\mu>0$ since if $\mu=0$ then it will lead to $\hat{\lambda} = \infty$ and this case is reduced to the population version of the average top-$k$ case in \cite{fan2017learning}.  As such, the population version of our \aorr~loss (equation \eqref{eq:aorr-loss}) is given by 
\begin{equation}
\begin{aligned}
(f^*_0, \lambda^*, \hat{\lambda}^*) = \mbox{arg}\ \underset{f, \lambda\geq 0}{\mbox{inf}} \underset{\hat{\lambda}\geq 0}{\mbox{sup}} \left\{\mathbb{E}[[s(Yf(X))-\lambda]_+ - [s(Yf(X))-\hat{\lambda}]_+]+(\nu\lambda-\mu \hat{\lambda})\right\}.
\end{aligned}
\label{eq:opt_ar_loss}
\end{equation}
It is difficult to directly work on the optima $f^*_0$ since the problem in equation \eqref{eq:opt_ar_loss} is a non-convex min-max problem and the standard min-max theorem does not apply here. Instead, we assume the existence of $\gl^*$ and $ \hgl^*$ in equation \eqref{eq:opt_ar_loss} and work with the minimizer $f^* = \arg\inf_{f} \L(f, \gl^*,\hgl^*)$ where $ \L(f, \gl^*,\hgl^*): = \mathbb{E}[[s(Yf(X))-\lambda^*]_+ - [s(Yf(X))-\hat{\lambda}^*]_+]+(\nu\lambda^*-\mu \hat{\lambda}^*).$ Now we can define the classification calibration for the \aorr~loss. 

\begin{definition} The \aorr~loss is called classification calibrated if, for any $x$, there is a minimizer $f^*=\arg\inf_{f} \L(f,\gl^*,\hgl^*)$ such as $f^*(x) >0$ if $\eta(x)>1/2$ and $f^*(x)<0$ if $\eta(x)<1/2.$
\end{definition}
We can then obtain the following theorem. Its proof can be found in Appendix \ref{appendix:proof_theorem_2}.  
\begin{theorem}
Suppose the individual loss $s: \mathbb{R} \rightarrow \mathbb{R}^+$ is  non-increasing, convex, differentiable at 0 and $s^{\prime}(0)<0.$  If  $0\leq \lambda^*< \hat{\lambda}^*$, then the \aorr~loss is classification calibrated.
\label{theorem:AR_calibration}
\end{theorem}

The assumptions in Theorem \ref{theorem:AR_calibration} are easily to be satisfied. Indeed, the commonly used individual losses such as the hinge loss $s(t)=[1-t]_+$ and the logistic loss $s(t)=\log_2(1+e^{-t})$ satisfy the conditions of non-increasing, convex, differentiable at 0, and $s^{\prime}(0)<0$. Furthermore, according to Theorem \ref{theorem:sorr} and $0\leq m< k\leq n$, we obtain $\lambda^*<\hat{\lambda}^*$.
Suppose the individual loss $s(t)$ is non-negative (this holds for the above-mentioned individual losses), we have $0\leq\lambda^*<\hat{\lambda}^*$. We also provide several examples that cannot make the assumptions hold. For example, the least square loss $s(t)=(1-t)^2$ is not a non-increasing individual loss. Therefore, using it as an individual loss cannot guarantee the \aorr~aggregate loss satisfies classification calibration. In addition, if $k\leq m$, the hypothesis $0\leq \lambda^*< \hat{\lambda}^*$ will not be held. So we also cannot obtain a classification calibrated \aorr~aggregate loss.

\subsection{Connection with Conditional Value at Risk } \label{sec:cvar}


\begin{figure}
\centering
    \includegraphics[trim=1 1 1 1, clip,keepaspectratio, width=0.6\textwidth]{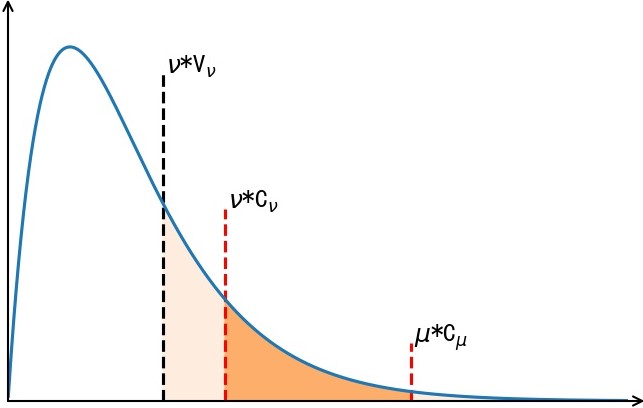}
  \caption{The yellow area between two red dash lines represents ICVaRs}
\label{fig:interpretation_CVaR}
\end{figure}

It is worthy of mentioning that the average of the top-$k$ method is also related to the risk measure called conditional value at risk (CVaR) at level $\alpha  = {k\over n }$ \citep[Chapter~6]{ shapiro2014lectures} in portfolio optimization for effective risk management. 

Let ($\Omega, \Sigma, P$) be a probability space. The \textit{conditional value at risk} (CVaR) at level $\alpha \in (0,1)$ of a random variable $s: \Omega\rightarrow \mathbb{R}$ is defined as $C_{\alpha}[s]:=\inf_{\lambda}\{\lambda+\frac{1}{\alpha}\mathbb{E}[[s-\lambda]_+]\}$. If $s$ is a continuous random variable, then $C_{\alpha}[s]=\mathbb{E}[s|s\geq V_{\alpha}[s]]$, where $V_{\alpha}[s]$ is called the \textit{value at risk} (VaR) and defined as $V_{\alpha}[s]:=\sup\{\lambda\in \mathbb{R}|\Pr(s\geq \lambda)\geq \alpha\}$. A sample-based estimate of $C_{\alpha}[s]$ can be denoted by $\widehat{C}_{\alpha}[s]:= \inf_{\lambda\in \mathbb{R}}\{\lambda+\frac{1}{n\alpha}\sum_{i=1}^n[s_i-\lambda]_+\}$, which is also called the empirical of $C_{\alpha}[s]$. The CVaR has a natural distributionally robust optimization interpretation \citep{shapiro2014lectures}. Therefore, from equation (\ref{eq:opt_ar_loss}), we know the population version of \aorr~$\mathcal{L}(f,\lambda, \hat{\lambda})=\underset{ \lambda}{\mbox{inf}} \underset{\hat{\lambda}}{\mbox{sup}} \left\{\mathbb{E}[[s(Yf(X))-\lambda]_+ - [s(Yf(X))-\hat{\lambda}]_+]+(\nu\lambda-\mu \hat{\lambda})\right\}$ can be expressed by difference of two CVaRs. We call it as Interval Conditional Value at Risks (ICVaRs).
\begin{equation}
    \begin{aligned}
    \mathcal{L}(f,\lambda, \hat{\lambda}) = \nu C_{\nu}[s(\theta)] - \mu C_{\mu}[s(\theta)].
    \end{aligned}
\end{equation}
Figure \ref{fig:interpretation_CVaR} is an interpretation of ICVaRs. We denote the empirical form of $\mathcal{L}(f,\lambda, \hat{\lambda})$ by $\widehat{\mathcal{L}}(f,\lambda, \hat{\lambda})$, and 
\begin{equation}
    \begin{aligned}
    \widehat{\mathcal{L}}(f,\lambda, \hat{\lambda})= \nu \widehat{C}_{\nu}[s(\theta)] - \mu \widehat{C}_{\mu}[s(\theta)].
    \end{aligned}
\end{equation}
The following theorem provides some deviation convergence bounds for estimating $\psi_{m,k}(S(\theta))$ from a finite number of independent samples.
\begin{theorem}\label{theorem:ICVaRs}
If $supp(s(\theta))\subseteq[a,b]$ and $s$ has a continuous distribution function, then for any $\delta\in(0,1]$,
\begin{equation*}
    \begin{aligned}
    &\Pr\left(\mathcal{L}(f,\lambda, \hat{\lambda})-\widehat{\mathcal{L}}(f,\lambda, \hat{\lambda}) \leq(b-a)\left[\sqrt{\frac{5k\ln(3/\delta)}{n^2}}+\sqrt{\frac{\ln(1/\delta)}{2n}}\right]\right)\geq 1-\delta,\\
    &\Pr\left(\mathcal{L}(f,\lambda, \hat{\lambda})-\widehat{\mathcal{L}}(f,\lambda, \hat{\lambda}) \geq -(b-a)\left[\sqrt{\frac{5m\ln(3/\delta)}{n^2}}+\sqrt{\frac{\ln(1/\delta)}{2n}}\right]\right)\geq 1-\delta.
    \end{aligned}
\end{equation*}
\end{theorem}
Proof can be found in Appendix \ref{proof:ICVaRs}. It should be mentioned that we can derive better generalization bounds in some mild conditions (sub-Gaussian, light-tailed, and heavy-tailed data distributions) according to \cite{prashanth2020concentration, thomas2019concentration}. Note that the non-asymptotic relation between the excess generalization induced by \aorr~and the target classification problem (e.g. excess misclassification error) is a fundamentally important question. However, it is very hard to establish this relation since the \aorr~involves that the difference of two possible convex losses, i.e.  $[s(yf_\theta(x))-\lambda]_+ -[s(yf_\theta(x))-\hat{\lambda}]_+$. We leave this relation for future study.

\subsection{Determine $k$ and $m$} \label{sec:determin_k_m}
Hyper-parameters $k$ and $m$  are very important for the final performance of the \aorr~aggregate loss. However, before actually training the model, it is difficult to determine their suitable values so that the model can achieve the best performance. We cannot fix them to some specific values. For different data sets, they may have different suitable values. Greedy search is a good method for us to find the optimal $k$ and $m$ in a finite time, but it only works on simple data sets that contain a small number of samples. For large data sets, the greedy search method is not efficient and could be very time-consuming. 

In practice, for large-scale data sets, we can decide on an approximate value of $m$ if we have prior knowledge about the faction of outliers in the data set. To avoid extra freedom due to the value of $k$, we follow a very popular adaptive setting that has been applied in previous works (e.g., \cite{kawaguchiordered}). At the beginning of training, $k$ equals to the size ($n$) of training data, $k=\lfloor \frac{n}{2} \rfloor$ once training accuracy $\geq 70\%$,  $k=\lfloor \frac{n}{4} \rfloor$ once training accuracy $\geq 80\%$, $k=\lfloor \frac{n}{8} \rfloor$ once training accuracy $\geq 90\%$, $k=\lfloor \frac{n}{16} \rfloor$ once training accuracy $\geq 95\%$, $k=\lfloor \frac{n}{32} \rfloor$ once training accuracy $\geq 99.5\%$. However, if we do not have prior knowledge about the fraction of outliers in the data set, the adaptive setting method still needs to try different values of $m$ from its feasible space and then find the optimal values of $k$ and $m$. Obviously, the time complexity is dominated by the size of $m$'s feasible set. Therefore, this method will take a long time to search for the optimal hyper-parameters if we apply it to a large data set.   

Many existing works on label corruption or label noise problem assume that training data is not clean and potentially includes noise and outliers. However, in general, there are many trusted samples are available. These trusted data can be collected to create a clean validation set. This assumption has been analyzed in \cite{charikar2017learning} and also been used by other works for designing a robust learning model \citep{hendrycks2018using, ren2018learning, veit2017learning, li2017learning}. \cite{hendrycks2018using} proposes a loss correction technique for deep neural network classifiers that use clean data to alleviate the influences of label noise. \cite{li2017learning} proposes a distillation framework by using a clean data set to reduce the risk of learning from noisy labels. We apply this assumption to our problem. Specifically, we extract a clean validation set from training data to determine the values of $k$ and $m$. According to Theorem \ref{theorem:sorr}, fix $\theta$, we will get the optimal values of $\lambda$ and $\hat{\lambda}$ if we know the values of $k$ and $m$. Based on the values of $\lambda$ and $\hat{\lambda}$, we can calculate how many individual losses are larger than them. These also provide us information about $k$ and $m$. Therefore, we can directly calculate the optimal $\lambda$ and $\hat{\lambda}$ based on the validation set instead of determining the optimal values of $k$ and $m$.

\begin{figure}
\centering
    \includegraphics[trim=1 1 1 1, clip,keepaspectratio, width=0.8\textwidth]{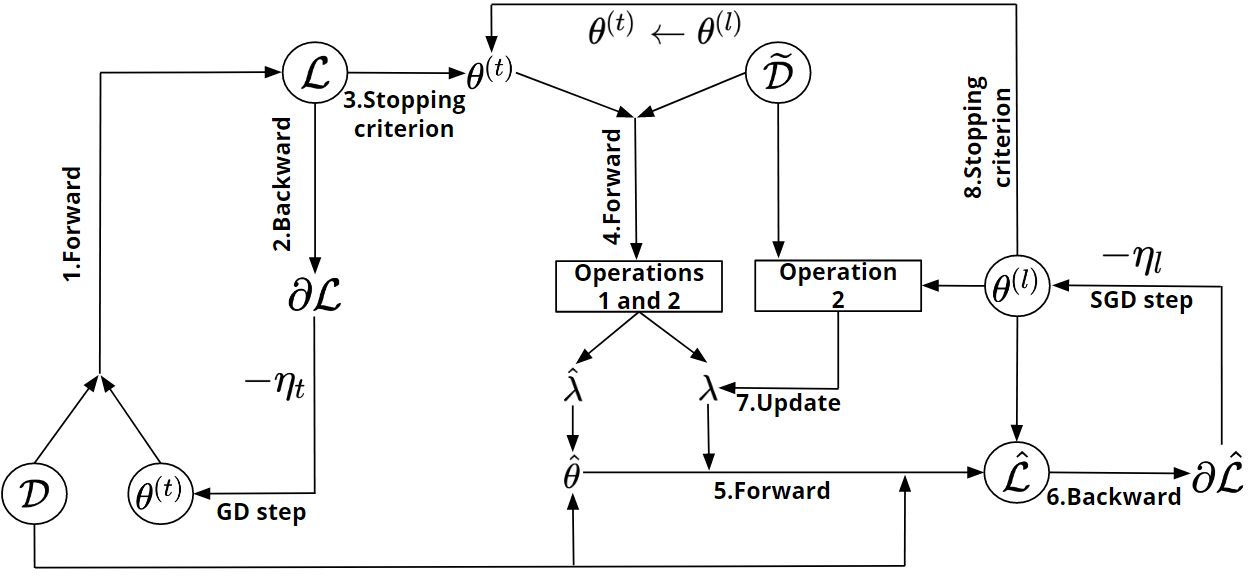}
  \caption{ The framework of learning hyper-parameters. Note that we calculate the mean value plus one standard deviation of all individual losses from $\widetilde{\mathcal{D}}$ and then regard it as operation 1 and regard the mean value minus two standard deviations of all individual losses from $\widetilde{\mathcal{D}}$ as operation 2 in practice.}
\label{fig:learning_hyperparameter_framework}
\end{figure}

According to the above analysis, we propose a framework (Figure \ref{fig:learning_hyperparameter_framework}) to combine hyper-parameters learning and Algorithm \ref{Alg0}. To better learn the parameters of $\lambda$ and $\hat{\lambda}$, we first use the average aggregate loss to ``warm-up'' the model for some epochs by training on the training data $\mathcal{D}$, which contains outliers. This pre-processing has been used in many existing works such as \cite{li2020dividemix, hendrycks2018using}. As shown in Figure \ref{fig:learning_hyperparameter_framework}, to ``warm-up'' the model, one needs to use forward and backward (see Steps 1 and 2) to update the model parameter $\theta^{(t)}$ with learning rate $\eta_t$. After a few epochs, with a stopping criterion (Step 3), we use model $\theta^{(t)}$ on the clean validation set $\widetilde{\mathcal{D}}$ to learn the individual losses (Step 4). Through simple operations 1 and 2, we can get $\hat{\lambda}$ and $\lambda$. For example, we can calculate the mean value plus one standard deviation of all individual losses from $\widetilde{\mathcal{D}}$ and then regard it as operation 1 and regard the mean value minus two standard deviations of all individual losses from $\widetilde{\mathcal{D}}$ as operation 2. Note that we should keep $\hat{\lambda}>\lambda$ according to Theorem \ref{theorem:sorr}.  When we have $\hat{\lambda}$, we can obtain sub-gradient $\hat{\theta}$ based on training data $\mathcal{D}$ and equation (\ref{eq:subgradient_phi_m}). With $\mathcal{D}$, $\hat{\theta}$ and $\lambda$, we use an inner loop and stochastic gradient descent method with learning rate $\eta_l$ to update the inner convex model parameter $\theta^{(l)}$. Meanwhile, apply $\theta^{(l)}$ on $\widetilde{\mathcal{D}}$ and re-use operation 2 to update the parameter $\lambda$. This procedure (Steps 5, 6, and 7) is similar to equation (\ref{eq:SGD}). With another stopping criterion (Step 8), we can update $\theta^{(t)}$ by using $\theta^{(l)}$ and redo steps 4, 5, 6, 7, and  8 repeatedly until the final stopping criterion is satisfied. We list detailed step-by-step pseudo-code in Algorithm \ref{Alg1}. For a given outer loop size $|t|$, an inner loop size $|l|$, the ``warm-up'' loop size $|p|$, training sample size $n$, and the validation sample size $\widetilde{n}$, the time complexity of Algorithm 2 is $O((|p|+|t|)n + |t|\cdot |l|\cdot \widetilde{n})$.

\begin{algorithm}[t]
    \caption{DCA for Minimizing \aorr~without Setting $k$ and $m$}\label{Alg1}
    \SetAlgoLined
    \textbf{Initialization:} $\theta^{(0)}$, $\eta_t$, $\eta_l$,  $\hat{\lambda}=0$, $\lambda=0$, flag$=$0, training data set $\mathcal{D}$ (with outliers), and validation data set $\widetilde{\mathcal{D}}$ (without outliers). \\

    
    \For{$t=0,1,...$}{
    \eIf{flag $==$ 0}{
        \For{$p=0,1,...$}{
        $\theta^{(p+1)} \leftarrow \theta^{(p)}-\frac{\eta_t}{|\mathcal{D}|} \sum_{i\in \mathcal{D}} \partial s_i(\theta)$
        
        \If{Stopping criterion}{
         $\hat{\lambda} \leftarrow$ operation1($\{s_j(\theta^{(p+1)})|j\in \widetilde{\mathcal{D}}\}$)\\
        $\lambda \leftarrow$ operation2($\{s_j(\theta^{(p+1)})|j\in \widetilde{\mathcal{D}}\}$)\\
        flag $\leftarrow$ 1,
        $\theta^{(0)} \leftarrow \theta^{(p+1)}$\\
        \textbf{break}
        }
        }
       }{
       Compute $\hat{\theta}^{(t)} \in \frac{1}{|\mathcal{D}|}\sum_{i \in \mathcal{D}}\partial s_i(\theta^{(t)})\cdot \mathbb{I}_{[s_i(\theta^{(t)})>\hat{\lambda}]}$
       
       \For{$l=0,1,...$}{
       Randomly sample $s_{i_l}(\theta^{(l)})$ from the collection of $\{s_i(\theta^{(l)})\}_{i\in\mathcal{D}}$\\
       $\theta^{(l+1)} \leftarrow \theta^{(l)}-\eta_l\left(\partial s_{i_l}(\theta^{(l)})\cdot \mathbb{I}_{[s_{i_l}(\theta^{(l)})>\lambda]} - \hat{\theta}^{ (t)}\right)$ \\
       $\lambda \leftarrow$ operation2($\{s_j(\theta^{(l+1)})|j\in \widetilde{\mathcal{D}}\}$)
       }
       
       Update $\theta^{(t+1)} \leftarrow \theta^{(l+1)}$ \\
       $\hat{\lambda} \leftarrow$ operation1($\{s_j(\theta^{(t+1)})|j\in \widetilde{\mathcal{D}}\}$),
       $\lambda \leftarrow$ operation2($\{s_j(\theta^{(t+1)})|j\in \widetilde{\mathcal{D}}\}$)\\
      }
    
    }
    
\end{algorithm}

\subsection{Experiments}

We empirically demonstrate the effectiveness of the \aorr~aggregate loss combined with two types of individual losses for binary classification, namely, the logistic loss and the hinge loss. For simplicity, we consider a linear prediction function $f(x;\theta) = \theta^T x$ with parameter $\theta$, and the $\ell_2$ regularizer $\frac{1}{2C}||\theta||_2^2$ with $C>0$. For binary classification, we apply the greedy search method to select optimal hyper-parameters $k$ and $m$. We use the MNIST data set with symmetric (i.e. uniformly random) label noise to verify the effectiveness of our Algorithm \ref{Alg1} for learning hyper-parameters and show the \aorr~aggregate loss can also be extended to multi-class classification. All algorithms are implemented in Python 3.6 and trained and tested on an Intel(R) Xeon(R) CPU W5590 @3.33GHz with 48GB of RAM.

\begin{figure*}[t!]
\captionsetup[subfigure]{justification=centering}
\centering
        \begin{subfigure}[b]{0.245\textwidth}
                \includegraphics[width=\linewidth]{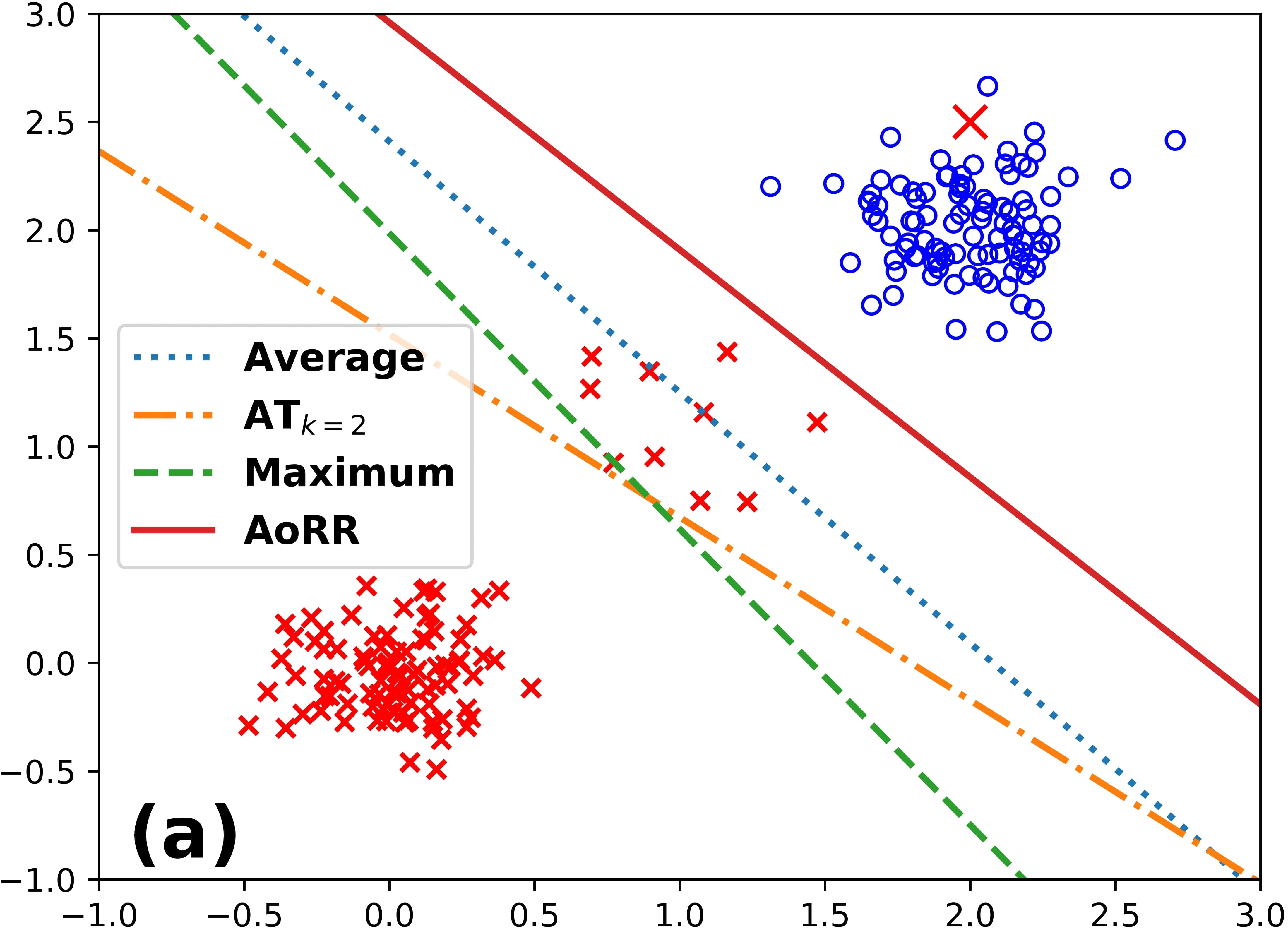}
        \end{subfigure}%
        \begin{subfigure}[b]{0.245\textwidth}
                \includegraphics[width=\linewidth]{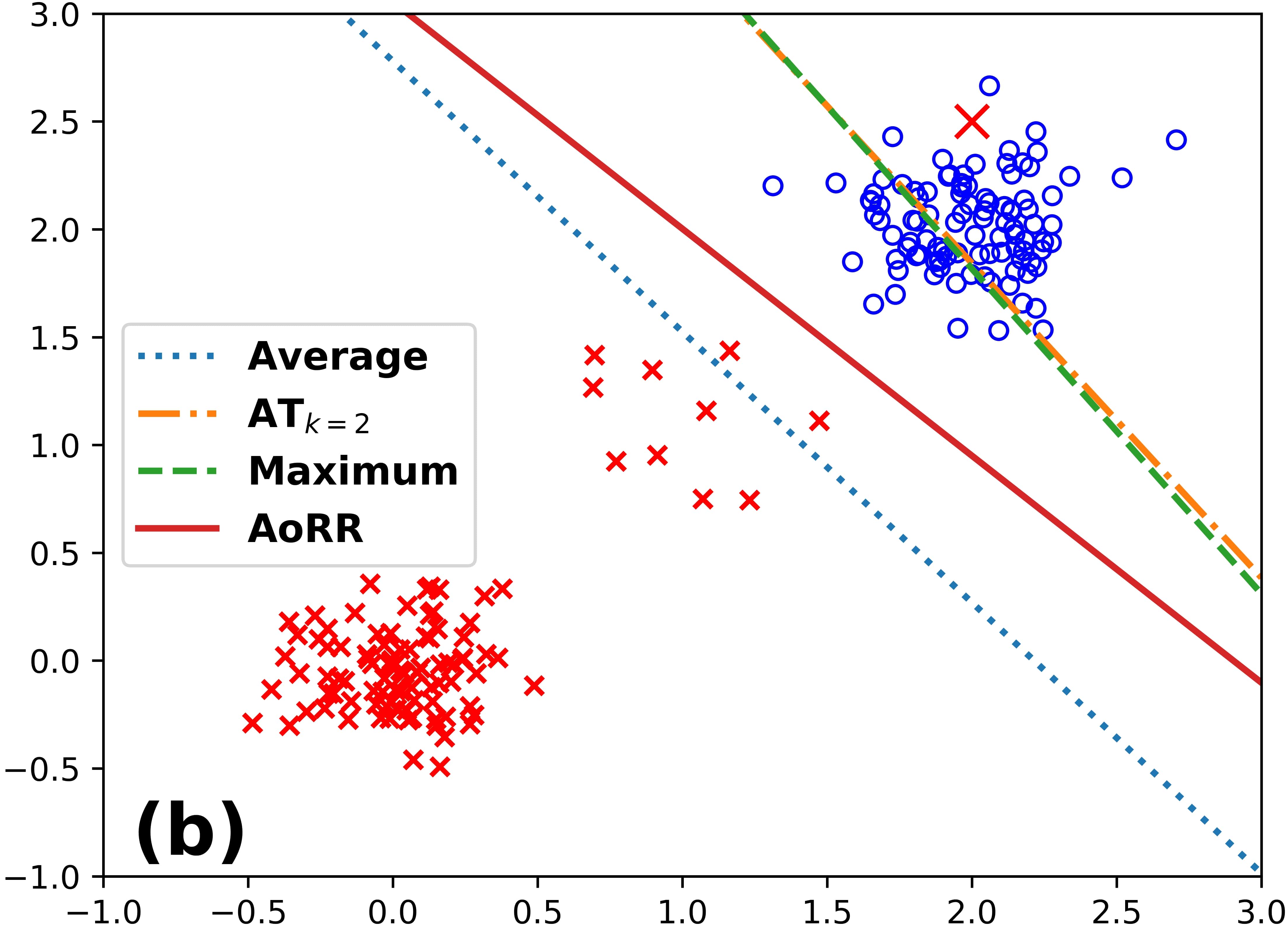}
        \end{subfigure}%
        \rulesep
        \begin{subfigure}[b]{0.245\textwidth}
                \includegraphics[width=\linewidth]{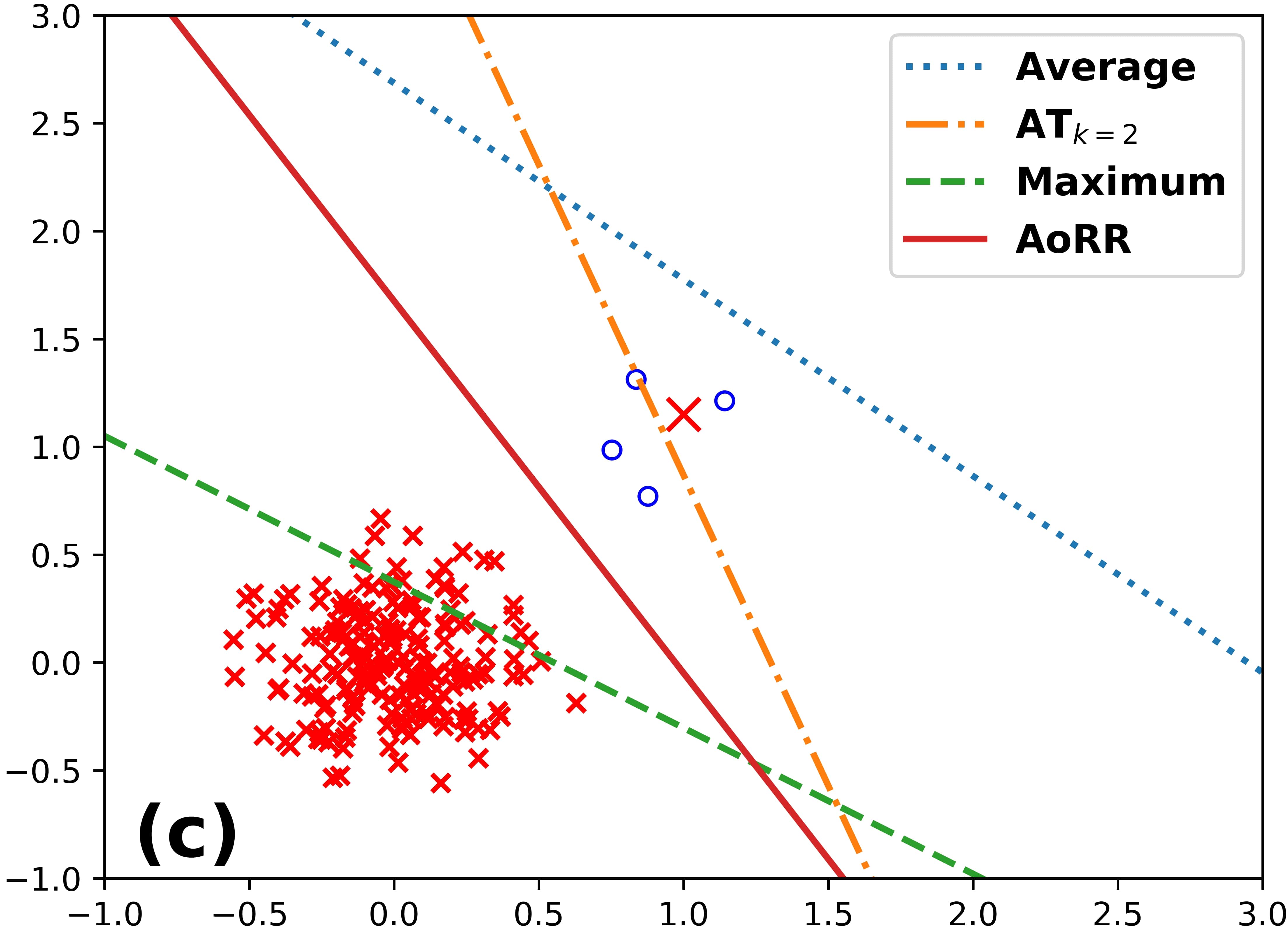}
        \end{subfigure}%
        \begin{subfigure}[b]{0.245\textwidth}
                \includegraphics[width=\linewidth]{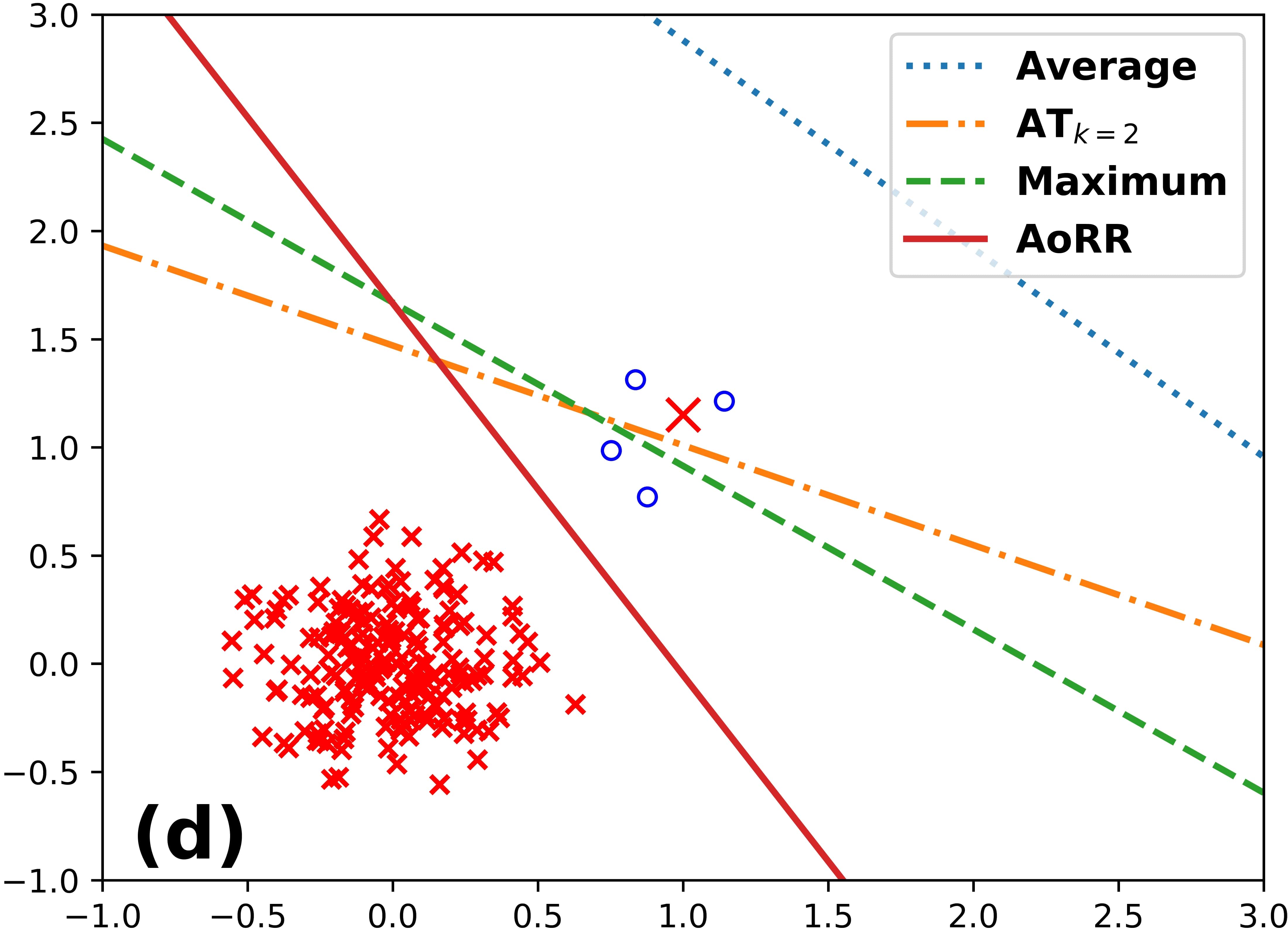}
        \end{subfigure}
        \bigskip
        \begin{subfigure}[b]{0.251\textwidth}            \includegraphics[width=\linewidth]{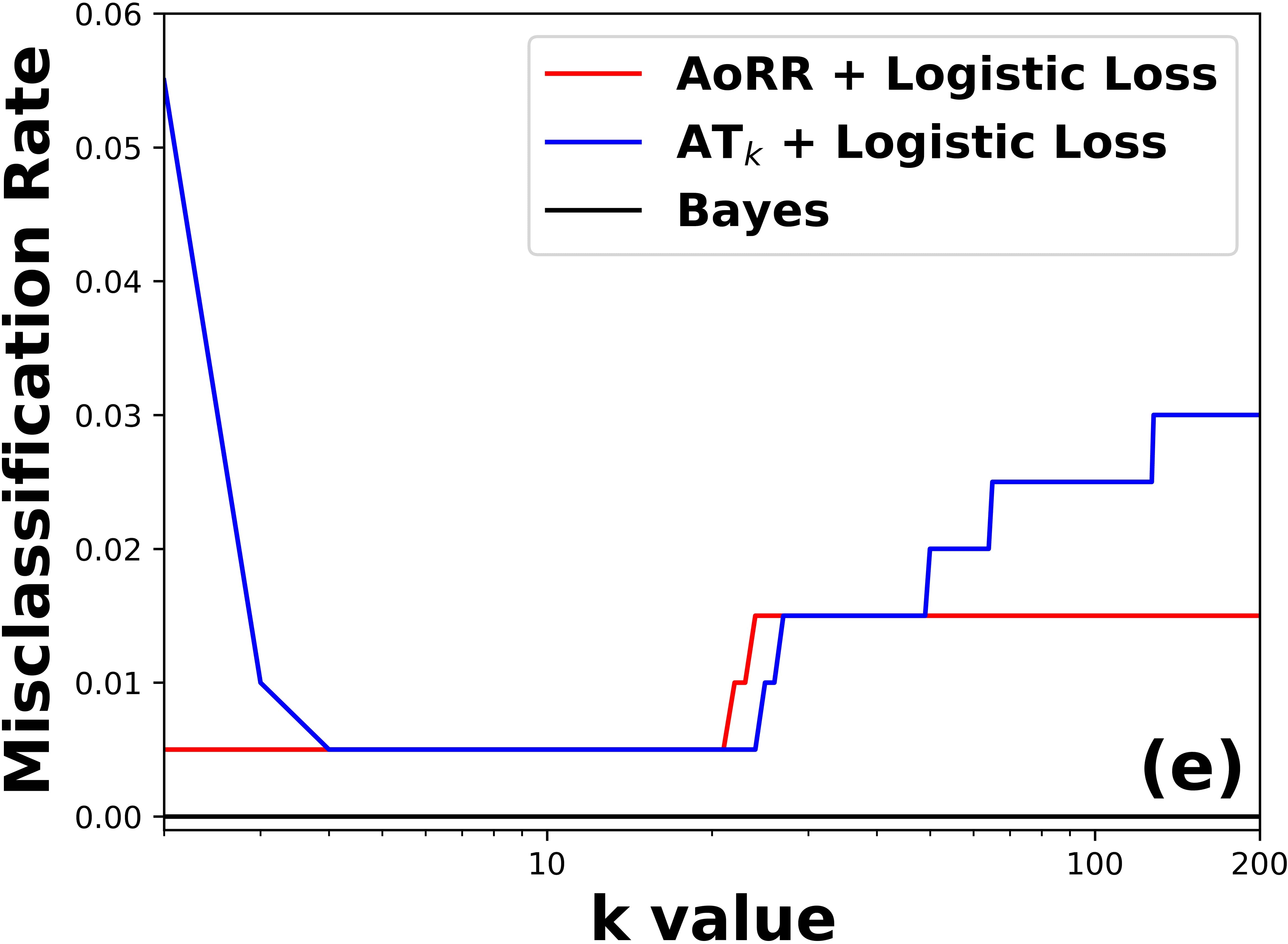}
        \end{subfigure}%
        \begin{subfigure}[b]{0.243\textwidth}
                \includegraphics[width=\linewidth]{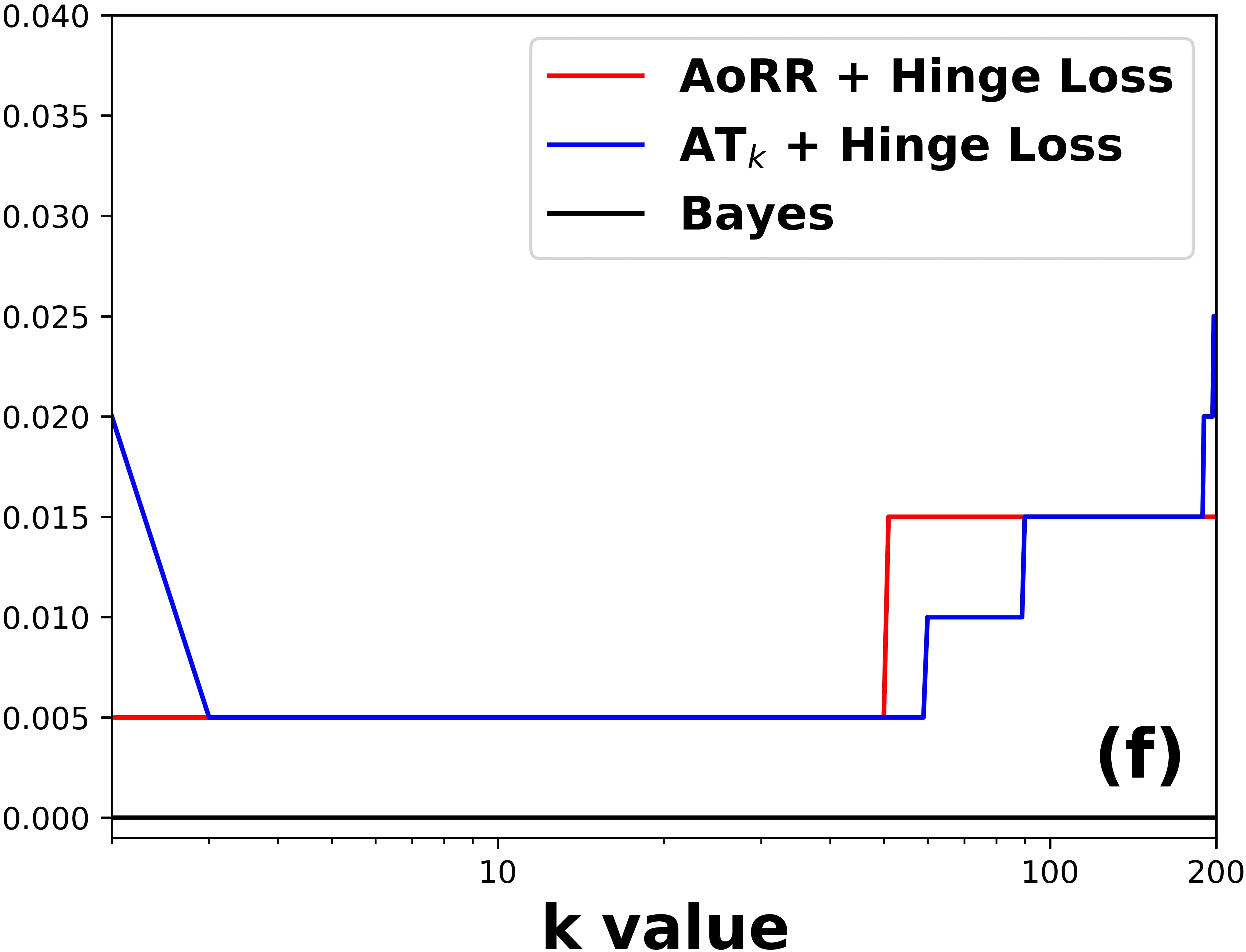}
        \end{subfigure}%
        \rulesep
        \begin{subfigure}[b]{0.243\textwidth}
                \includegraphics[width=\linewidth]{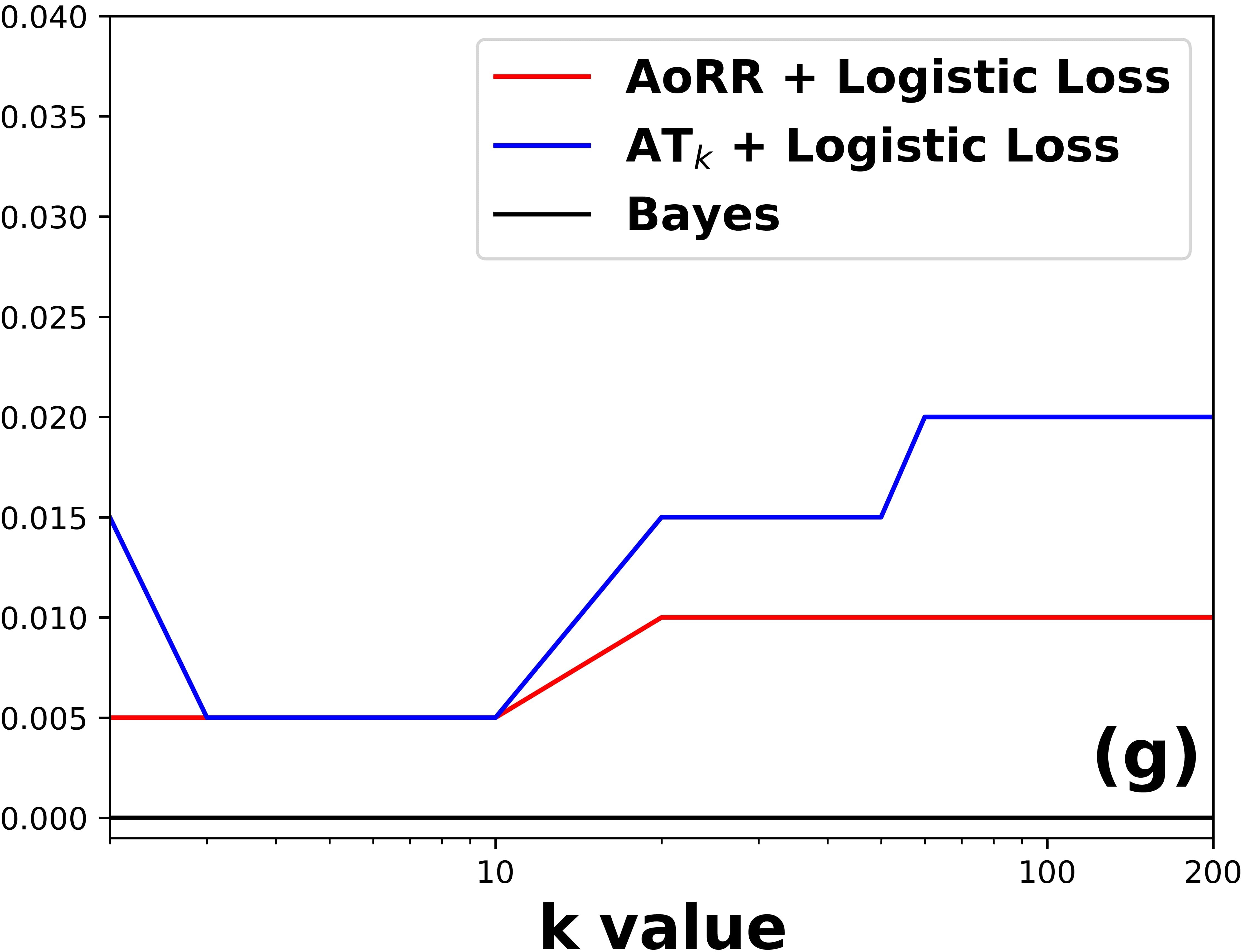}
        \end{subfigure}%
        \begin{subfigure}[b]{0.246\textwidth}
                \includegraphics[width=\linewidth]{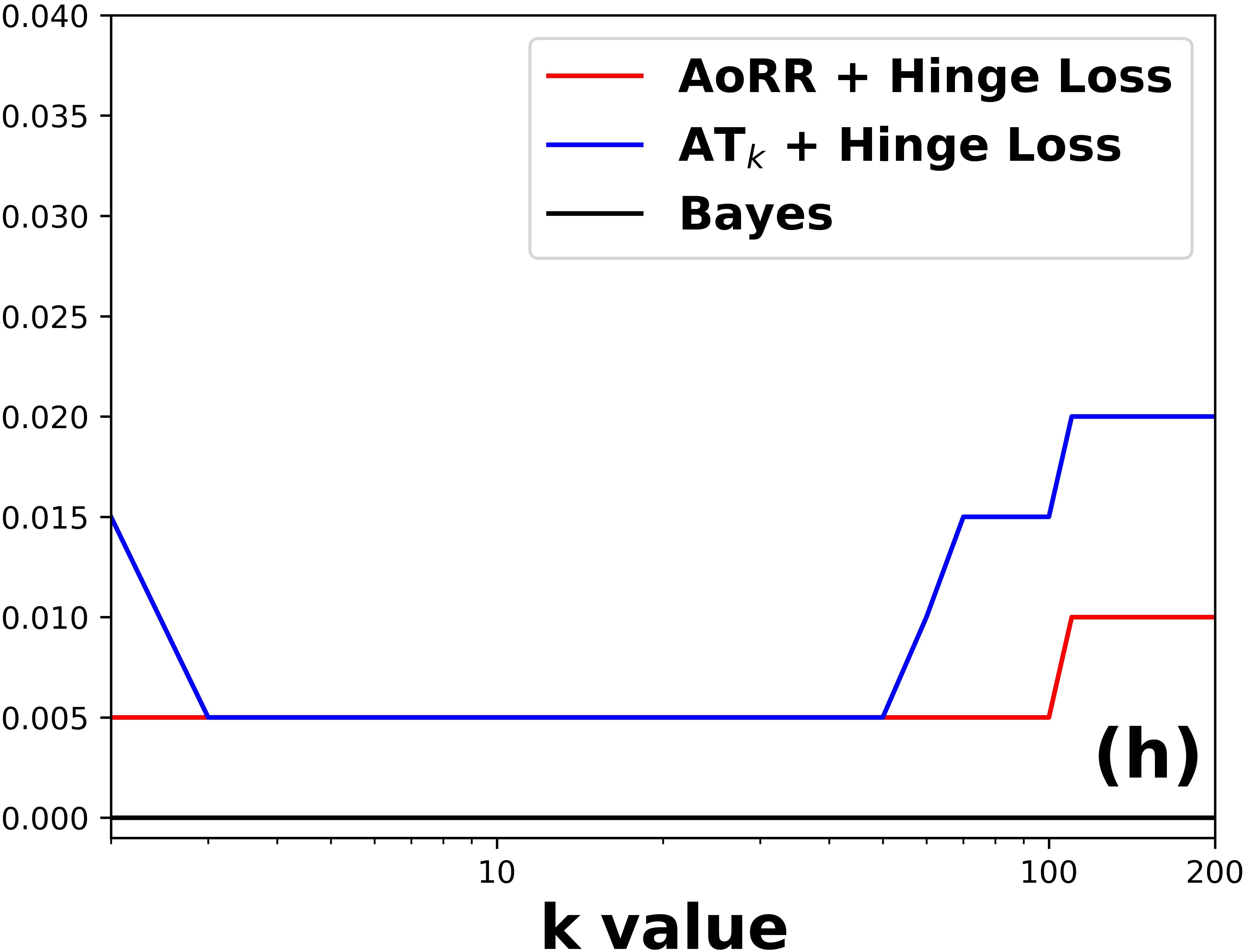}
        \end{subfigure}
        \vspace{-2em}
        \caption{ Comparison of different aggregate losses for binary classification on a balanced but multi-modal synthetic data set and with outliers with logistic loss (a) and hinge loss (b), and an imbalanced synthetic data set with outliers with logistic loss (c) and hinge loss (d). Outliers in data are shown as $\times$ in blue class. {The figures (e), (f), (g) and (h) show the misclassification rates of \aorr~w.r.t. different values of $k$ for each case and compare with the AT$_k$ and the optimal Bayes classifier.}}\label{fig: aggregate_loss_toy_example}
\end{figure*}

\subsubsection{\aorr~for Binary Classification}

\textbf{Synthetic data.} We generate two sets of 2D synthetic data (Figure \ref{fig: aggregate_loss_toy_example}). Each data set contains $200$ samples from Gaussian distributions with different means and variances. We consider both the case of the balanced (Figure \ref{fig: aggregate_loss_toy_example} (a,b)) and the imbalanced  (Figure \ref{fig: aggregate_loss_toy_example} (c,d)) data distributions, in the former the training data for the two classes are approximately equal while in the latter one class has a dominating number of samples in comparison to the other. The learned linear classifiers with different aggregate losses are shown in Figure \ref{fig: aggregate_loss_toy_example}. Both data sets have an outlier in the blue class (shown as $\times$). 


To optimally remove the effect of outliers, we need to set $k$ larger than the number of outliers in the training data set. Since there is one outlier in this synthetic data set, we select $k=2$ here as an example. As shown in Figure \ref{fig: aggregate_loss_toy_example}, neither the maximum loss nor the average loss performs well on the synthetic data set, due to the existence of outliers and the multi-modal nature of the data. Furthermore, 
Figure \ref{fig: aggregate_loss_toy_example} also shows that the AT$_k$ loss does not bode well: it is still affected by outliers. The reason can be that the training process with the AT$_k$ loss with $k=2$ will most likely pick up one individual loss from the outlier for optimization.  In contrast, the \aorr~loss with $k$=2 and $m$=1, which is equivalent to the top-$2$ or second-largest individual loss, yields better classification results. Intuitively, we avoid the direct effects of the outlier since it has the largest individual loss value. {Furthermore, we perform experiments to show misclassification rates of \aorr~with respect to different values of $k$ in Figure \ref{fig: aggregate_loss_toy_example} (e), (f), (g), (h) for each case and compare with the AT$_k$ loss and optimal Bayes classifier. The results show that for $k$ values other than 2, the \aorr~loss still exhibits an advantage over the AT$_k$ loss. Our experiments are based on a grid search for selecting the value of $k$ and $m$ because we found it is simple and often yields comparable performance.} 

\begin{figure*}[t]
\captionsetup[subfigure]{justification=centering}
\centering
        \begin{subfigure}[b]{0.24\textwidth}
                \includegraphics[width=\linewidth]{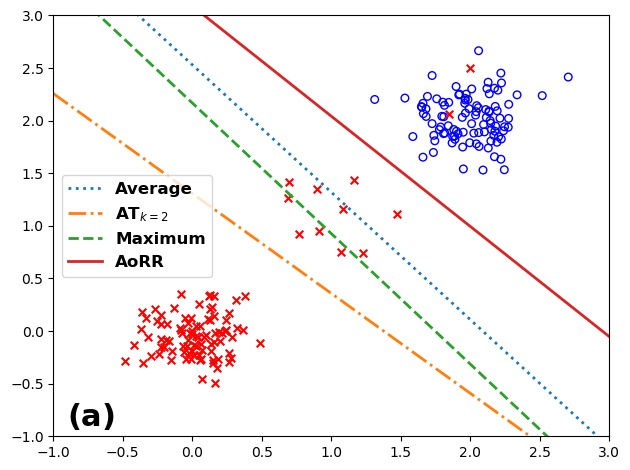}
                \label{fig:LogisticRegression_data4_2}
        \end{subfigure}%
        \begin{subfigure}[b]{0.24\textwidth}
                \includegraphics[width=\linewidth]{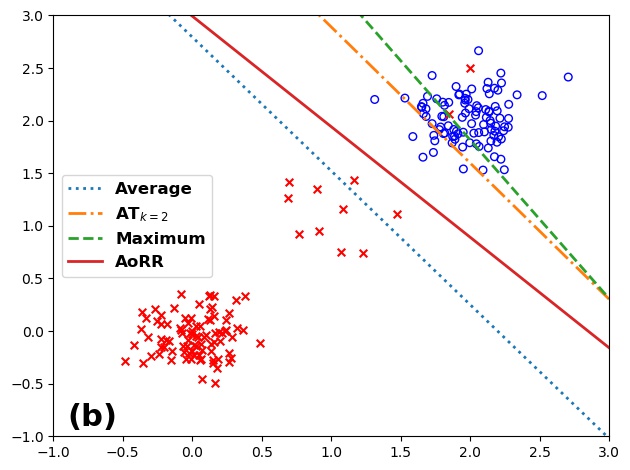}
                \label{fig:Hinge_data4_2}
        \end{subfigure}%
        \rulesep
        \begin{subfigure}[b]{0.24\textwidth}
                \includegraphics[width=\linewidth]{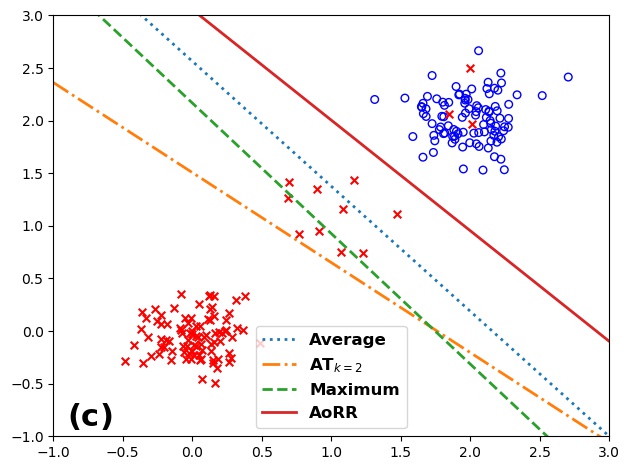}
                \label{fig:LogisticRegression_data4_3}
        \end{subfigure}%
        \begin{subfigure}[b]{0.24\textwidth}
                \includegraphics[width=\linewidth]{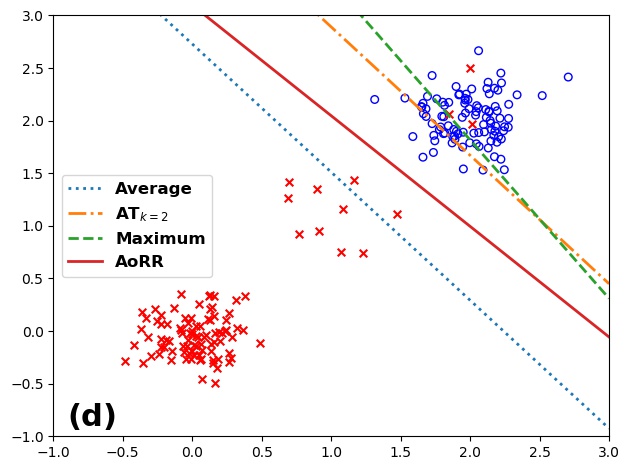}
                \label{fig:Hinge_data4_3}
        \end{subfigure}
        \begin{subfigure}[b]{0.24\textwidth}
                \includegraphics[width=\linewidth]{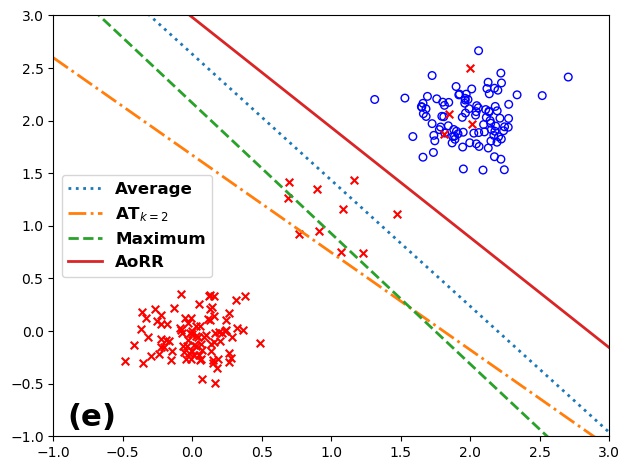}
                \label{fig:LogisticRegression_data4_4}
        \end{subfigure}%
        \begin{subfigure}[b]{0.24\textwidth}
                \includegraphics[width=\linewidth]{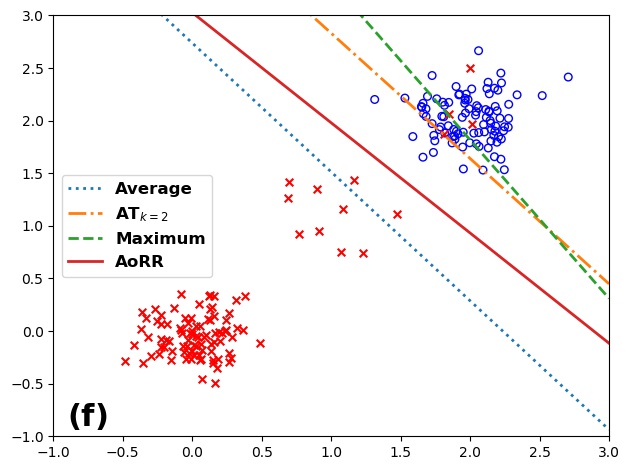}
                \label{fig:Hinge_data4_4}
        \end{subfigure}%
        \rulesep
        \begin{subfigure}[b]{0.24\textwidth}
                \includegraphics[width=\linewidth]{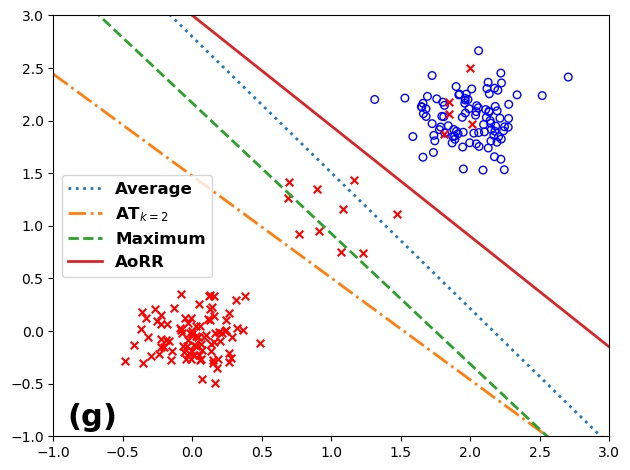}
                \label{fig:LogisticRegression_data4_5}
        \end{subfigure}%
        \begin{subfigure}[b]{0.24\textwidth}
                \includegraphics[width=\linewidth]{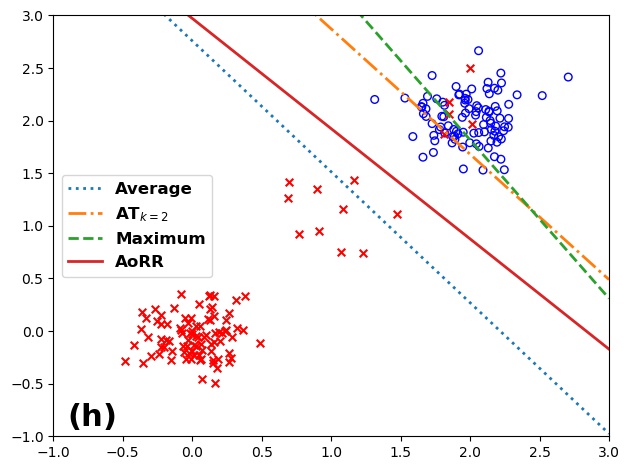}
                \label{fig:Hinge_data4_5}
        \end{subfigure}
        \begin{subfigure}[b]{0.24\textwidth}              \includegraphics[width=\linewidth]{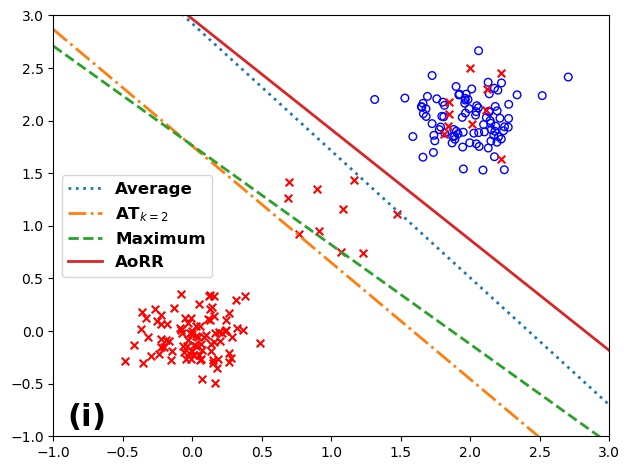}
                \label{fig:LogisticRegression_data4_10}
        \end{subfigure}%
        \begin{subfigure}[b]{0.24\textwidth}
                \includegraphics[width=\linewidth]{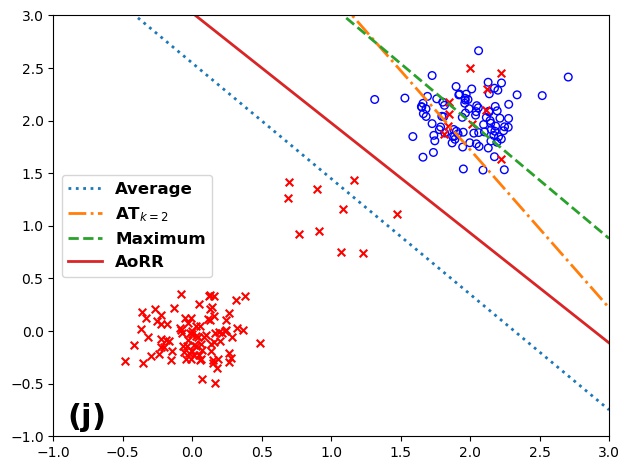}
                \label{fig:Hinge_data4_10}
        \end{subfigure}%
        \rulesep
        \begin{subfigure}[b]{0.24\textwidth}
                \includegraphics[width=\linewidth]{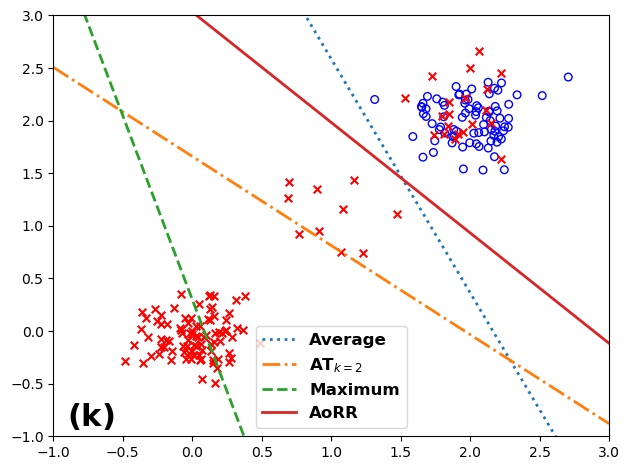}
                \label{fig:LogisticRegression_data4_20}
        \end{subfigure}%
        \begin{subfigure}[b]{0.24\textwidth}
                \includegraphics[width=\linewidth]{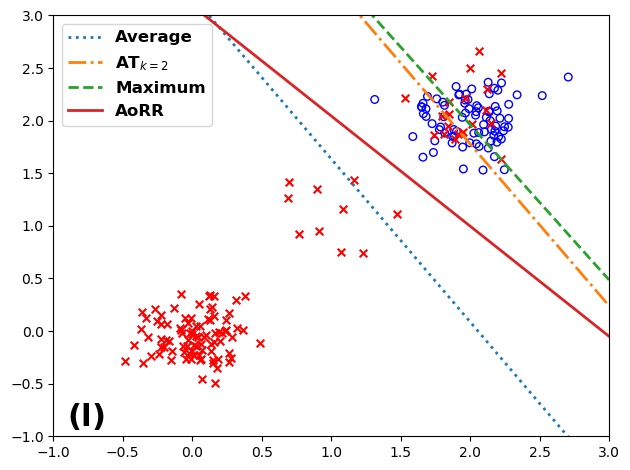}
                \label{fig:Hinge_data4_20}
        \end{subfigure}
        \caption{ Comparison of different aggregate losses on 2D synthetic data with 200 samples for binary classification with individual logistic loss (a, c, e, g, i, k) and individual hinge loss (b, d, f, h, j, l). Outliers are shown as $\times$ in blue class.}\label{fig: additional_toy_examples}
\end{figure*}

In order to evaluate the effects of different aggregate losses on more than one outlier, we also conduct additional experiments on the multi-modal toy example that includes more outliers.
We consider six cases as follows, 
\begin{compactitem}
    \item \textbf{Case 1 (2 outliers).} In Figure \ref{fig: additional_toy_examples} (a) and (b), there exist two outliers. Let hyper-parameters $k=3$ and $m=2$. 
    \item \textbf{Case 2 (3 outliers).} Figure \ref{fig: additional_toy_examples} (c) and (d) contain three outliers. In this scenario, $k=4$ and $m=3$.  
    \item \textbf{Case 3 (4 outliers).} Figure \ref{fig: additional_toy_examples} (e) and (f) include four outliers and we set $k=5$ and $m=4$. 
    \item \textbf{Case 4 (5 outliers).} There are five outliers in Figure \ref{fig: additional_toy_examples} (g) and (h). We set $k=6$ and $m=5$.
    \item \textbf{Case 5 (10 outliers).} Ten outliers have been included in Figure \ref{fig: additional_toy_examples} (i) and (j). Let $k=11$ and $m=10$ in this case.
    \item \textbf{Case 6 (20 outliers).} We create twenty outliers in Figure \ref{fig: additional_toy_examples} (k) and (l) and make $k=21$ and $m=20$.
\end{compactitem}
Seeing on cases 1, 2, 3, and 4, the linear classifier learned from average aggregate loss cross some red samples from minor distribution even though the data is separable. The reason is that the samples close to the decision boundary are sacrificed to reduce the total loss over the whole data set. 
The AT$_k$ loss selects $k$ largest individual losses which contain many outliers to train the classifier. It leads to the instability of the learned classifier. This phenomenon can be found when we compare all cases. Similarly, the maximum aggregate loss cannot fit this data very well in all cases. This loss is very sensitive to outliers.

In cases 5 and 6, the average aggregate loss with individual logistic loss achieves better results than with individual hinge loss. A possible reason is that for correctly classified samples with a margin greater than 1, the penalty caused by hinge loss is 0. However, it is non-zero when using logistic loss. Since many outliers in the blue class, to reduce the average loss, the decision boundary will close to the blue class. Especially, when we compare (i) and (k), it is obvious that average loss can achieve a better result while the number of outliers is increasing.

As we discussed, the hinge loss has less penalty for correctly classified samples than logistic loss. This causes outliers to be more prominent than normal samples while using the individual hinge loss. This result can be verified in the experiment when we compare the individual logistic loss and the individual hinge loss. For example, (i) and (j), (k) and (l), etc. We find the decision boundaries of maximum loss and AT$_k$ loss are close to outliers in the individual hinge loss scenario because both of them are sensitive to outliers in our cases.

\begin{table*}[]
\centering
\scalebox{0.8} { 
\begin{tabular}{|c|cccc|}
\hline
        Data Sets   &  \#Classes   &    \#Samples &  \#Features   & Class Ratio  \\ \hline
        Monk        &  2   & 432 & 6 & 1.12\\ 
        Australian  &  2   & 690 & 14 & 1.25\\ 
        Phoneme     &  2   & 5,404 & 5 & 2.41\\ 
        Titanic     &  2   & 2,201 & 3 & 2.10\\ 
        Splice      &  2   & 3,175 & 60 & 1.08\\ \hline
\end{tabular}
}
\vspace{1mm}
\caption{ Statistical information of five real data sets.}
\label{tab:aggregate_real_data sets}
\end{table*}

\begin{table*}[]
\centering
\setlength\tabcolsep{5pt}
\scalebox{0.8} { 
\begin{tabular}{|c|ccccc||ccccc|}
\hline
\multirow{2}{*}{Data Sets} & \multicolumn{5}{c||}{Logistic Loss} & \multicolumn{5}{c|}{Hinge Loss} \\ \cline{2-11} 
                  &  Maximum  & R\_Max &    Average &  AT$_k$   & \aorr  &    Maximum & R\_Max  &    Average &  AT$_k$   & \aorr    \\ \hline
        \multirow{2}{*}{Monk} &  22.41 & 21.69  &  20.46   &  16.76   &  \textbf{12.69}   & 22.04 & 20.61  & 18.61    &  17.04   & \textbf{13.17}   \\  
                  &  (2.95) & (2.62)  &  (2.02)   & (2.29)   &  \textbf{(2.34)}   & (3.08) & (3.38)  & (3.16)    &  (2.77)   & \textbf{(2.13)}   \\ \hline
\multirow{2}{*}{Australian} &  19.88& 17.65  &  14.27   &  11.7   &  \textbf{11.42}   &  19.82& 15.88  &  14.74   &  12.51   & \textbf{12.5}   \\  
                  &  (6.64)& (1.3)  &  (3.22)   &  (2.82)   &  \textbf{(1.01)}   &  (6.56)& (1.05)  &  (3.10)   &  (4.03)   & \textbf{(1.55)}   \\ \hline
\multirow{2}{*}{Phoneme} &  28.67& 26.71   &  25.50   &  24.17   & \textbf{21.95}    & 28.81&  24.21   &  22.88   &  22.88   &  \textbf{21.95}   \\  
                  &  (0.58)& (1.4)   &  (0.88)   &  (0.89)   & \textbf{(0.71)}    & (0.62)&  (1.7)   &  (1.01)   &  (1.01)   &  \textbf{(0.68)} \\ \hline
\multirow{2}{*}{Titanic} &  26.50& 24.15 &  22.77   &  22.44   & \textbf{21.69}    &  25.45& 25.08    &  22.82   &  22.02   &  \textbf{21.63}   \\  
                  &  (3.35)& (3.12)   &  (0.82)   &  (0.84)   & \textbf{(0.99)}    &  (2.52)& (1.2)   &  (0.74)   &  (0.77)   &  \textbf{(1.05)} \\\hline 
\multirow{2}{*}{Splice} &  23.57&  23.48&  17.25 &  16.12 & \textbf{15.59}    & 23.40&  22.82& 16.25 &  16.23  & \textbf{15.64}   \\  
                  &  (1.93)&  (0.76)  &  (0.93)   &  (0.97)   & \textbf{(0.9)}    & (2.10)&  (2.63)   & (1.12)    &  (0.97)   & \textbf{(0.89)}   \\ \hline
\end{tabular}
}
\caption{ Average error rate (\%) and standard derivation of
different aggregate losses combined with individual logistic loss and hinge loss over 5 data sets. The best results are shown in bold. (R\_Max: Robust\_Max) }
\label{tab:aggregate_real_experiments}

\end{table*}

\textbf{Real data.} We use five benchmark data sets from the UCI \citep{Dua:2019} and the KEEL \citep{alcala2011keel} data repositories (statistical information of each data set is given in Table \ref{tab:aggregate_real_data sets}). For each data set, we first randomly select $50\%$ samples for training, and the remaining $50\%$ samples are randomly split for validation and testing (each contains 25\% samples). Hyper-parameters $C$, $k$, and $m$ are selected based on the validation set. Specifically, parameter $C$ is chosen from $\{10^0, 10^1, 10^2, 10^3, 10^4, 10^5\}$, parameter $k\in\{1\}\cup [0.1:0.1:1]n$, where $n$ is the number of training samples, and parameter $m$ is selected in the range of $[1,k)$. The following results are based on the optimal values of $k$ and $m$ obtained based on the validation set. The random splitting of the training/validation/testing sets is repeated $10$ times and the average error rates, as well as the standard derivation on the testing set are reported in Table \ref{tab:aggregate_real_experiments}. {In \citep{shalev2016minimizing}, the authors introduce slack variables to indicate outliers and propose a robust version of the maximum loss. We term it as Robust\_Max loss and compare it to our method as one of the baselines}. As these results show, compared  to the maximum, {Robust\_Max}, average, and AT$_k$ losses, the \aorr~loss achieves the best performance on all five data sets with both individual logistic loss and individual hinge loss. For individual logistic  loss, the \aorr~loss significantly improves the classification performance on {\tt Monk} and {\tt Phoneme} data sets and a slight improvement on data sets {\tt Titanic} and {\tt Splice}. More specifically, the performance of maximum aggregate loss is very poor in all cases due to its high sensitivity to outliers or noisy data. {The optimization of the Robust\_Max loss uses convex relaxation on the domain of slack variables constraint and using $l_2$ norm to replace the $l_1$ norm in the constraint. Therefore, it can alleviate the sensitivity to outliers, but cannot exclude their influence.}  The average aggregate loss is more robust to noise and outliers than the maximum loss { and the Robust\_Max loss} on all data sets. However, as data distributions may be very complicated, the average loss may sacrifice samples from rare distributions to pursue a lower loss on the whole training set and obtains sub-optimal solutions accordingly. The AT$_k$ loss is not completely free from the influence of outliers and noisy data either, which can be observed in particular on the Monk data set. On the Monk data set, {in comparison to the AT$_k$ loss, the \aorr~loss reduce the misclassification rates by $4.07\%$ for the individual logistic loss  and $3.87\%$ for the individual hinge loss, respectively.} 

\begin{figure*}[t!]
\captionsetup[subfigure]{justification=centering}
\centering
        \begin{subfigure}[b]{0.256\textwidth}
                \includegraphics[width=\linewidth]{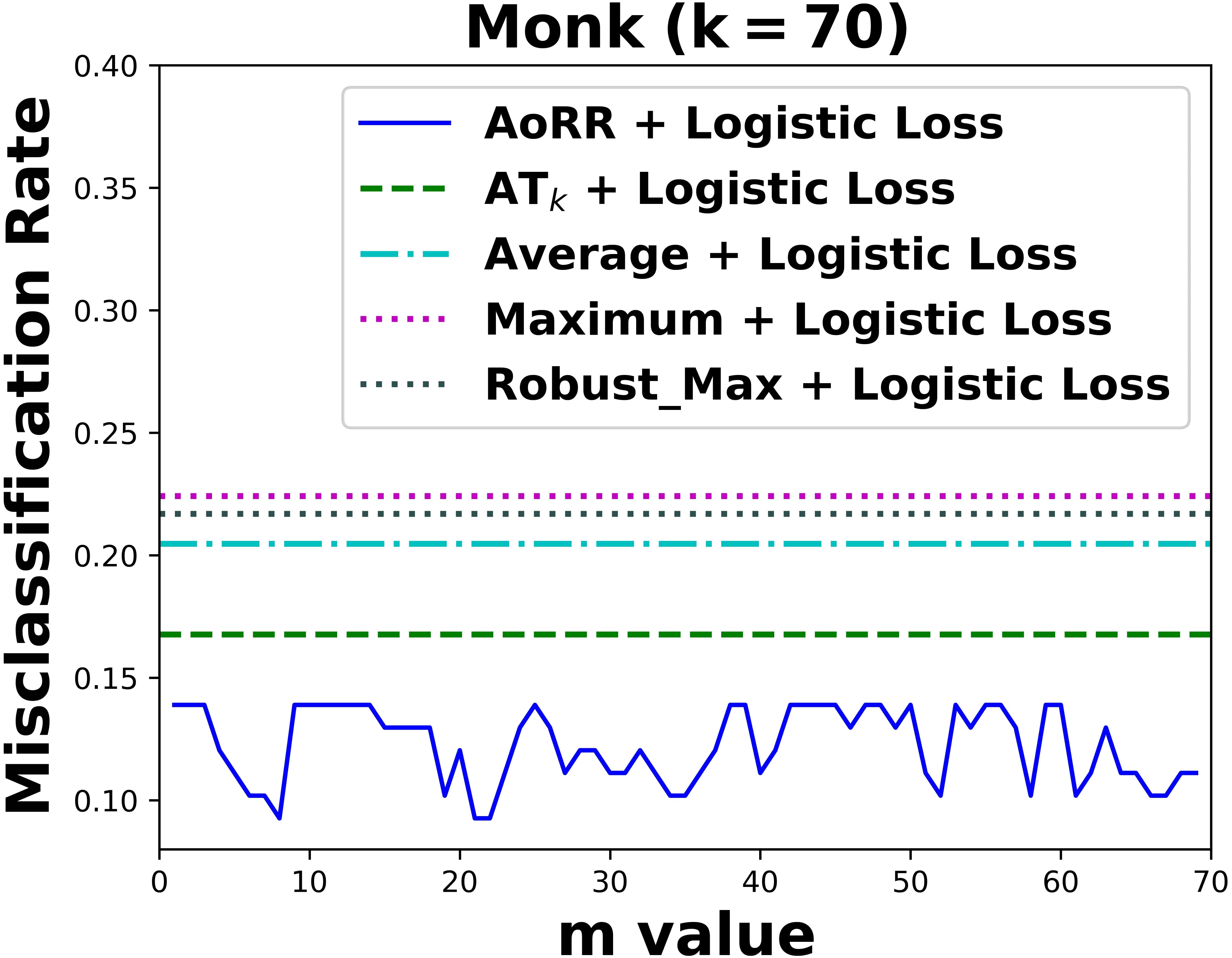}
        \end{subfigure}%
        \begin{subfigure}[b]{0.243\textwidth}
                \includegraphics[width=\linewidth]{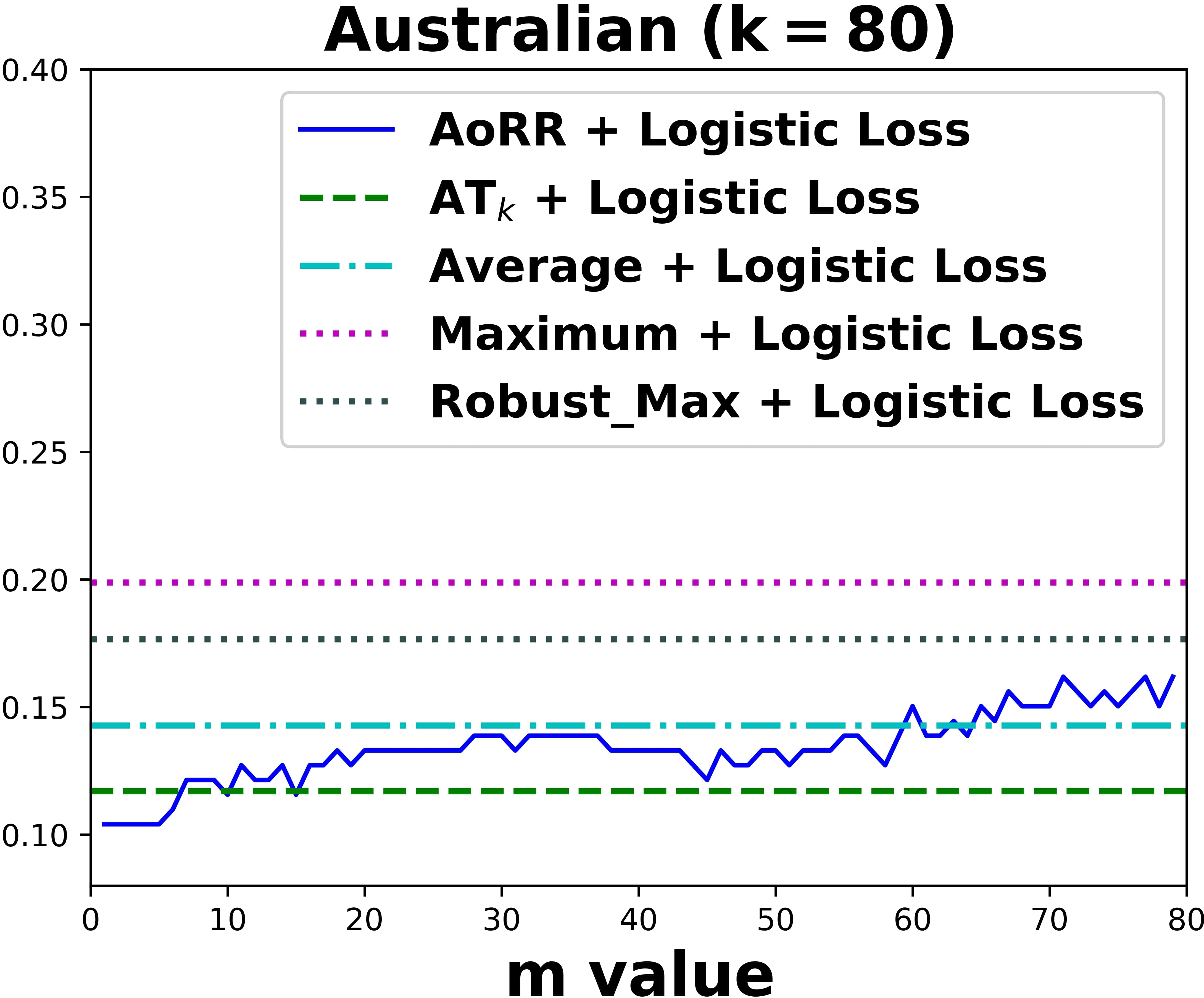}
        \end{subfigure}%
        \begin{subfigure}[b]{0.245\textwidth}
                \includegraphics[width=\linewidth]{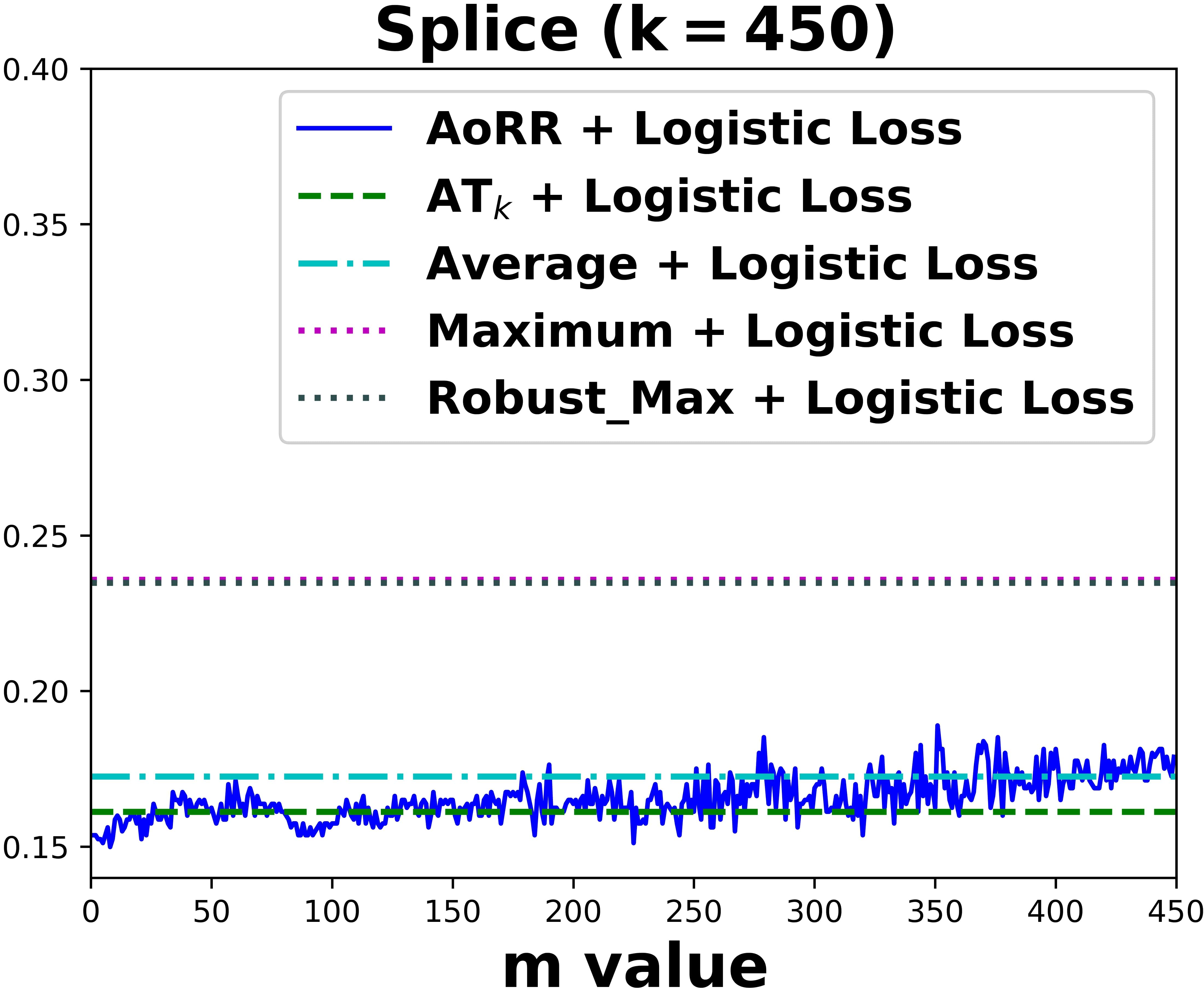}
        \end{subfigure}%
        \begin{subfigure}[b]{0.251\textwidth}
                \includegraphics[width=\linewidth]{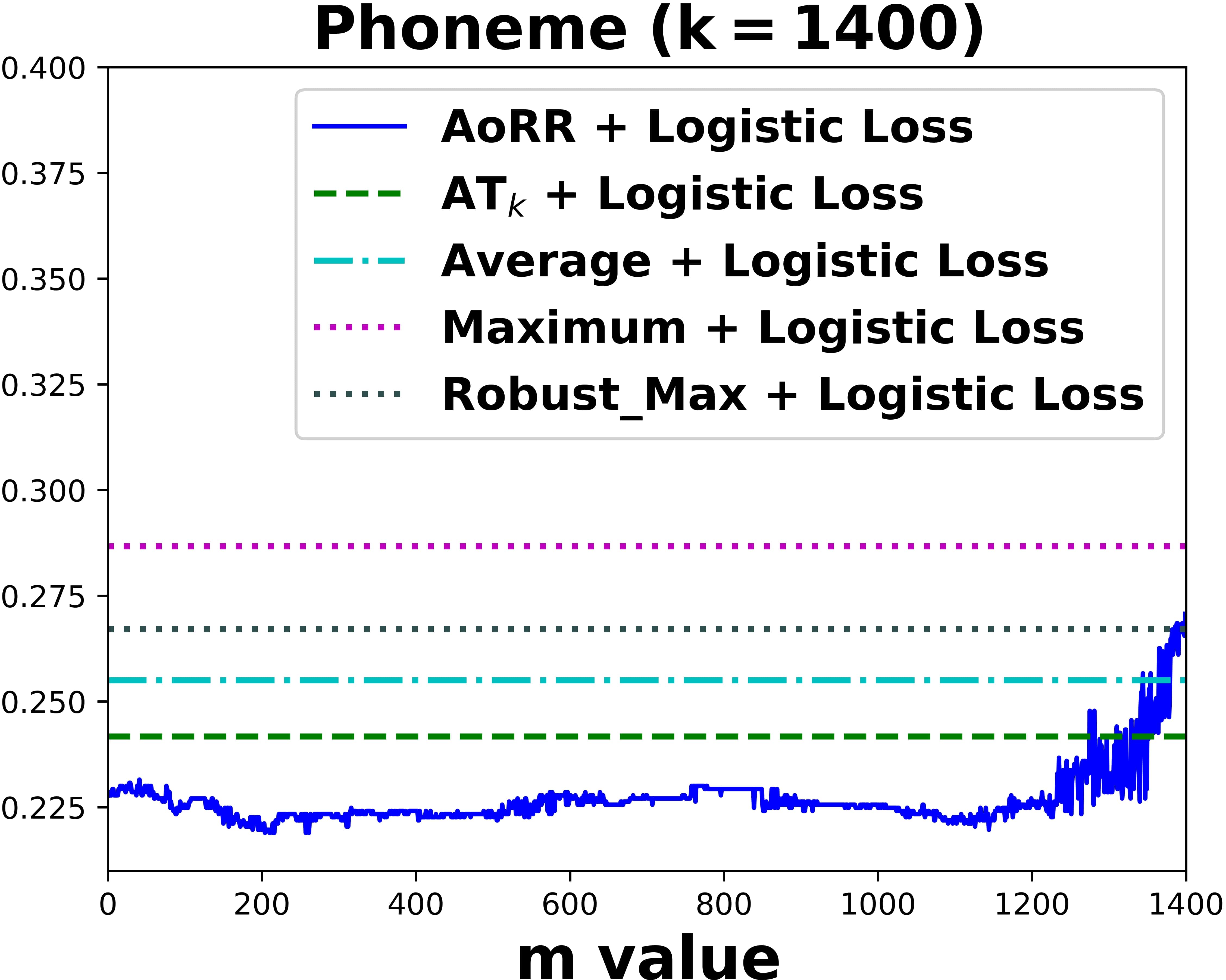}
        \end{subfigure}
        
        \begin{subfigure}[b]{0.256\textwidth}
                \includegraphics[width=\linewidth]{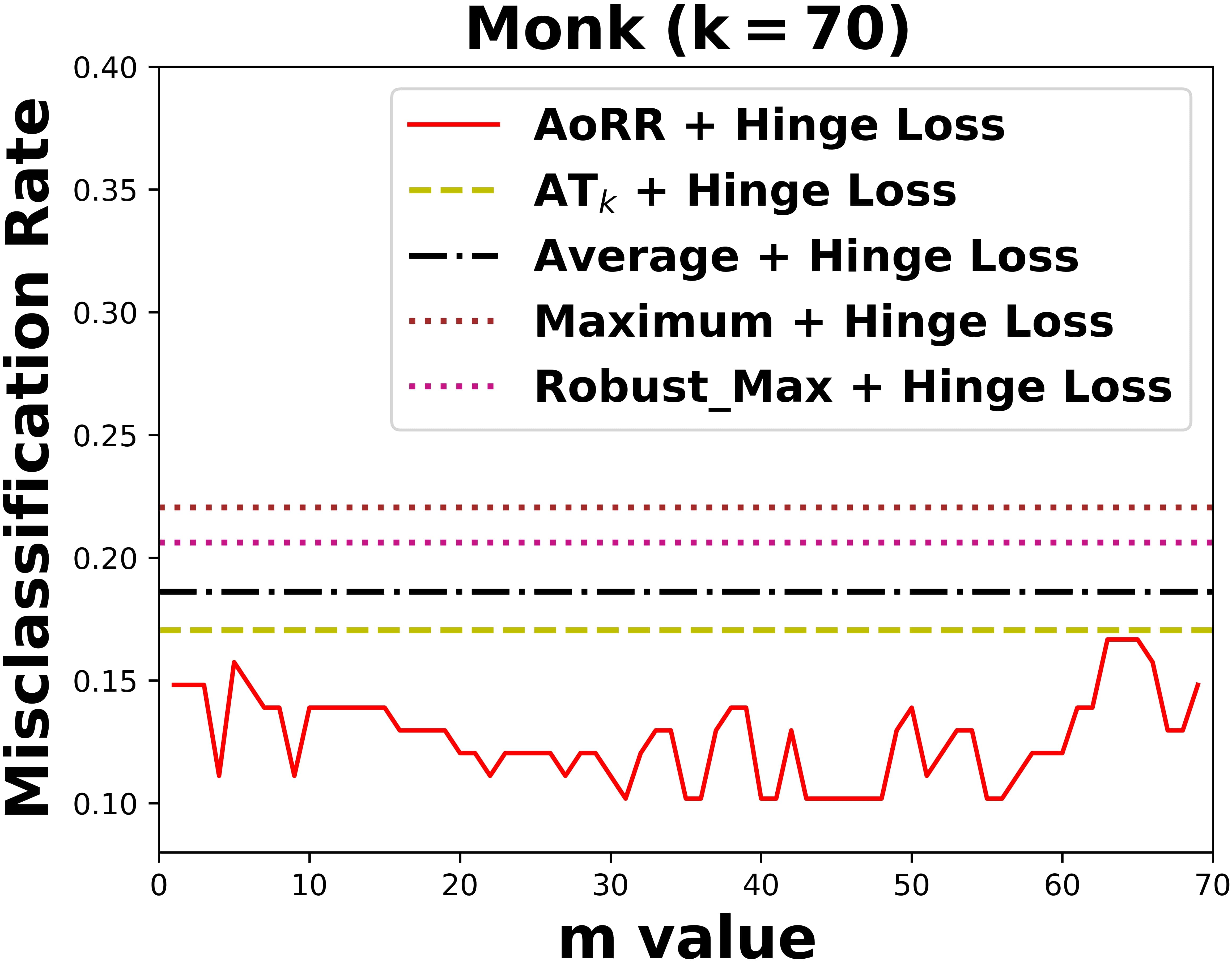}
        \end{subfigure}%
        \begin{subfigure}[b]{0.243\textwidth}
                \includegraphics[width=\linewidth]{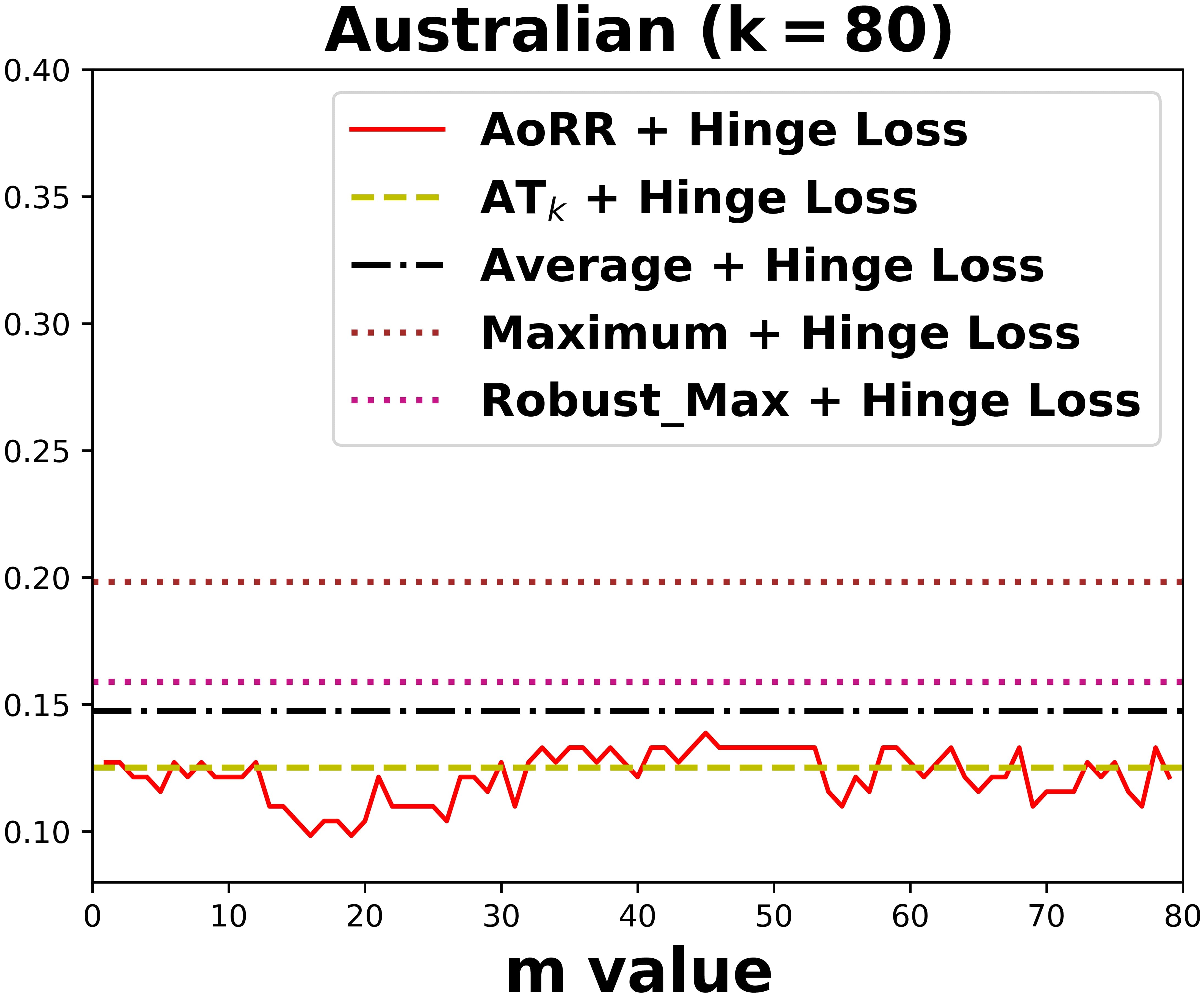}
        \end{subfigure}%
        \begin{subfigure}[b]{0.245\textwidth}
                \includegraphics[width=\linewidth]{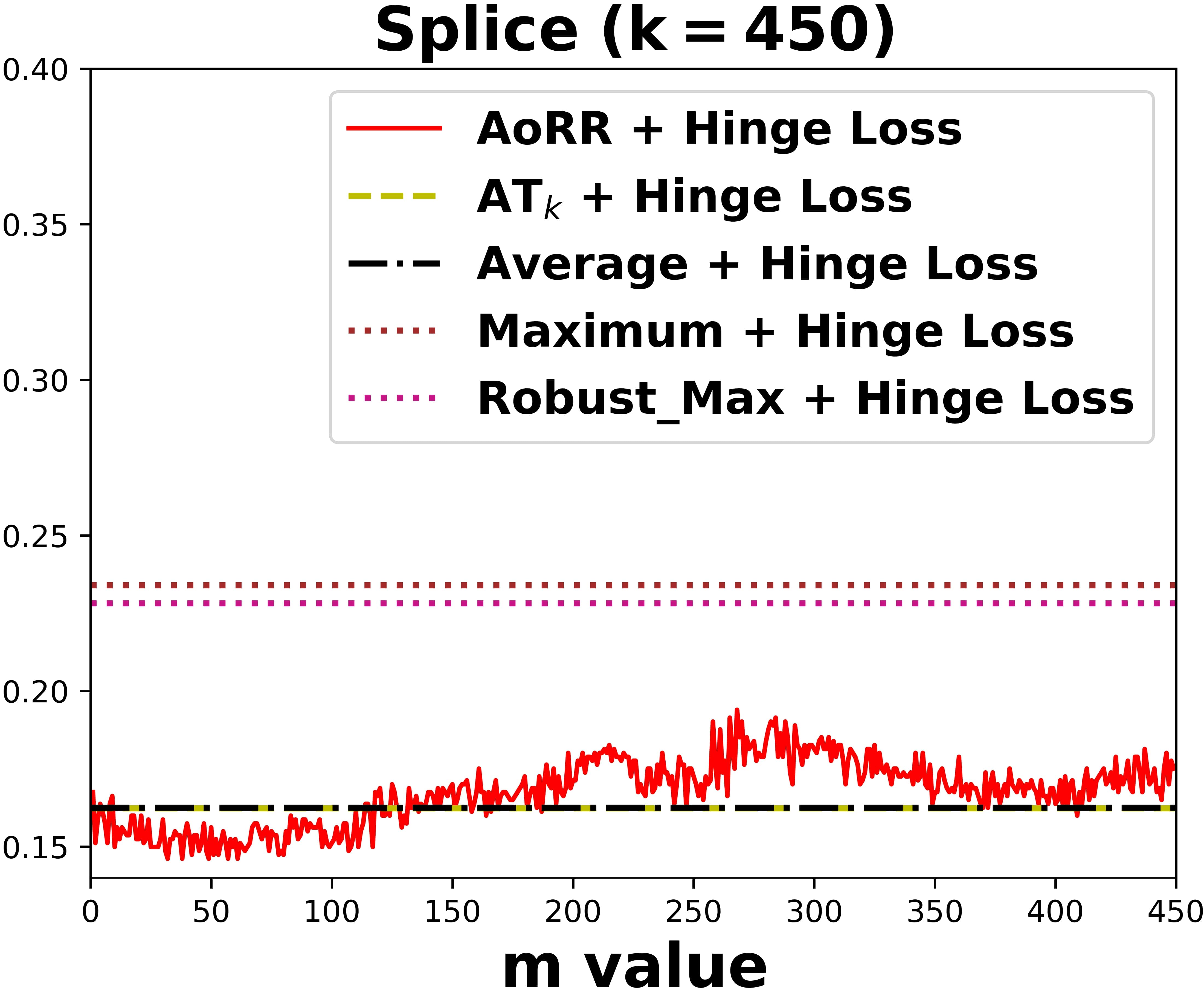}
        \end{subfigure}%
        \begin{subfigure}[b]{0.251\textwidth}
                \includegraphics[width=\linewidth]{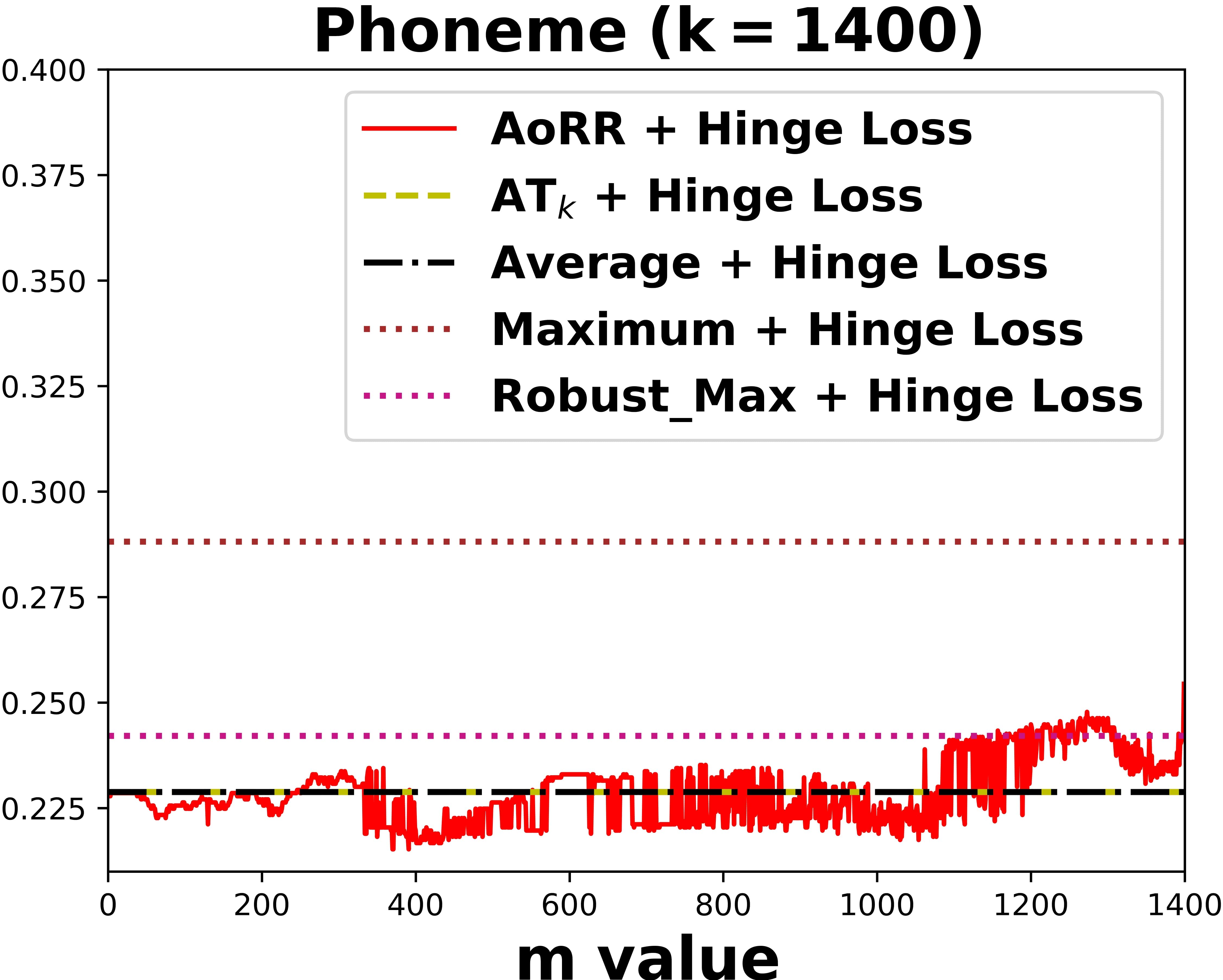}
        \end{subfigure}
        \caption{ Tendency curves of error rate of learning AoRR loss w.r.t. $m$ on four data sets.}\label{fig: aggregate_k_prime_value}
\end{figure*}

To further compare with the AT$_k$ loss, we investigate the influence of $m$ in the \aorr~loss. Specifically, we select the best $k$ value based on the AT$_k$ results and vary $m$ in the range of $[1,k-1]$. We use the individual logistic loss and plot tendency curves of misclassification error rates w.r.t $m$ in the first row of Figure  \ref{fig: aggregate_k_prime_value}, together with those from the average, maximum {and Robust\_Max} losses. As these plots show, on all four data sets, there is a clear range of $m$ with better performance than the corresponding AT$_k$ loss. We observe a trend of decreasing error rates with $m$ increasing. This is because outliers correspond to large individual losses, and excluding them from the training loss helps improve the overall performance of the learned classifier. However, when $m$ becomes large, the classification performance is decreasing, as many samples with small losses are included in the \aorr~objective and dominate the training process. Similar results based on the hinge loss can be found in the second row of Figure \ref{fig: aggregate_k_prime_value}.  
\subsubsection{\aorr~for Multi-class Classification} \label{sec:aorr_multiclass}


The \aorr~aggregate loss can also be extended to multi-class classification. Let $\mathcal{Y}=\{1,\cdots, l\}$, where $l$ represents the number of classes. We consider softmax loss as an individual loss. For sample $(x_i,y_i)$, it is defined as 

\begin{equation*}
    \begin{aligned}
    s_i(\theta) = -\log\Big(\frac{\exp(f_{y_i}(x_i;\theta))}{\sum_{j=1}^l \exp(f_j(x_i;\theta))}\Big),
    \end{aligned}
\end{equation*}
where $f_j(x_i;\theta)$ is the prediction of $j$-th class, and $y_i\in\mathcal{Y}$ is the ground-truth label of $x_i$. 

We use this multi-class classification task to verify the effectiveness of Algorithm \ref{Alg1}. We conduct experiments on the MNIST data set \citep{lecun1998gradient}, which contains $60,000$ training samples and $10,000$ testing samples that are images of handwritten digits. To create a validation set, We randomly extract $10,000$ samples from training samples. 
Therefore, the remaining training data size is $50,000$. To simulate outliers caused by errors that occurred when labeling the data in the training set, as in the work of \cite{wang2019symmetric}, we use the symmetric (uniform) noise creation method to randomly change labels of the training samples to one of the other class labels with a given proportion. The label is chosen at random with probability $p=0.2,0.3,0.4$. We use the average loss and AT$_k$ loss as baselines. For AT$_k$ loss, we also apply the same ``warm-up'' procedure and the same operation 2 to determine the hyper-parameter $k$. We extract the mean value plus one standard deviation of all individual losses from the validation set as operation 1's result. For operation 2, we use the mean value minus two standard deviations of all individual losses from the validation set. 
After multiple training epochs in the ``warm-up'' procedure, if the validation loss does not decrease, we stop the ``warm-up'' procedure. The result of this procedure can also be regarded as the result by using the average loss. After ``warm-up'', 
we do the training for AT$_k$ loss and \aorr~loss. 

\begin{table}[t]
\centering
\scalebox{0.8} {
\begin{tabular}{|c|c|c|c|}
\hline
\diagbox{Methods}{Noise Level}& 0.2 & 0.3 & 0.4 \\ \hline
Average & 89.69 (0.08) & 88.71 (0.10) & 87.77 (0.02) \\ \hline
AT$_k$ & 89.71 (0.07) & 88.73 (0.11) & 87.62 (0.04) \\ \hline
\aorr & \textbf{92.42 (0.03)} & \textbf{92.26 (0.07)} & \textbf{91.87 (0.11)} \\ \hline
\end{tabular}
}
\vspace{1mm}
\caption{ Testing accuracy (\%) of different aggregate losses combined with individual logistic loss on MNIST with different levels of symmetric noisy labels. The average accuracy and standard deviation of 5 random runs are reported and the best results are shown in bold.}
\label{tab:learning_hyperparameter_framework}
\end{table}

\begin{table}[t]
\centering
\scalebox{0.8} {
\begin{tabular}{|c|c|c|c|}
\hline
\diagbox{$m$}{Noise Level}& 0.2 & 0.3 & 0.4 \\ \hline
Ground Truth & 10000 & 15000 & 20000 \\ \hline
Estimation & 11722 (23) & 16431 (17) & 21026 (24) \\ \hline
\end{tabular}
}
\vspace{1mm}
\caption{ The estimation of hyper-parameter $m$. The average and standard deviation of 5 random runs are reported.}
\label{tab:m_estimation}
\end{table}

All models randomly run 5 times. The performance of testing accuracy is reported in Table \ref{tab:learning_hyperparameter_framework}. From Table \ref{tab:learning_hyperparameter_framework}, we can find that \aorr~outperforms all other two baseline methods for all different noise levels. Specifically, our \aorr~method obtains 2.71\% improvement on $p=0.2$, 3.53\% improvement on $p=0.3$, and 4.25\% improvement on $p=0.4$ when comparing to AT$_k$. It is obvious that Algorithm \ref{Alg1} works successfully and effectively. In addition, we also compare the estimated $m$ value based on $\hat{\lambda}$ to the optimal $m$ because we know the number of noisy data in the training set according to the noise level. From Table \ref{tab:m_estimation}, we can find all estimated values are close to the ground truth values. 
It should be noted that we can modify operation 2 to slightly increase the return value $\hat{\lambda}$ so that the difference between the estimated value and the ground truth will be smaller.

\section{\tkml~Loss for Multi-label Learning} \label{sec:tkml}

We use \sorr~to construct the individual loss for multi-label/multi-class classification, where a sample $x$ can be associated with a set of labels $\oldemptyset \not = Y \subset \{1,\cdots,l\}$. Our goal is to construct a linear predictor $f_{\Theta}(x) = \Theta^T x$ with $\Theta = (\theta_1,\cdots,\theta_l)$.  The final classifier outputs labels for $x$ with the top $k$ ($1 \le k < l$) prediction scores, \ie, $\theta^\top_{[1]}x \geq  \theta^\top_{[2]}x \geq \cdots \geq  \theta^\top_{[k]}x$. In training, the classifier is expected to include as many true labels as possible in the top $k$ outputs. This can be evaluated by the ``margin'', \ie, the difference between the $(k+1)$-th largest score of all the labels, $\theta^\top_{[k+1]}x$ and the lowest prediction score of all the ground-truth labels, $\min_{y \in Y}\theta^\top_{y}x$. If we have $\theta^\top_{[k+1]}x < \min_{y \in Y}\theta^\top_{y}x$, then all ground-truth labels have prediction scores ranked in the top $k$ positions. If this is not the case, then at least one ground-truth label has a prediction score not ranked in the top $k$. This induces the following metric for multi-label classification as $\mathbb{I}_{[\theta^\top_{[k+1]}x \ge \min_{y \in {Y}}\theta^\top_{y}x]}$. Replacing the indicator function with the hinge function and let $S(\theta)=\{s_{j}(\theta)\}_{j=1}^l$, where $s_{j}(\theta)=\big[1+ \theta^\top_{j}x - \min_{y \in {Y}}\theta^\top_{y}x\big]_+$, we obtain a continuous surrogate loss, as    $\psi_{k,k+1}(S(\theta))= s_{[k+1]}(\theta)$.
We term this loss as the {\em \underline{t}op-$\underline{k}$ \underline{m}ulti-\underline{l}abel} (\tkml) loss. According to the existing works from \cite{crammer2003family} and \cite{lapin2017analysis}, the conventional multi-label loss is defined as $\big[1+ \max_{y \not \in Y}\theta^\top_{y}x -   \min_{y \in {Y}}\theta^\top_{y}x\big]_+$. When $k = |Y|$, we have the following proposition and its proof can be found in Appendix \ref{appendix:proof_proposition1},
\begin{Proposition}\label{prop:tkml}
The \tkml~loss is a lower-bound to the conventional multi-label loss \citep{crammer2003family}, as
$ 
\big[1+ \max_{y \not \in Y}\theta^\top_{y}x -   \min_{y \in {Y}}\theta^\top_{y}x\big]_+ \ge \psi_{|Y|,|Y|+1}(S(\theta)).
$
\end{Proposition}
The \tkml~loss generalizes the conventional multi-class loss ($|Y| =k=1$) and the top-$k$ consistent $k$-guesses multi-class classification \citep{yang2020consistency} ($1 = |Y| \leq k < l$). A similar learning objective is proposed in \cite{lapin2015top} corresponds to $s_{[k]}(\theta)$, however, as proved in \cite{yang2020consistency}, it is not multi-class top-$k$ consistent. 
{Another work in  \cite{chang2017robust} proposes a robust top-k multi-class SVM based on the convex surrogate  of $s_{[k]}(\theta)$ to address the outliers by using a hyperparameter to cap the values of the individual losses. This approach is different from ours since we directly address the original top-k multi-class SVM problem using our \tkml~loss without introducing its convex surrogate and it is consistent.} For a set of training data $(x_1,Y_1),\cdots,(x_n,Y_n)$, if we denote $\psi_{k,k+1}(S(x,Y;\Theta)) = \psi_{k,k+1}(S(\theta))$, the data loss on \tkml~can be written as $\L_{\tkml}(\Theta)=\frac{1}{n}\sum_{i=1}^n \psi_{k,k+1}(S(x_i,Y_i;\Theta))$, which can be optimized using the Algorithm \ref{Alg0}.

\subsection{Experiments} \label{Sec:multi-label_experiment}
We use the same $\ell_2$ regularizer, $R(\theta)=\frac{1}{2C}||\theta||_2^2$ and cross-validate hyper-parameter $C$ in the range 10$^0$ to 10$^5$, extending it when the optimal value appears.

\begin{table*}[]
\centering
\scalebox{0.8} { 
\begin{tabular}{|c|cccc|}
\hline
        Data Sets   &  \#Samples   &    \#Features &  \#Labels   & $\overline{c}$  \\ \hline
        Emotions   &  593   & 72 & 6 & 1.81\\ 
        Scene      &  2,407 & 294 & 6 & 1.06\\ 
        Yeast      &  2,417 & 103 & 14 & 4.22\\ \hline
\end{tabular}
}
\vspace{1mm}
\caption{ Statistical information of each data set for multi-label learning, where $\overline{c}$ represents the average number of positive labels per instance.}
\label{tab:multi_label_data sets}
\end{table*}

\textbf{Multi-label classification.} We use three benchmark data sets (Emotions, Scene, and Yeast) from the KEEL data repository to verify the effectiveness of our \tkml~loss. 
More details about these three data sets can be found in Table \ref{tab:multi_label_data sets}. For comparison, we compare \tkml~with logistic regression (LR) model (i.e. minimize a surrogate hamming loss \citep{zhang2013review}), and a ranking based method (LSEP, \cite{li2017improving}). For these two baseline methods, we use a sigmoid operator on the linear predictor as $f_{\Theta}(x)=1/(1+\exp(-\Theta^Tx)).$  Since \tkml~is based on the value of $k$, we use five different $k$ values ($k\in\{1,2,3,4,5\}$) to evaluate the performance. For each data set, we randomly partition it to 50\%/25\%/25\% samples for training/validation/testing, respectively. This random partition is repeated 10 times, and the average performance on testing data is reported in Table \ref{tab:multilabel_experiments}. 
We use a metric (top $k$ multi-label accuracy) $\frac{1}{n}\sum_{i=1}^n \mathbb{I}_{[(Z_i\subseteq Y_i) \vee (Y_i\subseteq Z_i)]}$ to evaluate the performance, where $n$ is the size of the sample set. For instance, $(x_i, Y_i)$ with $Y_i$ be its ground-truth set, $f_\Theta(x_i) \in \mathbb{R}^l$ be its predicted scores, and $Z_i$ be a set of top $k$ predictions according to $f_\Theta(x_i)$. This metric reflects the performance of a classifier can get as many true labels as possible in the top $k$ range. More settings can be found in Appendix \ref{sec:multi_label_settings}. 

\begin{table}[t]
\centering
\scalebox{0.8} { 
\begin{tabular}{|c|c|c|c|c|c|c|}
\hline
                Data Sets  & Methods & $k$=1 & $k$=2 & $k$=3 & $k$=4 & $k$=5 \\ \hline
\multirow{3}{*}{Emotions} & LR & 73.54(3.98) & 57.48(3.35) & 73.20(4.69) & 86.60(3.02) & 96.46(1.71) \\ \cline{2-7} 
                  & LSEP & 72.18(4.56) & 55.85(3.37) & 72.18(3.74) & 85.58(2.92) & 95.85(1.07) \\ \cline{2-7} 
                  & \tkml & \textbf{76.80(2.66)} & \textbf{62.11(2.85)} & \textbf{77.62(2.81)} & \textbf{90.14(2.22)} & \textbf{96.94(0.63)} \\ \hline
\multirow{3}{*}{Scene} & LR & 73.2(0.57) & 85.31(0.47) & \textbf{94.79(0.79)} & \textbf{97.88(0.63)} & \textbf{99.7(0.30)} \\ \cline{2-7} 
                  & LSEP & 69.22(3.43) & 83.83(4.83) & 92.46(4.78) & 96.35(3.5) & 98.56(1.94) \\ \cline{2-7} 
                  & \tkml & \textbf{74.06(0.45)} & \textbf{85.36(0.79)} & 88.92(1.47) & 91.94(0.87) & 95.01(0.61) \\ \hline
\multirow{3}{*}{Yeast} & LR & \textbf{77.57(0.91)} & \textbf{70.59(1.16)} & \textbf{52.65(1.23)} & 43.26(1.16) & 43.49(1.33) \\ \cline{2-7} 
                  & LSEP & 75.5(1.03) & 66.84(2.9) & 49.72(1.26) & 41.90(1.91) & 43.01(1.02) \\ \cline{2-7} 
                  & \tkml & 76.94(0.49) & 67.19(2.79) & 45.41(0.71) & \textbf{43.47(1.06)} & \textbf{44.69(1.14)} \\ \hline
\end{tabular}
}
\vspace{1mm}
\caption{ Top $k$ multi-label accuracy with its standard derivation (\%) on three data sets. The best performance is shown in bold.}
\label{tab:multilabel_experiments}
\vspace{-4mm}
\end{table}

From Table \ref{tab:multilabel_experiments}, we note that the \tkml~loss in general improves the performance on the Emotions data set for all different $k$ values. These results illustrate the effectiveness of the \tkml~loss. More specifically, our \tkml~method obtains 4.63\% improvement on $k=2$ and 4.42\% improvement on $k=3$ when comparing to LR. This rate of improvement becomes higher (6.26\% improvement on $k=2$) when compare to LSEP. We also compare the performance of the method based on the \tkml~loss on different $k$ values. If we choose the value of $k$ close to the number of the ground-truth labels, the corresponding classification method outperforms the two baseline methods. For example, in the case of the Emotions data set, the average number of positive labels per instance is $1.81$, and our method based on the \tkml~loss achieves the best performance for $k=1,2$. As another example, the average number of true labels for the Yeast data set is 4.22, so the method based on the \tkml~loss achieves the best performance for $k=4,5$. 

\begin{table}[t]
\centering
\scalebox{0.8} {
\begin{tabular}{|c|c|c|c|}
\hline
\diagbox{Methods}{Data Sets}& Emotions & Scene & Yeast \\ \hline
LR & 74.85 & 71.6 & 73.56 \\ \hline
LSEP & 82.66 & 85.43 & 74.26 \\ \hline
\tkml & \textbf{84.82} & \textbf{86.38} & \textbf{74.32} \\ \hline
\end{tabular}
}
\caption{ AP (\%) results on three data sets. The best performance is shown in bold.}
\label{tab:ap_results}
\end{table}

We also adopt a widely used multi-label learning metric named average precision (AP) for performance evaluation. It is calculated by \citep{zhang2013review}
\[
\mbox{AP}=\frac{1}{n}\sum_{i=1}^n\frac{1}{|Y_i|}\sum_{j\in Y_i}\frac{|\{\tau \in Y_i| rank_f(x_i,\tau)<rank_f(x_i,j)\}|}{rank_f(x_i,j)},
\]
where $rank_f(x_i,j)$ returns the rank of $f_j(x_i)$ in descending according to $\{f_a(x_i)\}_{a=1}^l$.
From Table \ref{tab:ap_results}, we can find our \tkml~method outperforms the other two baseline approaches on all data sets. For the Emotions data set, the AP score of \tkml~is 2.16\% higher than the LSEP method and near 10\% higher than the LR. The performance is also slightly improved on Scene and Yeast data sets.  These results demonstrate the effectiveness of our \tkml~method.

\textbf{Robustness analysis.} As a special case of \aorr, the \tkml~loss exhibit similar robustness with regards to outliers, which can be elucidated with experiments in the multi-class setting (\ie, $k$=1 and $|Y|$=1). 
We conduct experiments on the MNIST data set \citep{lecun1998gradient}, which contains $60,000$ training samples and $10,000$ testing samples.
To simulate outliers caused by errors that occurred when labeling the data, as in the work of \cite{wang2019symmetric}, we use the asymmetric (class-dependent) noise creation method \citep{patrini2017making, zhang2018generalized} to randomly change labels of the training data (2$\rightarrow$7, 3$\rightarrow$8, 5$\leftrightarrow $6, and 7$\rightarrow$1) with a given proportion. The flipping label is chosen at random with probability $p = 0.2, 0.3, 0.4$. As a baseline, we use the top-$k$ multi-class SVM (SVM$_{\alpha}$) \citep{lapin2015top}. The performance is evaluated with the top 1, top 2, $\cdots$, top 5 accuracy on the testing samples. More details about the settings can be found in Appendix \ref{sec:tkml_multi_class_settings}.

\begin{table}[t]
\centering
\scalebox{0.8} {
\begin{tabular}{|c|c|c|c|c|c|c|}
\hline
           Noise Level & Methods & Top-1 Acc. & Top-2 Acc. & Top-3 Acc.& Top-4 Acc. & Top-5 Acc. \\ \hline
\multirow{2}{*}{0.2} & SVM$_\alpha$ & 78.33(0.18) & 90.66(0.29) & 95.12(0.2) & 97.28(0.09) & 98.49(0.1)  \\  
                  & \tkml~ & \textbf{83.06(0.94)} & \textbf{94.17(0.19)} & \textbf{97.24(0.13)} & \textbf{98.47(0.05)} & \textbf{99.22(0.01)} \\ \hline
\multirow{2}{*}{0.3} & SVM$_\alpha$ & 74.65(0.17) & 89.31(0.24) & 94.14(0.2) & 96.73(0.23) & 98.19(0.07)  \\ 
                  & \tkml~ & \textbf{80.13(1.24)} & \textbf{93.37(0.1)} & \textbf{96.81(0.22)} & \textbf{98.21(0.05)} & \textbf{99.08(0.05)} \\ \hline
\multirow{2}{*}{0.4} & SVM$_\alpha$ & 68.32(0.32) & 86.71(0.42) & 93.14(0.49) & 96.16(0.32) & 97.84(0.18) \\ 
                  & \tkml~ & \textbf{75(1.15)} & \textbf{92.41(0.14)} & \textbf{96.2(0.13)} & \textbf{97.95(0.1)} & \textbf{98.89(0.04)} \\ \hline
\end{tabular}
}
\caption{ Testing accuracy (\%) of two methods on MNIST with different levels of asymmetric noisy labels. The average accuracy and standard deviation of 5 random runs are reported and the best results are shown in bold. (Acc.: Accuracy)}
\label{tab:MSVM_real_experiments}
\end{table}

\begin{figure*}[t]
\captionsetup[subfigure]{justification=centering}
\centering
        \begin{subfigure}[b]{0.33\textwidth}
                \includegraphics[width=\linewidth]{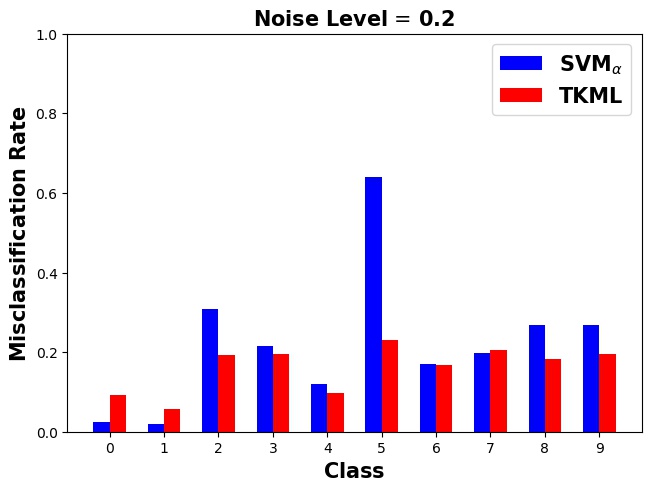}
                \label{fig:mnist_noise_20}
        \end{subfigure}%
        \begin{subfigure}[b]{0.33\textwidth}
                \includegraphics[width=\linewidth]{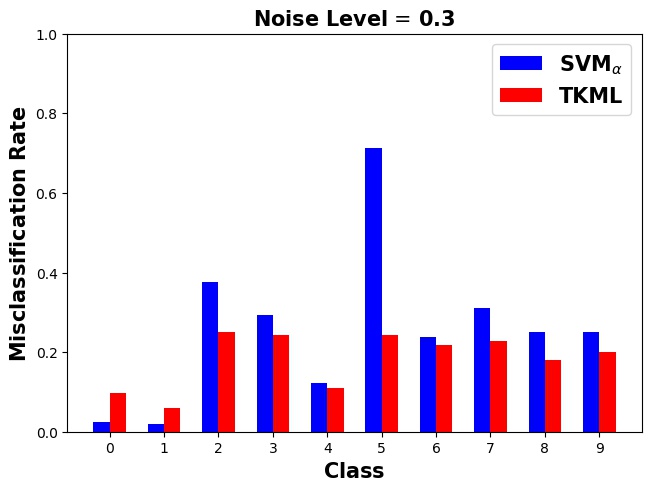}
                \label{fig:mnist_noise_30}
        \end{subfigure}%
        \begin{subfigure}[b]{0.33\textwidth}
                \includegraphics[width=\linewidth]{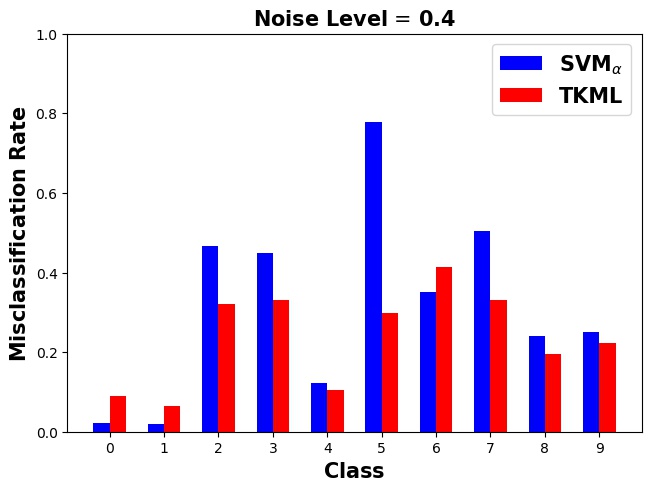}
                \label{fig:mnist_noise_40}
        \end{subfigure}%
        \caption{ The class-wise error rates of two methods with different noise level data.}\label{fig: MSVM_missclassification_error_by_class}
\vspace{-5mm}
\end{figure*}

From Table \ref{tab:MSVM_real_experiments}, it is clear that our method \tkml~consistently outperforms the baseline SVM$_{\alpha}$ among all top 1-5 accuracies. The gained improvement in performance is getting more significant as the level of noise increases. Since our flipping method only works between two different labels, we expected the  performance of \tkml~has some significant improvements on top 1 and 2 accuracies. Indeed, this  expectation is correctly verified as Table \ref{tab:MSVM_real_experiments} clearly indicates that the performance of our method is better than SVM$_\alpha$ by nearly 7\% accuracy (see Top-1 accuracy in the noise level 0.4). These results also demonstrate our optimization framework works well. 

To evaluate our method is better than SVM$_\alpha$ especially on the flipped class, we plot the class-wise error rate w.r.t different noise level data. As seen in Figure \ref{fig: MSVM_missclassification_error_by_class}, our method \tkml~outperforms SVM$_\alpha$ on the flipping classes such as 2 and 3, especially in class 5. As the noise level increases, the performance gap becomes more pronounced. For flipping class 7, the performance in this class is increased when the noise level increases from 0.3 to 0.4. \tkml~also gets good performance in class 6 when the noise level is 0.2 and 0.3.

\section{Combination of \aorr~and \tkml} \label{sec:tkml-aorr}

The \tkml~individual loss applies to the multi-label learning problem at the label level. However, it may also be the case that the training samples for a multi-label learning problem include outliers. It is thus a natural idea to combine the \tkml~individual loss at the label level and the \aorr~aggregate loss at the sample level to construct a more robust learning objective. 

To the best of our knowledge, considering robustness for multi-label learning at both the label and the sample level has not been extensively studied in the literature. The most relevant work is \cite{rawat2020doubly}, which considers robustness at both levels for multi-class classification (\ie, out of multiple class labels, only one is the true label). The doubly-stochastic mining method in \cite{rawat2020doubly} uses the AT$_k$ aggregate loss on the sample level and top-$k$ multi-class SVM loss on the label level. As we discussed before, AT$_k$ loss cannot eliminate the outliers and top-$k$ multi-class SVM does not satisfy the multi-class top-$k$ consistent. Furthermore, the proposed method from their work cannot deal with the top-$k$ multi-label learning problem where there are multiple true labels for each instance. 

We consider a learning objective that combines \aorr~and \tkml~for multi-label learning.
Theorem \ref{theorem:aorr-equal} tells us that we can get the same optimal solution of \sorr~problem (equation (\ref{eq:0})) by solving the problem of equation (\ref{eq:aorr_newform}). However, it is hard to deal with the cardinality constraint $\|q\|_0=n-m$ in equation (\ref{eq:aorr_newform}). Thus, we need to reformulate the equation (\ref{eq:aorr_newform}) in order to remove the cardinality constraint. Before we introduce the new formulation, we provide an useful Lemma as follows,
\begin{lemma}
\label{concave}
$\sum_{i=m+1}^n s_{[i]}$ is a concave function of the elements of $S$. Furthermore, we have $\sum_{i=m+1}^ns_{[i]} = \max_{\lambda\in\mathbb{R}}\bigg\{(n-m)\lambda-\sum_{i=1}^{n}[\lambda-s_i]_+\bigg\}$, of which $s_{[m]}$ is an optimum solution.
\end{lemma}
Proof can be found in Appendix \ref{proof_concave}. According to equation (\ref{eq:aorr_newform}), we can substitute the optimal $q^*$ to the constraint. Using Lemma \ref{concave}, we get 
\begin{equation*}
    \begin{aligned}
    &\min_{\lambda} (k-m)\lambda + \sum_{i=m+1}^{n}[[s(\theta)-\lambda]_+]_{[i]}\\
    =& \min_{\lambda} (k-m)\lambda + \max_{\hat{\lambda}}\Big\{(n-m)\hat{\lambda}-\sum_{i=1}^n [\hat{\lambda}-[s_i(\theta)-\lambda]_+]_+\Big\}.
    \end{aligned}
\end{equation*}
Finally, we obtain a new objective function that the combination of \tkml~and \aorr~as follows,
\begin{equation}
    \begin{aligned}
    \mathcal{L}_{\tkml-\aorr}(S(\theta))&={\psi_{m,k}(S(\theta)) \over k-m}  =\min_{\lambda}\max_{\hat{\lambda}} \  \lambda +\frac{n-m}{k-m}\hat{\lambda}-\frac{1}{k-m}\sum_{i=1}^n [\hat{\lambda}-[s_i(\theta)-\lambda]_+]_+\\
    &={1 \over k-m}\min_{\lambda}\max_{\hat{\lambda}} \  (k-m)\lambda -m\hat{\lambda}+\sum_{i=1}^n \hat{\lambda}-[\hat{\lambda}-[s_i(\theta)-\lambda]_+]_+.
    \end{aligned}
\label{eq:combine}
\end{equation}

For $s_i(\theta)$, we use \tkml~individual loss to replace it. So we have $s_i(\theta) = [1+\theta_{[k^{\prime}+1]}^\top x_i -\min_{y\in Y_i}\theta^\top_y x_i]_+$, where we use $k^{\prime}$ in the label level to distinguish $k$ in the sample level. Then, we develop a heuristic algorithm for learning $\mathcal{L}_{\tkml-\aorr}(S(\theta))$.  For a set of training data $\{(x_i,Y_i)\}_{i=1}^n$, where sample $x_i$ is associated with a set of labels $\oldemptyset \not = Y_i \subset \{1,\cdots,l\}$. The algorithm iteratively updates the parameters  $\theta$, $\lambda$ and $\hat{\lambda}$ based on the $\mathcal{L}_{\tkml-\aorr}(S(\theta))$ over training samples with the following steps:
 \begin{equation}
    \begin{aligned}
    &\theta^{(t+1)} = \theta^{(t)}- \eta_l\Big(\frac{1}{k-m}\sum_{i=1}^n\partial s_i(\theta^{(t)})\cdot \mathbb{I}_{[\hat{\lambda}^{(t)}>[s_i(\theta^{(t)})-\lambda^{(t)}]_+]} \cdot \mathbb{I}_{[s_i(\theta^{(t)})>\lambda^{(t)}]}\Big),\\
    &\lambda^{(t+1)} = \lambda^{(t)} - \eta_l\Big(1 -\frac{1}{k-m}\sum_{i=1}^n \mathbb{I}_{[\hat{\lambda}^{(t)}>[s_i(\theta^{(t)})-\lambda^{(t)}]_+]} \cdot \mathbb{I}_{[s_i(\theta^{(t)})>\lambda^{(t)}]}\Big),\\
    &\hat{\lambda}^{(t+1)} = \hat{\lambda}^{(t)} + \eta_l\Big(\frac{n-m}{k-m}-\frac{1}{k-m}\sum_{i=1}^n \mathbb{I}_{[\hat{\lambda}^{(t)}>[s_i(\theta^{(t)})-\lambda^{(t)}]_+]}\Big),
    \end{aligned}
\label{eq:sorr-dl-updaterule}
 \end{equation}
where $0\leq m< k\leq n$, $1 \leq k^{\prime}<l$, $\eta_l$ is the step size, and $\partial s_i(\theta^{(t)})$ denotes the (sub)gradient of $s_i(\theta^{(t)})$ with respect to $\theta^{(t)}$. Algorithm \ref{Alg2} describes the pseudo-code of the proposed heuristic algorithm. For a given loop size $|t|$ and training sample size $n$, the time complexity of Algorithm 3 is $O(|t|\cdot n)$.

\begin{algorithm}[t]
    \caption{Combination of \aorr~and \tkml}\label{Alg2}
    \SetAlgoLined
    \textbf{Initialization:} $\theta^{(0)}$, $\lambda^{(0)}$, $\hat{\lambda}^{(0)}$, $\eta_t$, $k^{\prime}$, $k$, and $m$ \\

    
    \For{$t=0,1,...$}{
    
    Compute $s_{i}(\theta^{(t)})=[1+\theta_{[k^{\prime}+1]}^{(t)\top} x_i -\min_{y\in Y_i} \theta_y^{(t)\top} x_i]_+$, $\forall i\in \{1,\cdots, n\}$ 
    
    Update parameters $\theta^{(t+1)}$, $\lambda^{(t+1)}$, and $\hat{\lambda}^{(t+1)}$ with equation (\ref{eq:sorr-dl-updaterule})

    }
\end{algorithm}


    
    

It should be mentioned that the mini-batch SGD method can be applied to Algorithm \ref{Alg2} when the training sample size is large. We can also use more complex non-linear models such as deep neural networks to replace the linear model in the \tkml~individual loss $s_i(\theta)$.
Furthermore, Algorithm \ref{Alg2} with mini-batch $B$ setting will become a general algorithm. It can be generalized to other types of existing algorithms for multi-class learning with setting different hyper-parameter values of $k^{\prime}$, $k$, and $m$. For example, if $|Y_i|=1=k^{\prime}, \forall i$, and  $m=0$, it becomes OSGD algorithm \citep{kawaguchiordered}, which updates the model parameter based on the average loss over the top-$k$ largest individual losses in the mini-batch. If $|Y_i|=1=k^{\prime}, \forall i$, $m=|B|-k$, and $k=|B|$, it becomes ITLM algorithm \citep{shen2019learning}, which selects the bottom-$k$ smallest individual losses in each mini-batch and then uses them to update the model parameter. If $|Y_i|=1=k^{\prime}, \forall i$, $m=|B|-1$, and $k=|B|$, our algorithm is MKL-SGD \citep{shah2020choosing}. This method selects a sample with the smallest individual loss in the mini-batch set, then use the gradient of this individual loss to update the model parameters. In addition, if $|Y_i|=1, \forall i$, and $m=0$, our Algorithm is similar to the doubly-stochastic mining method in \cite{rawat2020doubly}.

\subsection{Experiments}
\begin{table}[t]
\centering
\scalebox{0.8} { 
\begin{tabular}{|c|c|c|c|c|c|c|}
\hline
                Noise Level  & Methods & $k^{\prime}$=1 & $k^{\prime}$=2 & $k^{\prime}$=3 & $k^{\prime}$=4 & $k^{\prime}$=5 \\ \hline
\multirow{3}{*}{0} & \tkml-Average & 73.78 (1.59)&73.64 (1.53)&43.90 (2.99)&34.94 (3.78)&43.55 (0.96) \\ \cline{2-7} 
                  & \tkml-AT$_k$ & 74.38 (1.56)&73.84 (1.25)&49.44 (1.76)&43.16 (2.16)&45.99 (2.13) \\ \cline{2-7} 
                  & \tkml-\aorr & \textbf{74.94 (1.82)}&\textbf{73.88 (1.26)}&\textbf{50.25 (1.39)}&\textbf{45.93 (2.15)}&\textbf{46.13 (1.77)} \\ \hline
\multirow{3}{*}{0.1} & \tkml-Average & 73.08 (1.55)&73.24 (1.52)&43.80 (4.78)&34.22 (4.72)&43.48 (0.65) \\ \cline{2-7} 
                  & \tkml-AT$_k$ & 74.28 (1.36)&73.44 (1.34)&49.26 (0.80)&41.12 (1.91)&45.72 (1.97) \\ \cline{2-7} 
                  & \tkml-\aorr & \textbf{74.55 (1.62)}&\textbf{73.68 (1.52)}&\textbf{50.19 (0.81)}&\textbf{45.52 (2.35)}&\textbf{46.05 (2.20)} \\ \hline
\multirow{3}{*}{0.2} & \tkml-Average & 72.71 (3.68)&72.64 (0.88)&43.38 (2.00)&33.96 (4.31)&43.30 (1.09) \\ \cline{2-7} 
                  & \tkml-AT$_k$ & 73.88 (1.03)& 73.24 (1.25)& 49.15 (1.45)& 39.24 (1.15)& 45.62 (1.70) \\ \cline{2-7} 
                  & \tkml-\aorr & \textbf{74.36 (1.61)}& \textbf{73.50 (1.38)}&\textbf{ 49.73 (1.41)}& \textbf{45.48 (2.35)}& \textbf{46.00 (1.59)} \\ \hline
\multirow{3}{*}{0.3} & \tkml-Average & 71.32 (3.86)& 71.94 (1.06)& 43.16 (2.78)& 33.78 (3.03)& 41.44 (4.09) \\ \cline{2-7} 
                  & \tkml-AT$_k$ & 73.78 (1.09)& 73.04 (1.21)& 47.48 (1.90)& 36.90 (1.91)& 45.06 (2.01) \\ \cline{2-7} 
                  & \tkml-\aorr & \textbf{74.31 (1.40)}& \textbf{73.48 (1.36)}& \textbf{49.69 (1.07)}& \textbf{44.71 (2.11})& \textbf{45.79 (1.98)} \\ \hline
\end{tabular}
}
\vspace{1mm}
\caption{ Top $k$ multi-label accuracy and its standard derivation (\%) on the Yeast data set with different levels of symmetric noisy labels. The average best performance is shown in bold based on 10 random runs.}
\label{tab:combine_experiments}
\vspace{-4mm}
\end{table}

\begin{figure*}[t]
\captionsetup[subfigure]{justification=centering}
\centering
        \begin{subfigure}[b]{0.33\textwidth}
                \includegraphics[width=\linewidth]{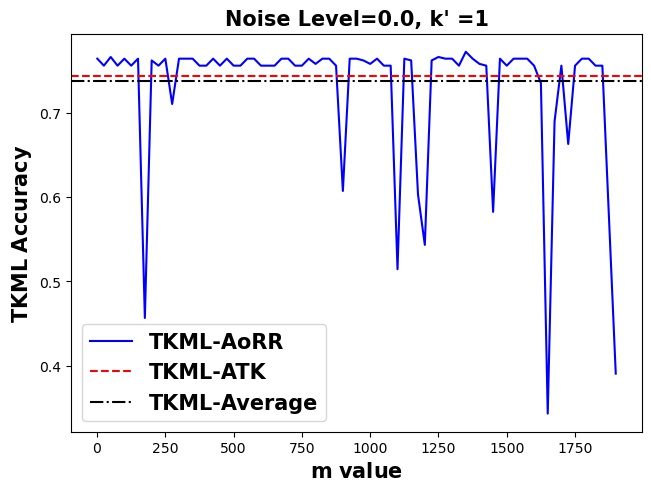}
                \label{fig:yeast_noise_0_k_1}
        \end{subfigure}%
        \begin{subfigure}[b]{0.33\textwidth}
                \includegraphics[width=\linewidth]{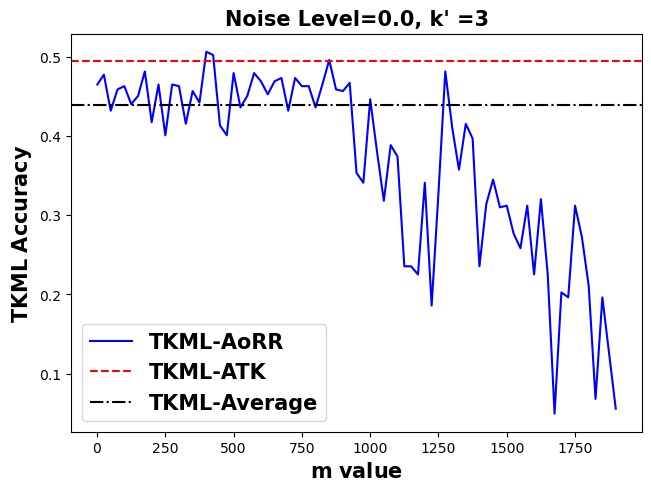}
                \label{fig:yeast_noise_0_k_3}
        \end{subfigure}%
        \begin{subfigure}[b]{0.33\textwidth}
                \includegraphics[width=\linewidth]{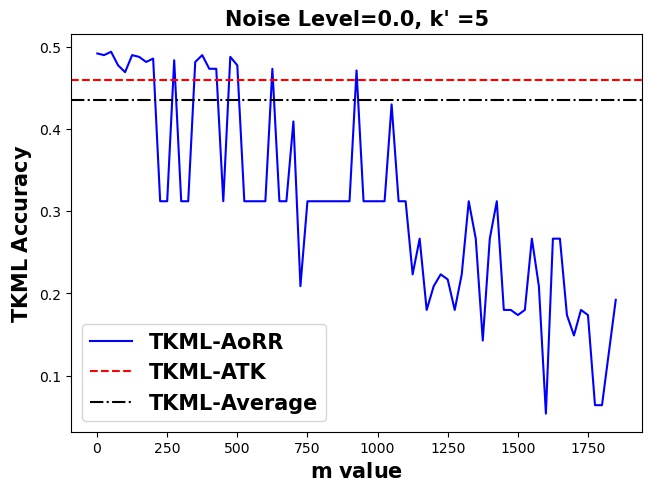}
                \label{fig:yeast_noise_0_k_5}
        \end{subfigure}%
        
        
        \begin{subfigure}[b]{0.33\textwidth}
                \includegraphics[width=\linewidth]{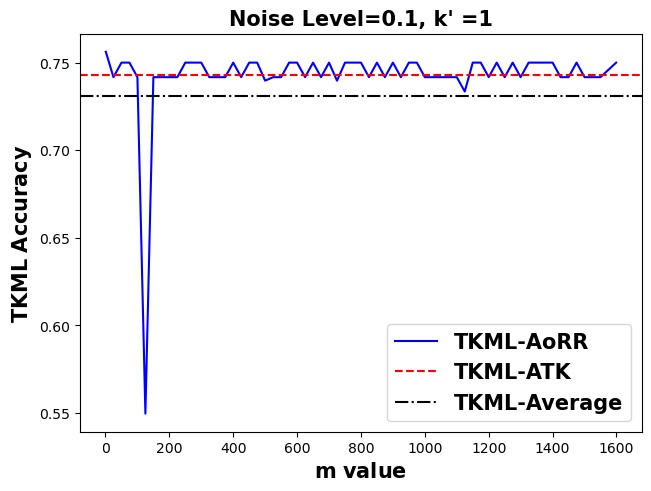}
                \label{fig:yeast_noise_10_k_1}
        \end{subfigure}%
        \begin{subfigure}[b]{0.33\textwidth}
                \includegraphics[width=\linewidth]{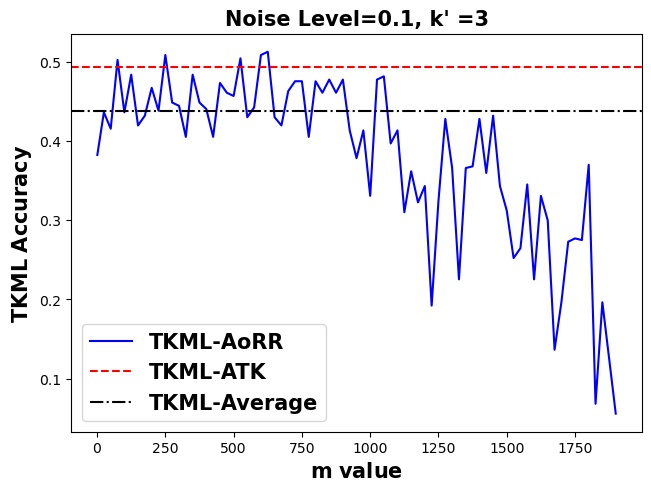}
                \label{fig:yeast_noise_10_k_3}
        \end{subfigure}%
        \begin{subfigure}[b]{0.33\textwidth}
                \includegraphics[width=\linewidth]{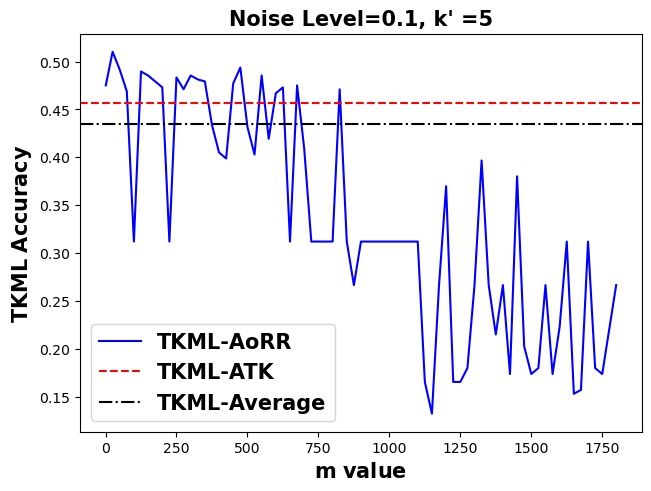}
                \label{fig:yeast_noise_10_k_5}
        \end{subfigure}%
        
        
        \begin{subfigure}[b]{0.33\textwidth}
                \includegraphics[width=\linewidth]{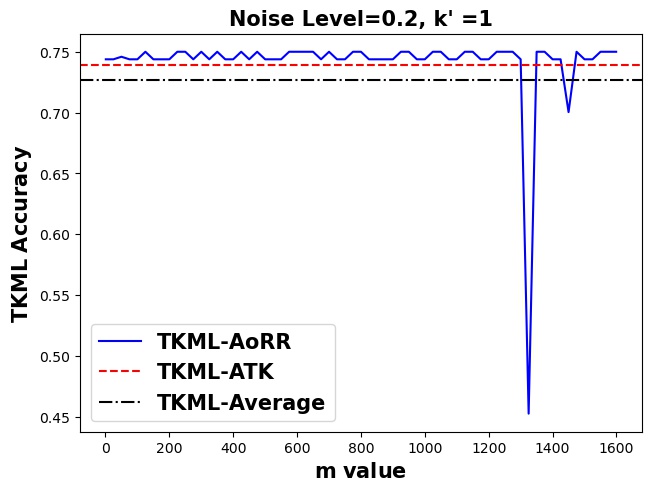}
                \label{fig:yeast_noise_20_k_1}
        \end{subfigure}%
        \begin{subfigure}[b]{0.33\textwidth}
                \includegraphics[width=\linewidth]{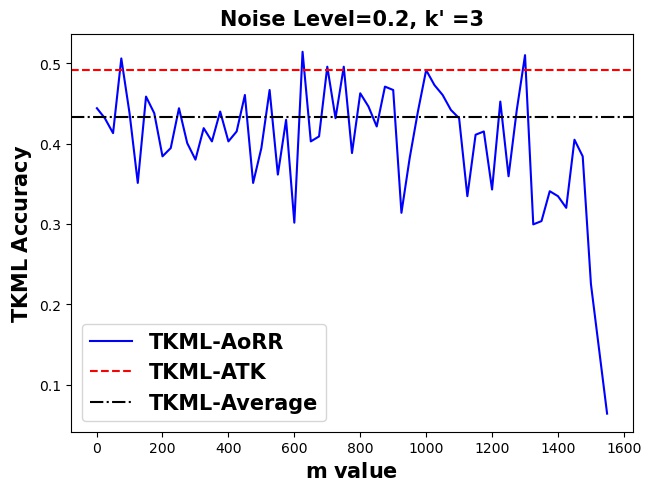}
                \label{fig:yeast_noise_20_k_3}
        \end{subfigure}%
        \begin{subfigure}[b]{0.33\textwidth}
                \includegraphics[width=\linewidth]{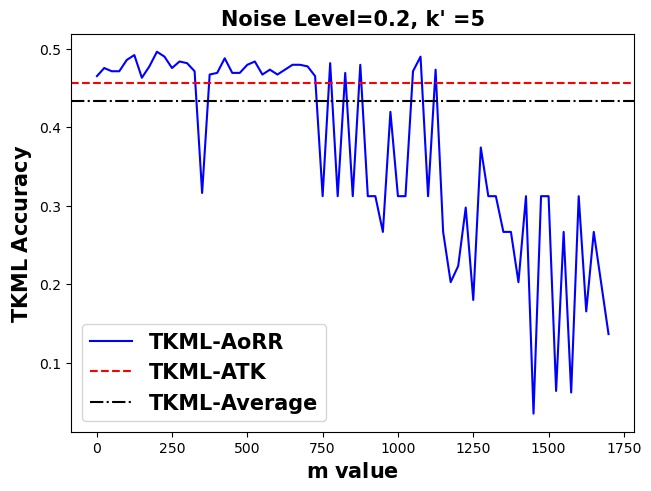}
                \label{fig:yeast_noise_20_k_5}
        \end{subfigure}%
        
        
        \begin{subfigure}[b]{0.33\textwidth}
                \includegraphics[width=\linewidth]{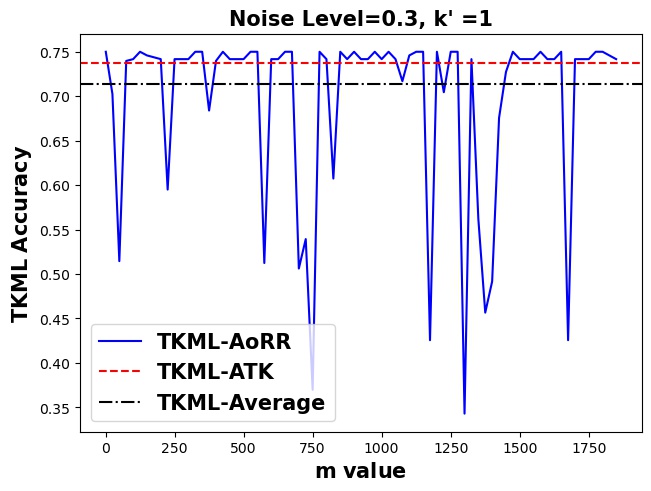}
                \label{fig:yeast_noise_30_k_1}
        \end{subfigure}%
        \begin{subfigure}[b]{0.33\textwidth}
                \includegraphics[width=\linewidth]{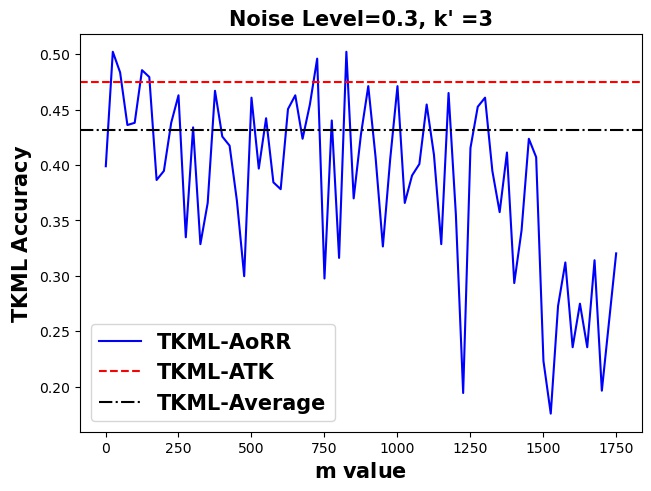}
                \label{fig:yeast_noise_30_k_3}
        \end{subfigure}%
        \begin{subfigure}[b]{0.33\textwidth}
                \includegraphics[width=\linewidth]{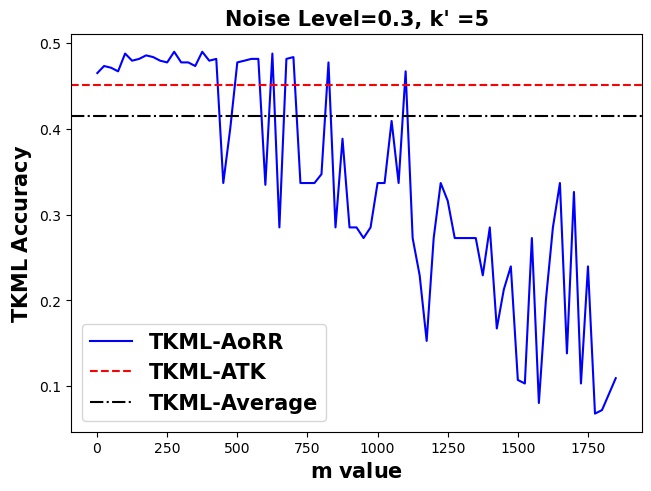}
                \label{fig:yeast_noise_30_k_5}
        \end{subfigure}%
        \caption{ Tendency curves of \tkml~accuracy of learning \tkml-\aorr~loss w.r.t. $m$ on four noise level settings ($p=0.0, 0.1, 0.2, 0.3$). The left figures are based on the $k'=1$ setting. The middle figures are based on the $k'=3$ setting. The right figures are based on the $k'=5$ setting.}\label{fig: combine_method_trend_m}
\vspace{-5mm}
\end{figure*}

The Yeast data set is selected to show the efficiency  of our loss function equation (\ref{eq:combine}) and Algorithm \ref{Alg2}. We randomly split Yeast data into two parts, which are 80\% samples for training and 20\% samples for testing. This random partition is repeated 10 times. The average performances on testing data are reported and they are based on the optimal values of hyper-parameters. To compare the combined \tkml~and \aorr~method (\tkml-\aorr), we consider other two combination methods, which use the average (\tkml-Average) and the AT$_k$ (\tkml-AT$_k$) aggregate loss on the sample level for top-$k$ multi-label learning.  Assuming the training data size is $n$, the hyper-parameter $k$ is selected in the range of $[1, n)$ for \tkml-AT$_k$ method. After finding the optimal value $k$ for \tkml-AT$_k$ method, we apply it to our \tkml-\aorr~method.  Then hyper-parameter $m$ in our method is selected in the range of $[1,k)$. Note that we can also apply the method from Section \ref{sec:determin_k_m} to determine the hyper-parameters $k$ and $m$ in practice. However, we use this grid search method to show the performance changes of our method as the hyper-parameters change in the experiment. On the other hand, since we use \aorr~aggregate method, we should expect that our method also exhibits  robustness with regards to outliers. Therefore, we introduce noise data to the training set of Yeast. Following \cite{li2020towards}, we generate symmetric label noise. Specifically, we randomly choose training samples with probability $p=0.1, 0.2, 0.3$ and change each of their labels to another random label. We use top-$k$ multi-label accuracy as a metric to evaluate the performance as we have applied in Section \ref{Sec:multi-label_experiment}.  All results are shown in the Table \ref{tab:combine_experiments}. More details about the settings can be found in Appendix \ref{sec:tkml_aorr_settings}.

From Table \ref{tab:combine_experiments}, we can find that our \tkml-\aorr~method outperforms the other two methods in all noise levels and all $k^{\prime}$ values. Especially, there is more than 10\% improvement when compared  with \tkml-Average method in $k^{\prime}=4$. The results show that our algorithm works efficiently. As the noise level increases, the top-$k$ multi-label accuracy of the same method is decreasing. The performance declines slowly with the increase of noise level while using our method. For example, from noise level $0$  to noise level $0.3$, the performance only has less than 1\% difference in $k^{\prime} = 1, 2, 3, 5$. For $k^{\prime} =4$, the gap is 1.22\% from noise level 0 to 0.3 using our method. However,  the performance is decreased near 10\% for the \tkml-AT$_k$ method. As we mentioned before, the performance of the \tkml-AT$_k$ method is better than the \tkml-Average method. But it cannot eliminate the influence of the outliers. Therefore, the performance can be further improved by using our method. To see how the performance change with setting different values of the hyper-parameter $m$, we show the tendency curves of \tkml~accuracy for the learning \tkml-\aorr~loss w.r.t. $m$ in Figure \ref{fig: combine_method_trend_m}. From Figure \ref{fig: combine_method_trend_m}, we can find there is a clear range of $m$ with better performance than the corresponding \tkml-Average and \tkml-AT$_k$ loss.

\section{Conclusion} \label{sec:conclusion}
In this work, we have introduced a general approach to form learning objectives, \ie, the sum of ranked range, which corresponds to the sum of a consecutive sequence of sorted values of a set of real numbers. We show that \sorr~can be expressed as the difference between two convex problems and optimized with the difference-of-convex algorithm (DCA). We also show that \sorr~can be reformed as a bilevel optimization problem.

We explored two applications in machine learning of the minimization of the \sorr~framework, namely the \aorr~aggregate loss for binary/multi-class classification at the data sample level and the \tkml~individual loss for multi-label/multi-class classification at the data label level. For the  \aorr~aggregate loss, we discussed its connection with  CVaRs and explore its generalization bound. We also proposed a method for learning hyper-parameters, which can make \aorr~works more efficiently in practice. Furthermore, we combine the \aorr~aggregate loss at the sample level and the \tkml~individual loss at the label level to enhance the robustness in the top-$k$ multi-label learning. A heuristic algorithm is proposed to optimize the combined loss. We conducted extensive experiments to show the effectiveness of the proposed frameworks on achieving superior generalization and robust performance on synthetic and real data sets. 


There are several potential \textbf{limitations} to our proposed methods. First, Algorithm \ref{Alg0} and Algorithm \ref{Alg1} are based on the DCA. We use a stochastic sub-gradient method to iterative optimize the model parameter in the inner loop, which may introduce bias about the optimal value of the learned model parameters $\theta$. Furthermore, using the inner loop to find the optimal solution in each outer loop may introduce a high time complexity. As we analyzed the time complexity of Algorithm \ref{Alg0} and Algorithm \ref{Alg1}, both of them are increased by increasing the size of the inner loop size $|l|$.  Second, in Algorithm \ref{Alg1}, we need a clean validation set to learn the hyper-parameters $k$ and $m$. However, it is very hard to guarantee the extracted validation set from the training set is clean. If it is not clean, the learned result through Algorithm \ref{Alg1} may not reliable. Third, although experiments show that Algorithm \ref{Alg2} is successful and effective, its theoretical convergence has not been studied. The learning objective (\ref{eq:combine}) is a minimax optimization problem, which is worth exploring its convergence property  and optimization error.

For \textbf{future works}, we will try to address the above-mentioned limitations of our proposed methods. We also plan to further study the statistical consistency of \tkml~loss for multi-label learning and the extension of our \sorr~framework with deep neural network structures. Testing the vulnerability of the deep learning models that incorporate the \tkml~loss is also worthy to explore. We have done such works in \cite{hu2021tkml}. However, how to protect \tkml~based models against adversarial attacks is important for future work.



\acks{This work is supported by NSF research grants (IIS-1816227,  IIS-2008532, DMS-2110836, and IIS-2110546) as well as an Army Research Office grant (agreement number: W911 NF-18-1-0297).}


\newpage

\appendix
\numberwithin{equation}{section}
\numberwithin{theorem}{section}
\renewcommand{\thesection}{{\Alph{section}}}
\renewcommand{\thesubsection}{\Alph{section}.\arabic{subsection}}
\renewcommand{\thesubsubsection}{\Roman{section}.\arabic{subsection}.\arabic{subsubsection}}
\setcounter{secnumdepth}{-1}
\setcounter{secnumdepth}{3}




\section{Proofs}

\subsection{Proof of Theorem \ref{theorem:sorr}}
\label{proof_theorem_sorr}
\begin{proof}
To prove Theorem 1, we need the following lemma.
\begin{lemma}[\cite{ogryczak2003minimizing}]. $\phi_k(S)$ is a convex function of the elements of $S$. Furthermore, for any $i\in [1,n]$, we have $\sum_{i=1}^k s_{[i]}=min_{\lambda\in \mathbb{R}}\{k\lambda+\sum_{i=1}^n[s_i-\lambda]_+\}$, of which $s_{[k]}$ is an optimum solution.
\label{lemma:convex}
\end{lemma}

From Lemma \ref{lemma:convex}, we have

\begin{equation*}
\begin{aligned}
\min_{\theta}\psi_{m,k}(S(\theta)) 
&= \min_{\theta} \bigl[ \phi_k(S(\theta)) -\phi_m(S(\theta))\bigr]               \\
&= \min_{\theta}\bigg[ \min_{\lambda\in \mathbb{R}}\Big\{k\lambda+\sum_{i=1}^n[s_i(\theta)-\lambda]_+\Big\}-\min_{\hat{\lambda}\in \mathbb{R}}\Big\{m\hat{\lambda}+\sum_{i=1}^n[s_i(\theta)-\hat{\lambda}]_+\Big\}\bigg].
\end{aligned}
\end{equation*}
If the optimal solution $\theta^*$ is achieved, from Lemma \ref{lemma:convex}, we get $\lambda = s_{[k]}$ and $\hat{\lambda}=s_{[m]}$. Therefore, $\hat{\lambda}> \lambda$ because $k>m$.  
\end{proof}

\subsection{Proof of Equation \eqref{eq:subgradient_phi_m}}
\label{prooftheorem1}
\begin{proof}
Before introducing the sub-gradient of $\phi_m(S(\theta))$, we provide a very useful characterization of differentiable properties of the optimal value function \cite[Proposition A.22]{bertsekas1971control}, which is also an extension of Danskin's theorem \citep{danskin2012theory}.

\begin{lemma}

Let $\phi:\mathbb{R}^n \times \mathbb{R}^m \rightarrow (-\infty, \infty]$ be a function and let $Y$ be a compact subset of $\mathbb{R}^m$. Assume further that for every vector $y\in Y$ the function $\phi(\cdot,y):\mathbb{R}^n \rightarrow (-\infty, \infty]$ is a closed proper convex function. Consider the function $f$ defined as $f(x)=sup_{y\in Y} \phi(x,y)$, then if $f$ is finite somewhere, it is a closed proper convex function. Furthermore, if int$(dom f) \neq \oldemptyset$ and $\phi$ is continuous on the set int$(dom f) \times Y$, then for every $x\in int(dom f)$ we have $\partial f(x)=conv\{\partial \phi(x,\overline{y})|\overline{y}\in\overline{Y}(x)\}$, where $\overline{Y}(x)$ is the set $\overline{Y}(x) = \{\overline{y}\in Y|\phi(x,\overline{y})=max_{y\in Y} \phi(x,y)\}$

\label{lemma:Bertsekas_lemma} 
\end{lemma}

We apply Lemma \ref{lemma:Bertsekas_lemma} with a new notation $\phi_m(\theta, \hat{\lambda}) = m\hat{\lambda}+\sum_{i=1}^n[s_i(\theta)-\hat{\lambda}]_+$. Suppose $\theta\in \mathbb{R}^n$ and $\hat{\lambda} \in \mathbb{R}$, the function $\phi_m: \mathbb{R}^n \times \mathbb{R} \rightarrow (-\infty, \infty]$. Let Y be a compact subset of $\mathbb{R}$ and for every $\hat{\lambda} \in Y$, it is obvious that the function $\phi_m(\cdot, \hat{\lambda}): \mathbb{R}^n \rightarrow (-\infty, \infty]$ is a closed proper convex function w.r.t $\theta$ from the second term of Eq.(\ref{eq:0}).

Consider a function $f$ defined as $f(\theta)=\sup_{\hat{\lambda}\in Y} \phi(\theta,\lambda)$, since $f$ is finite somewhere, it is a closed proper convex function. The interior of the effective domain of $f$ is nonempty, and that $\phi_m$ is continuous on the set $int(dom f) \times Y$. The condition of lemma \ref{lemma:Bertsekas_lemma} is satisfied. 

$\forall \theta\in int(dom f)$, we have
\[
\partial f(\theta)=conv\{\partial \phi_m(\theta,\overline{\lambda})|\overline{\lambda}\in\overline{Y}(\theta)\},
\]
where 
\[
\overline{Y}(\theta) = \{\overline{\lambda}\in Y|\phi_m(\theta,\overline{\lambda})=max_{\hat{\lambda}\in Y}\phi_m(\theta,\hat{\lambda})\}=\{\overline{\lambda}\in Y|-\phi_m(\theta,\overline{\lambda})=-min_{\hat{\lambda}\in Y}\phi_m(\theta,\hat{\lambda})\}.
\]
As we know $-min_{\hat{\lambda}\in Y}\phi_m(\theta,\hat{\lambda})= -\phi_m(S(\theta))$. This means the subdifferential of $f$ w.r.t $\theta$ exists when we set the optimal value of $\hat{\lambda}$.

From the above and the lemma \ref{lemma:convex}, we can get the sub-gradient $\hat{\theta} \in \partial \phi_m(S(\theta)) = \sum_{i=1}^n \partial s_i(\theta)\cdot \mathbb{I}_{[s_i(\theta)>\hat{\lambda}]}$, where $\hat{\lambda}$ equals to $s_{[m]}(\theta)$.
\end{proof}

\subsection{Proof of Theorem \ref{theorem:aorr-equal}}\label{appendix:proof_theorem_aorr_equal}
\begin{proof}
The proof of this proposition is similar to the proof of proposition 1 in the paper \cite{fujiwara2017dc}. However, they only discuss the linear case of $s_i(\theta)$. Under the constraints, we can rewrite the formula in Theorem \ref{theorem:aorr-equal}  as
\begin{equation*}
    \begin{aligned}
    (k-m)\lambda + \sum_{i=1}^n q_i[s_i(\theta)-\lambda]_+ &= (k-m)\lambda + \sum_{i=1}^n [s_i(\theta)-\lambda]_+ - \sum_{i=1}^n (1-q_i)[s_i(\theta)-\lambda]_+\\
    &=k\lambda + \sum_{i=1}^n [s_i(\theta)-\lambda]_+ - \sum_{i=1}^n (1-q_i)\{[s_i(\theta)-\lambda]_++\lambda\}.
    \end{aligned}
\end{equation*}
The last equality holds because $\sum_{i=1}^n(1-q_i)=n-(n-m)=m$. 

For the term $\sum_{i=1}^n (1-q_i)\{[s_i(\theta)-\lambda]_++\lambda\}$, we assume $s_i(\theta^*)$, $\forall i$, are sorted in descending order when getting the optimal model parameter $\theta^*$. For example, $s_1(\theta^*)\geq s_2(\theta^*)\geq \cdots \geq s_n(\theta^*)$. Since $\lambda^*\geq 0$, the optimal $q^*$ should be $q_1^*=\cdots=q_m^*=0$, $q_{m+1}^*=\cdots=q_n^*=1$. Note that $\lambda^*$ must be an optimal solution of the problem
\begin{equation*}
    \begin{aligned}
    \min_{\lambda} (k-m)\lambda + \sum_{i=m+1}^n q_i^*[s_i(\theta^*)-\lambda]_+. 
    \end{aligned}
\end{equation*}
From Lemma \ref{lemma:convex}, we know $s_{m+1}(\theta^*)\geq \lambda^*$, which implies that $s_i(\theta^*)-\lambda^*\geq 0$ holds for $q_i<1$. Therefore, $\sum_{i=1}^n (1-q_i)\{[s_i(\theta)-\lambda]_++\lambda\} = \sum_{i=1}^n (1-q_i)s_i(\theta)$. Furthermore, we know 
\begin{equation*}
    \begin{aligned}
    \min_{\hat{\lambda}}\Big\{m\hat{\lambda}+\sum_{i=1}^n[s_i(\theta)-\hat{\lambda}]_+\Big\} = \max_{q}\Big\{\sum_{i=1}^n (1-q_i)s_i(\theta)\Big|\ q_i\in [0,1], \ ||q||_0 =n-m\Big\}.
    \end{aligned}
\end{equation*}
Then we get 
\begin{equation*}
    \begin{aligned}
    &\min_{\lambda,q}(k-m)\lambda+\sum_{i=1}^n q_i[s_i(\theta)-\lambda]_+\ \ \ s.t. \ q_i\in [0,1], \ ||q||_0 =n-m \\
    =&\min_{\lambda}\Big\{k\lambda+\sum_{i=1}^n[s_i(\theta)-\lambda]_+\Big\}-\min_{\hat{\lambda}}\Big\{m\hat{\lambda}+\sum_{i=1}^n[s_i(\theta)-\hat{\lambda}]_+\Big\}.
    \end{aligned}
\end{equation*}
\end{proof}

\subsection{Proof of Theorem \ref{theorem:AR_calibration}}\label{appendix:proof_theorem_2}
\begin{proof} Without loss of generality,  by normalization we can assume $s(0)=1$ which can be satisfied by scaling. 
For any fixed $x\in \X$, by the definition of $f^* = \arg\inf \L(f, \gl^*, \hgl^*)$, we know that  
$$ f^*(x)= t^* = \arg\inf_{t\in \R}  \mathbb{E}\Bigl[[s(Yt)-\lambda^*]_+ - [s(Yt)-\hat{\lambda}^*]_+ \Big|  X = x \Bigr].$$
Notice the assumption $\hat{\lambda}^* >  \lambda^*$ and recall $\eta(x) = P(y=1|  x).$  We need to show that $t^* >0 $ for $\eta(x)>1/2$ and $t^*<0$ if $\eta(x)<1/2.$ Indeed, if $t^*\neq 0 $, then,  
 by the definition of $t^*$,
we have that 
\begin{align*} 
\mathbb{E}\Bigl[[s(Yt^*)-\lambda^*]_+ - [s(Yt^*)-\hat{\lambda^*}]_+ \Big|  X = x \Bigr] < \mathbb{E}\Bigl[[s(-Yt^*)-\lambda^*]_+ - [s(-Yt^*)-\hat{\lambda}^*]_+  \Big|  X = x \Bigr].     
\end{align*}
The above inequality  is identical to 
$$ \bigl[\big((s(t^*) -\lambda^*)_+ - (s(t^*) -\hat{\lambda}^*)_+\big)  - \big((s(-t^*) -\lambda^*)_+ - (s(-t^*) -\hat{\lambda}^*)_+\big) \bigr] \bigl[2\eta(x)- 1\bigr] <0.$$

Since $\hat{\lambda}^*> \lambda^*$, we have that  that $g(s) =  (s-\lambda^*)_+ - (s -\hat{\lambda}^*)_+ $ is a non-decreasing function of variable $s$. Then, if $\eta(x)> {1\over 2}$ we must have $g(s(t^*))<g(s(-t^*))$ which indicates $s(t^*) < s(-t^*).$ From the non-increasing property of $s$ on $\mathbb{R}$, $s(t)$ is also a convex function and $s'(0)<0$ immediately indicates $t^*>0$. Likewise, we can show that $t^*<0$ for $\eta(x)<1/2.$


To prove $t=0$ is not a minimizer, without loss of generality, assume $\eta(x)>\frac{1}{2}$. We need to consider two conditions as follows,

1. If $0\leq \lambda^* < \hat{\lambda}^*\leq 1$ and $s(0)=1$, then
\begin{align*}
A&=\mathbb{E}\Bigl[[s(0)-\lambda^*]_+ - [s(0)-\hat{\lambda}^*]_+ \Big|  X = x \Bigr] \\
&=[1-\lambda^*]_+ - [1-\hat{\lambda}^*]_+\\
&=\hat{\lambda}^* - \lambda^*.
\end{align*}
Since $s^{\prime}(0)<0$ and $s$ is non-increasing, there exists $t^0>t^*=0>-t^0$, and $s(-t^0)>s(0)\geq \hat{\lambda}^*>s(t^0)>\lambda^*$. Let
\begin{align*} 
B&=\mathbb{E}\Bigl[[s(Yt^0)-\lambda^*]_+ - [s(Yt^0)-\hat{\lambda}^*]_+ \big|  X= x \Bigr]\\
&=\Big([s(t^0)-\lambda^*]_+ - [s(t^0)-\hat{\lambda}^*]_+ \Big) \eta(x) +  \Big([s(-t^0)-\lambda^*]_+ - [s(-t^0)-\hat{\lambda}^*]_+ \Big) \Big(1-\eta(x)\Big)\\
&=\Big([s(-t^0)-\lambda^*]_+ - [s(-t^0)-\hat{\lambda}^*]_+ \Big)\\
&+\Bigl[\Big([s(t^0)-\lambda^*]_+ - [s(t^0)-\hat{\lambda}^*]_+ \Big)-\Big([s(-t^0)-\lambda^*]_+ - [s(-t^0)-\hat{\lambda}^*]_+ \Big)\Bigr]\eta(x)\\
&=\hat{\lambda}^* - \lambda^*+\Bigl[s(t^0)-\lambda^*-(\hat{\lambda}^* - \lambda^*)\Bigr]\eta(x).
\end{align*}
Then
\begin{align*}
B-A=(s(t^0)-\hat{\lambda}^*)\eta(x)<0.
\end{align*}
Therefore, $t=0$ is not a minimizer.

2. If $0\leq \lambda^* \leq 1 < \hat{\lambda}^*$ and $s(0)=1$, then
\begin{align*}
&\frac{d}{dt}\mathbb{E}[[s(Yt)-\lambda^*]_+ - [s(Yt)-\hat{\lambda}^*]_+]|_{t=0} \\
&=\frac{d}{dt}[\eta(x)([s(t)-\lambda^*]_+ - [s(t)-\hat{\lambda}^*]_+)+(1-\eta(x))([s(-t)-\lambda^*]_+ - [s(-t)-\hat{\lambda}^*]_+)]|_{t=0}\\
&=\frac{d}{dt}[\eta(x)(s(t)-\lambda^*)+(1-\eta(x))(s(-t)-\lambda^*)]|_{t=0}\\
&=[\eta(x)s^{\prime}(t)-(1-\eta(x))s^{\prime}(-t)]|_{t=0}\\
&=(2\eta(x)-1)s^{\prime}(0)<0.
\end{align*}
Thus $t = 0$ is not a minimizer. 
\end{proof}

\subsection{Proof of Theorem \ref{theorem:ICVaRs}}
\begin{proof}\label{proof:ICVaRs}
First, we introduce a lemma from \cite{brown2007large} and \cite[Theorem~1 and 2]{ thomas2019concentration} as follows,
\begin{lemma} (\cite{brown2007large, thomas2019concentration})
Let $C_{\alpha}[s]=\mathbb{E}[s|s\geq V_{\alpha}[s]]$ and $\widehat{C}_{\alpha}[s]:= \inf_{\lambda\in \mathbb{R}}\{\lambda+\frac{1}{n\alpha}\sum_{i=1}^n[s_i-\lambda]_+\}$, if $supp(s)\subseteq[a,b]$ and $s$ has a continuous distribution function, then for any $\delta\in(0,1]$,
\begin{equation*}
    \begin{aligned}
    &\Pr\left(C_{\alpha}[s] \leq \widehat{C}_{\alpha}[s] + (b-a)\sqrt{\frac{5\ln(3/\delta)}{\alpha n}}\right)\geq 1-\delta,\\
    &\Pr\left(C_{\alpha}[s] \geq \widehat{C}_{\alpha}[s] - \frac{b-a}{\alpha}\sqrt{\frac{\ln(1/\delta)}{2 n}}\right)\geq 1-\delta.
    \end{aligned}
\end{equation*}
\end{lemma}
This lemma tells us that bound the deviation of the empirical CVaR from the true CVaR with high probability. Based on this lemma, we can get 
\begin{equation}
    \begin{aligned}
    &\Pr\left(\nu(C_{\nu}[s(\theta)]-\widehat{C}_{\nu}[s(\theta)])\leq \nu(b-a)\sqrt{\frac{5\ln(3/\delta)}{\nu n}}\right)\geq 1-\delta,
    \end{aligned}
\label{eq:lemma_cvar_1}
\end{equation}
\begin{equation}
    \begin{aligned}
    \Pr\left(\nu(C_{\nu}[s(\theta)]-\widehat{C}_{\nu}[s(\theta)])\geq - (b-a)\sqrt{\frac{\ln(1/\delta)}{2 n}}\right)\geq 1-\delta,
    \end{aligned}
\label{eq:lemma_cvar_2}
\end{equation}
\begin{equation}
    \begin{aligned}
    &\Pr\left(-\mu(C_{\mu}[s(\theta)]-\widehat{C}_{\mu}[s(\theta)])\geq -\mu(b-a)\sqrt{\frac{5\ln(3/\delta)}{\mu n}}\right)\geq 1-\delta,
    \end{aligned}
\label{eq:lemma_cvar_3}
\end{equation}
\begin{equation}
    \begin{aligned}
    \Pr\left(-\mu(C_{\mu}[s(\theta)]-\widehat{C}_{\mu}[s(\theta)])\leq  (b-a)\sqrt{\frac{\ln(1/\delta)}{2 n}}\right)\geq 1-\delta.
    \end{aligned}
\label{eq:lemma_cvar_4}
\end{equation}
Combining Eq.(\ref{eq:lemma_cvar_1}) and Eq.(\ref{eq:lemma_cvar_4}), we obtain 
\begin{equation}
    \begin{aligned}
    \Pr\left(\mathcal{L}(f,\lambda, \hat{\lambda})-\widehat{\mathcal{L}}(f,\lambda, \hat{\lambda}) \leq(b-a)\left[\sqrt{\frac{5\nu\ln(3/\delta)}{n}}+\sqrt{\frac{\ln(1/\delta)}{2n}}\right]\right)\geq 1-\delta.
    \end{aligned}
\label{eq:lemma_cvar_5}
\end{equation}
Combining Eq.(\ref{eq:lemma_cvar_2}) and Eq.(\ref{eq:lemma_cvar_3}), we obtain 
\begin{equation}
    \begin{aligned}
    \Pr\left(\mathcal{L}(f,\lambda, \hat{\lambda})-\widehat{\mathcal{L}}(f,\lambda, \hat{\lambda}) \geq -(b-a)\left[\sqrt{\frac{5\mu\ln(3/\delta)}{n}}+\sqrt{\frac{\ln(1/\delta)}{2n}}\right]\right)\geq 1-\delta.
    \end{aligned}
\label{eq:lemma_cvar_6}
\end{equation}
Substituting $\nu = k/n$ and $\mu = m/n$ into Eq.(\ref{eq:lemma_cvar_5}) and  Eq.(\ref{eq:lemma_cvar_6}), we get desired results. 
\end{proof}

\subsection{Proof of Proposition \ref{prop:tkml}}
\label{appendix:proof_proposition1}
\begin{proof}
We just need to prove that $\max_{y \not \in Y}\theta_{y}^\top x \ge \theta_{[|Y|+1]}^\top x$. If this is not the case, then for any label $y \not \in Y$, then its rank in the ranked list is no more than $|Y|+2$, then the sum of total number of such labels is not larger than $l - (|Y|+2) + 1 = l - |Y|-1$. And the total number of labels will be $|Y|+|\{y\not\in Y\}| \le l-1 \not = l$, which is a contradiction. 
\end{proof}

\subsection{Proof of Lemma \ref{concave}}
\begin{proof}\label{proof_concave}
\begin{equation*}
    \begin{aligned}
    \sum_{i=m+1}^ns_{[i]}&=\sum_{i=1}^{n}s_{i}-\sum_{i=1}^{m}s_{[i]}\\
    &=\sum_{i=1}^{n}s_{i} - \min_{\lambda}\bigg\{m\lambda+\sum_{i=1}^n[s_i-\lambda]_+\bigg\}\\
    &=-\min_{\lambda}\bigg\{-\sum_{i=1}^{n}(s_{i}-\lambda) -(n-m)\lambda +\sum_{i=1}^n[s_i-\lambda]_+\bigg\}\\
    &=-\min_{\lambda}\bigg\{-(n-m)\lambda+\sum_{i=1}^{n}[\lambda-s_i]_+\bigg\}\\
    &=\max_{\lambda}\bigg\{(n-m)\lambda-\sum_{i=1}^{n}[\lambda-s_i]_+\bigg\}
    \end{aligned}.
\label{eq:non-targeted-atk}
\end{equation*}
The second equation holds because of Lemma \ref{lemma:convex}. The fourth equation holds because the fact of $[a]_+-a=[-a]_+$.
Define $L(\lambda) = (n-m)\lambda-\sum_{i=1}^{n}[\lambda-s_i]_+$. Let $\lambda = \alpha\lambda_1+(1-\alpha)\lambda_2$, where $1\geq\alpha\geq0$, we have 
\begin{equation*}
    \begin{aligned}
    &L(\alpha\lambda_1+(1-\alpha)\lambda_2) \\
    &= \alpha (n-m)\lambda_1+(1-\alpha)(n-m)\lambda_2
    -\sum_{i=1}^n [\alpha\lambda_1+(1-\alpha)\lambda_2-s_i]_+
    \\
    &=\alpha(n-m)\lambda_1+(1-\alpha)(n-m)\lambda_2
    -\sum_{i=1}^m [\alpha(\lambda_1-s_i)+(1-\alpha)(\lambda_2-s_i)]_+\\
    &\geq \alpha(n-m)\lambda_1+(1-\alpha)(n-m)\lambda_2-\alpha\sum_{i=1}^n[\lambda_1-s_i]_+-(1-\alpha)\sum_{i=1}^n[\lambda_2-s_i]_+\\
    &= \alpha L(\lambda_1)+(1-\alpha)L(\lambda_2).
    \end{aligned}
\end{equation*}
Therefore, $\sum_{i=m+1}^n s_{[i]}$ is a concave function.
\end{proof}

\section{Additional Experimental Details}




\subsection{Training Settings on Toy Examples for Aggregate Loss}
To reproduce the experimental results of \aorr~on synthetic data, we provide the details about the settings when we are training the model in Table \ref{tab:aggregate_synthetic_data sets_settings}. For example, the learning rate, the number of epochs for the outer loop, and the number of epochs for the inner loop. 

\begin{table}[h]
\centering
\scalebox{0.8} {
\begin{tabular}{|c|c|c|c|c||c|c|c|}
\hline
\multirow{2}{*}{Data Sets} & \multirow{2}{*}{Outliers} & \multicolumn{3}{c||}{Logistic loss} & \multicolumn{3}{c|}{Hinge loss} \\ \cline{3-8} 
                 & &LR    &   \# OE   &   \# IE    &   LR    &   \# OE    &    \# IE    \\ \hline
\multirow{7}{*}{\makecell{Multi-modal data}} &1 &  0.01     &  100     &   1000    &     0.01  &    5   &  1000     \\ \cline{2-8} 
                  &   2 &0.01&  100     &  1000     &   0.01    &   5    &  1000     \\ \cline{2-8} 
                  &   3 &0.01&   100    &  1000     &  0.01     &   5    & 1000      \\ \cline{2-8} 
                  &   4 &0.01&   100    &  1000     &   0.01    &  5     &   1000    \\ \cline{2-8} 
                  &   5 &0.01&   100    &   1000    &   0.01    &  5     &  1000     \\ \cline{2-8} 
                  &   10&0.01&   100    &  1000     &  0.01     &   5    &  1000     \\ \cline{2-8} 
                  &  20 &0.01&   100    &  1000     &   0.01    &  5     &  1000     \\ \hline
 \makecell{Imbalanced  data} &1 &   0.01    &   100    &   1000    &  0.01     & 5      &  1000     \\ \hline
\end{tabular}
}
\begin{tablenotes}\scriptsize
\centering
\item[*] LR: Learning Rate, OE: Outer Epochs, IE: Inner epochs
\end{tablenotes}
\vspace{1mm}
\caption{\aorr~settings on toy experiments.}
\label{tab:aggregate_synthetic_data sets_settings}
\vspace{-5mm}
\end{table}



\subsection{Training Settings on Real Data Sets for Aggregate Loss}
We provide a reference for setting parameters to reproduce our \aorr~ experiments on real data sets. Table \ref{tab:aggregate_real_data sets_logistic_settings} contains the settings for individual logistic loss. Table \ref{tab:aggregate_real_data sets_hinge_settings} is for individual hinge loss. Table \ref{tab:aggregate_MNIST_settings} is for multi-class learning.
 
\begin{table}[H]
\centering
\scalebox{0.8} {
\begin{tabular}{|c|cccccc|}
\hline
        Data Sets   &  $k$   &  $m$ & $C$ &  \# Outer epochs  & \# Inner epochs & Learning rate \\ \hline
        Monk        & 70    & 20 & $10^4$ & 5 & 2000 & 0.01\\ 
        Australian  & 80    & 3 & $10^4$ & 10 & 1000 & 0.01\\ 
        Phoneme     & 1400  & 100 & $10^4$& 10 & 1000 &0.01\\ 
        Titanic     & 500   & 10  &$10^4$ & 10 & 1000 &0.01\\ 
        Splice      & 450   & 50 &$10^4$ & 10 & 1000 &0.01\\ \hline
\end{tabular}
}
\vspace{1mm}
\caption{\aorr~settings on real data sets for individual logistic loss.}
\label{tab:aggregate_real_data sets_logistic_settings}
\end{table}

\begin{table}[h]
\centering
\scalebox{0.8} {
\begin{tabular}{|c|cccccc|}
\hline
        Data Sets   &  $k$   &  $m$ & $C$ &  \# Outer epochs  & \# Inner epochs & Learning rate \\ \hline
        Monk        & 70    & 45 & $10^4$ & 5 & 1000 & 0.01\\ 
        Australian  & 80    & 3 & $10^4$ & 5 & 1000 & 0.01\\ 
        Phoneme     & 1400  & 410 & $10^4$& 10 & 500 &0.01\\ 
        Titanic     & 500   & 10  &$10^4$ & 5 & 500 &0.01\\ 
        Splice      & 450   & 50 &$10^4$ & 10 & 1000 &0.01\\ \hline
\end{tabular}
}
\vspace{1mm}
\caption{\aorr~settings on real data sets for individual hinge loss.}
\label{tab:aggregate_real_data sets_hinge_settings}
\vspace{-5mm}
\end{table}

\begin{table}[h]
\centering
\scalebox{0.8} {
\begin{tabular}{|c|cccc|}
\hline
        Data Sets   &  \makecell{Learning rate\\(``warm-up'')}   &  \makecell{Learning rate\\(``non-warm-up'')} & \# Outer epochs  & \# Inner epochs \\ \hline
        MNIST        & 0.4    & 0.5 & 20 & 5000 \\ \hline
\end{tabular}
}
\vspace{1mm}
\caption{\aorr~settings on MNIST for multi-class learning.}
\label{tab:aggregate_MNIST_settings}
\vspace{-5mm}
\end{table}



\subsection{Training Settings for Multi-label Learning}
\label{sec:multi_label_settings}
The settings for \tkml~on three real data sets are shown in Table \ref{tab:multi_label_settings}. 

\begin{table}[h]
\centering
\scalebox{0.8} {
\begin{tabular}{|c|cccc|}
\hline
        Data Sets   &  $C$   &    \#Outer epochs &  \#Inner epochs   & Learning rate  \\ \hline
        Emotions   &  $10^4$   & 20 & 1000 & 0.1\\ 
        Scene      &  $10^4$ & 20 & 1000 & 0.1\\ 
        Yeast      &  $10^4$ & 20 & 1000 & 0.1\\ \hline
\end{tabular}
}
\vspace{1mm}
\caption{\tkml~settings on each data set.}
\label{tab:multi_label_settings}
\vspace{-5mm}
\end{table}

\subsection{Training Settings for Multi-class Learning}
\label{sec:tkml_multi_class_settings}
Training settings for the MNIST data set in different noise level can be found in Table \ref{tab:multi_class_settings}.

\begin{table}[H]
\centering
\scalebox{0.8} {
\begin{tabular}{|c|ccc|}
\hline
        Noise level   & \#Outer epochs &  \#Inner epochs   & Learning rate  \\ \hline
        0.2   &  27 & 2000 & 0.1\\ 
        0.3      &  25 & 2000 & 0.1\\ 
        0.4      &  21 & 2000 & 0.1\\ \hline
\end{tabular}
}
\vspace{1mm}
\caption{\tkml~settings on the MNIST data set in different noise levels.}
\label{tab:multi_class_settings}
\vspace{-5mm}
\end{table}

\subsection{Training Settings for \tkml-\aorr~Learning}
\label{sec:tkml_aorr_settings}
Training settings for the Yeast data set in different noise level can be found in Table \ref{tab:tkml_aorr_settings}.

\begin{table}[H]
\centering
\scalebox{0.8} {
\begin{tabular}{|c|cc|}
\hline
        Noise level   & \#Epochs    & Learning rate  \\ \hline
        0   &  1000 & 0.3\\ 
        0.1   &  1000 & 0.3\\ 
        0.2      &  1000  & 0.3\\ 
        0.3      &  1000 & 0.3\\ \hline
\end{tabular}
}
\vspace{1mm}
\caption{\tkml-\aorr~settings on the Yeast data set in different noise levels.}
\label{tab:tkml_aorr_settings}
\vspace{-5mm}
\end{table}

\vskip 0.2in
\bibliography{ArXiv_version}

\end{document}